\documentclass[english,notitlepage,superscriptaddress,nofootinbib]{revtex4-1}
\usepackage[T1]{fontenc}
\usepackage{color}
\usepackage{babel}
\usepackage{mathrsfs}
\usepackage{amsmath}
\usepackage{amssymb}
\usepackage{graphicx}
\usepackage[unicode=true,pdfusetitle,
 bookmarks=true,bookmarksnumbered=false,bookmarksopen=false,
 breaklinks=false,pdfborder={0 0 0},pdfborderstyle={},backref=false,colorlinks=true]
 {hyperref}
\hypersetup{
 linkcolor=burntgreen,citecolor=red,urlcolor=burntorange}

\makeatletter

\newcommand{\lyxdot}{.}

\usepackage{color}
\definecolor{burntorange}{rgb}{0.8, 0.33, 0.0}
\definecolor{charcoal}{rgb}{0.21, 0.27, 0.31}
\definecolor{coolblack}{rgb}{0.0, 0.28, 0.49}
\definecolor{burntgreen}{rgb}{0.05, 0.45, 0.27}
\definecolor{burntblue}{rgb}{0.05, 0.27, 0.8}

\renewcommand{\doi}[1]{\href{http://dx.doi.org/#1}{\texttt{doi:#1}}}

\setcitestyle{authoryear,round}
\makeatother

\begin{document}
\title{Shaping the learning landscape in neural networks around wide flat
minima}
\author{Carlo Baldassi}
\affiliation{Artificial Intelligence Lab, Institute for Data Science and Analytics,
Bocconi University, Milano, Italy}
\affiliation{Istituto Nazionale di Fisica Nucleare, Sezione di Torino, Italy}
\author{Fabrizio Pittorino}
\affiliation{Artificial Intelligence Lab, Institute for Data Science and Analytics,
Bocconi University, Milano, Italy}
\affiliation{Politecnico di Torino, Italy}
\author{Riccardo Zecchina}
\affiliation{Artificial Intelligence Lab, Institute for Data Science and Analytics,
Bocconi University, Milano, Italy}
\affiliation{International Centre for Theoretical Physics, Trieste, Italy}
\begin{abstract}
Learning in Deep Neural Networks (DNN) takes place by minimizing a
non-convex high-dimensional loss function, typically by a stochastic
gradient descent (SGD) strategy. The learning process is observed
to be able to find good minimizers without getting stuck in local
critical points, and that such minimizers are often satisfactory at
avoiding overfitting. How these two features can be kept under control
in nonlinear devices composed of millions of tunable connections is
a profound and far reaching open question. In this paper we study
basic non-convex one- and two-layer neural network models which learn
random patterns, and derive a number of basic geometrical and algorithmic
features which suggest some answers. We first show that the error
loss function presents few extremely wide flat minima (WFM) which
coexist with narrower minima and critical points. We then show that
the minimizers of the cross-entropy loss function overlap with the
WFM of the error loss. We also show examples of learning devices for
which WFM do not exist. From the algorithmic perspective we derive
entropy driven greedy and message passing algorithms which focus their
search on wide flat regions of minimizers. In the case of SGD and
cross-entropy loss, we show that a slow reduction of the norm of the
weights along the learning process also leads to WFM. We corroborate
the results by a numerical study of the correlations between the volumes
of the minimizers, their Hessian and their generalization performance
on real data.

\tableofcontents{}
\end{abstract}
\maketitle

\section{Introduction}

Artificial Neural Networks (ANN), currently also known as Deep Neural
Networks (DNN) when they have more than two layers, are powerful nonlinear
devices used to perform different types of learning tasks \citep{mackay2003information}.
From the algorithmic perspective, learning in ANN is in principle
a hard computational problem in which a huge number of parameters,
the connection \emph{weights}, need to be optimally tuned. Yet, at
least for supervised pattern recognition tasks, learning has become
a relatively feasible process in many applications across domains
and the performances reached by DNNs have had a huge impact on the
field of Machine Learning (ML).

DNN models have evolved very rapidly in the last decade, mainly by
an empirical trial and selection process guided by heuristic intuitions.
As a result, current DNN are in a sense akin to complex physical or
biological systems, which are known to work but for which a detailed
understanding of the principles underlying their functioning remains
unclear. The tendency to learn efficiently and to generalize with
limited overfitting are two properties that often coexist in DNN and
yet a unifying theoretical framework is still missing.

Here we provide analytical results on the geometrical structure of
the loss landscape of ANN which shed light on the success of Deep
Learning (DL) \citep{lecun2015deep} algorithms and allow us to design
novel efficient algorithmic schemes.

We focus on non-convex one- and two-layer ANN models that exhibit
sufficiently complex behavior and yet are amenable to detailed analytical
and numerical studies. Building on methods of statistical physics
of disordered systems, we analyze the complete geometrical structure
of the minimizers of the loss function of ANN learning random patterns
and discuss how the current DNN models are able to exploit such structure,
e.g. starting from the choice of the loss function, avoiding algorithmic
traps and reaching rare solutions which belong to wide flat regions
of the weight space. In our study the notion of flatness is given
in terms of the volume of the weights around a minimizer which do
not lead to an increase of the loss value. This generalizes the so
called Local Entropy of a minimizer \citep{baldassi_subdominant_2015},
defined for discrete weights as the log of the number of optimal weights
assignments within a given Hamming distance from the reference minimizer.
We call these regions High Local Entropy (HLE) regions for discrete
weights or Wide Flat Minima (WFM) for continuous weights. Our results
are derived analytically for the case of random data and corroborated
by numerics on real data. In order to eliminate ambiguities which
may arise from changes of scale of the weights, we control the norm
of the weights in each of the units that compose the network. The
outcomes of our study can be summarized as follows.

(i) We show analytically that ANN learning random patterns possess
the structural property of having extremely robust regions of optimal
weights\emph{, }namely wide flat minima of the loss, whose existence
is important to achieve convergence in the learning process. Though
these wide minima are rare compared to the dominant critical points
(absolute narrow minima, local minima or saddle points in the loss
surface), they can be accessed by a large family of simple learning
algorithms. We also show analytically that other learning machines,
such as the Parity Machine, do not possess wide flat minima.

(ii) We show analytically that the choice of the cross-entropy loss
function has the effect of biasing learning algorithms toward HLE
or WFM regions.

(iii) We derive a greedy algorithm -- Entropic Least Action Learning
(eLAL) -- and a message passing algorithm -- focusing Belief Propagation
(fBP) -- which zoom in their search on wide flat regions of minimizers.

(iv) We compute the volumes associated to the minimizers found by
different algorithms using Belief Propagation.

(v) We show numerically that the volumes correlate well with the spectra
of the Hessian on computationally tractable networks and with the
generalization performance on real data. The algorithms which search
for WFM display a spectrum which is much more concentrated around
zero eigenvalues compared to plain SGD.

Our results on random patterns support the conclusion that the minimizers
which are relevant for learning are not the most frequent isolated
and narrow ones (which also are computationally hard to sample) but
the rare ones which are extremely wide. While this phenomenon was
recently disclosed for the case of discrete weights \citep{baldassi_subdominant_2015,baldassi_unreasonable_2016},
here we demonstrate that it is present also in non convex ANN with
continuous weights. Building on these results we derive novel algorithmic
schemes and shed light on the performance of SGD with the cross-entropy
loss function. Numerical experiments suggest that the scenario generalizes
to real data and is consistent with other numerical results on deeper
ANN \citep{keskar2016large}.

\section{\label{sec:HLE/WFM-regions-exist}HLE/WFM regions exist in non convex
neural devices storing random patterns}

In what follows we analyze the geometrical structure of the weights
space by considering the simplest non convex neural devices storing
random patterns: the single layer network with discrete weights and
the two layer networks with both continuous and discrete weights.
The choice of random patterns, for which no generalization is possible,
is motivated by the possibility of using analytical techniques from
statistical physics of disordered systems and by the fact that we
want to identify structural features which do not depend on specific
correlation patterns of the data.

\subsection{The simple example of discrete weights}

In the case of binary weights it is well known that even for the single
layer network the learning problem is computationally challenging.
Therefore we begin our analysis by studying the so called binary perceptron,
which maps vectors of $N$ inputs $\xi\in\left\{ -1,1\right\} ^{N}$
to binary outputs as $\sigma\left(W,\xi\right)=\textrm{sign}\left(W\cdot\xi\right)$,
where $W\in\left\{ -1,1\right\} ^{N}$ is the \emph{synaptic} weights
vector $W=\left(w_{1},\,w_{2},\,...,\,w_{N}\right)$.

Given a training set composed of $\alpha N$ input patterns $\xi^{\mu}$
with $\mu\in\left\{ 1,\dots,\alpha N\right\} $ and their corresponding
desired outputs $\sigma^{\mu}\in\left\{ -1,1\right\} ^{\alpha N}$,
the learning problem consists in finding a \emph{solution} $W$ such
that $\sigma\left(W,\xi^{\mu}\right)=\sigma^{\mu}$ for all $\mu$.
The entries $\xi_{i}^{\mu}$ and the outputs $\sigma^{\mu}$ are random
unbiased i.i.d.~variables. As discussed in \citet{krauth-mezard}
(but see also the rigorous bounds in \citet{ding2019capacity}), perfect
classification is possible with probability $1$ in the limit of large
$N$ up to a critical value of $\alpha$, usually denoted as $\alpha_{c}$;
above this value, the probability of finding a solution drops to zero.
$\alpha_{c}$ is called the \emph{capacity} of the device.

The standard analysis of this model is based on the study of the zero
temperature limit of the Gibbs measure with a loss (or energy) function
${\cal L}_{\mathrm{NE}}$ which counts the number of errors (NE) over
the training set:

\begin{equation}
{\cal L}_{\mathrm{NE}}=\sum_{\mu=1}^{\alpha N}\Theta\left(-\sigma^{\mu}\sigma\left(W,\xi^{\mu}\right)\right)\label{eq:L_NE}
\end{equation}
where $\Theta\left(x\right)$ is the Heaviside step function, $\Theta\left(x\right)=1$
if $x>0$ and $0$ otherwise. The Gibbs measure is given by

\begin{equation}
P\left(W\right)=\frac{1}{Z\left(\beta\right)}\,\exp\left(-\beta\sum_{\mu=1}^{\alpha N}\Theta\left(-\sigma^{\mu}\sigma\left(W,\xi^{\mu}\right)\right)\right)
\end{equation}
where $\beta\ge0$ is the inverse temperature parameter. For large
values of $\beta$, $P\left(W\right)$ concentrates on the minima
of ${\cal L}_{\mathrm{NE}}$. The key analytical obstacle for the
computation of $P\left(W\right)$ is the evaluation of the normalization
factor, the partition function $Z$:

\begin{equation}
Z\left(\beta\right)=\sum_{\left\{ w_{i}=\pm1\right\} }\exp\left(-\beta\sum_{\mu=1}^{\alpha N}\Theta\left(-\sigma^{\mu}\sigma\left(W,\xi^{\mu}\right)\right)\right)
\end{equation}
In the the zero temperature limit ($\beta\to\infty)$ and below $\alpha_{c}$
the partition function simply counts all solutions to the learning
problem,

\begin{align}
Z_{\infty} & =\lim_{\beta\to\infty}Z\left(\beta\right)=\sum_{\left\{ W\right\} }\mathbb{X}_{\xi}\left(W\right)
\end{align}
where $\mathbb{X}_{\xi}\left(W\right)=\prod_{\mu=1}^{\alpha N}\Theta\left(\sigma^{\mu}\sigma\left(W,\xi^{\mu}\right)\right)$
is a characteristic function which evaluates to one if all patterns
are correctly classified, and to zero otherwise.

$Z_{\infty}$ is an exponentially fluctuating quantity (in $N$),
and its most probable value is obtained by exponentiating the average
of $\log Z_{\infty}$, denoted by $\left\langle \log Z_{\infty}\right\rangle _{\xi}$,
over the realizations of the patterns

\begin{equation}
Z_{\infty,\mathrm{typical}}\simeq\exp\left(N\left\langle \ln Z_{\infty}\right\rangle _{\xi}\right).\label{eq:Z_typ}
\end{equation}
The calculation of $\left\langle \log Z_{\infty}\right\rangle _{\xi}$
has been done in the 80s and 90s by the replica and the cavity methods
of statistical physics and, as mentioned above, the results predict
that the learning task undergoes a threshold phenomenon at $\alpha_{c}=0.833$,
where the probability of existence of a solution jumps from one to
zero in the large $N$ limit \citep{krauth-mezard}. This result has
been put recently on rigorous grounds by \citet{ding2019capacity}.
Similar calculations predict that for any $\alpha\in\left(0,\alpha_{c}\right)$,
the vast majority of the exponentially numerous solutions on the hypercube
$W\in\left\{ -1,1\right\} ^{N}$ are isolated, separated by a $O\left(N\right)$
Hamming mutual distance \citep{huang2014origin}. In the same range
of $\alpha$, there also exist an even larger number of local minima
at non-zero loss, a result that has been corroborated by analytical
and numerical findings on stochastic learning algorithms which satisfy
detailed balance \citep{horner1992dynamics}. Recently it became clear
that by relaxing the detailed balance condition it was possible to
design simple algorithms which can solve the problem efficiently \citep{braunstein2006learning,baldassi2007efficient,baldassi2009generalization}.

\subsection{Local entropy theory \label{subsec:Local-entropy-binary-perc}}

The existence of effective learning algorithms indicates that the
traditional statistical physics calculations, which focus on set of
solutions that dominate the zero temperature Gibbs measure (i.e.~the
most numerous ones), are effectively blind to the solutions actually
found by such algorithms. Numerical evidence suggests that in fact
the solutions found by heuristics are not at all isolated: on the
contrary, they appear to belong to regions with a high density of
nearby other solutions. This puzzle has been solved very recently
by an appropriate large deviations study \citep{baldassi_subdominant_2015,baldassi_local_2016,baldassi2016learning,baldassi_unreasonable_2016}
in which the tools of statistical physics have been used to study
the most probable value of the \emph{local entropy} of the loss function,
i.e.~a function that is able to detect the existence of regions with
an $O\left(N\right)$ radius containing a high density of solutions
even when the number of these regions is small compared to the number
of isolated solutions. For binary weights the local entropy function
is the (normalized) logarithm of the number of solutions $W^{\prime}$
at Hamming distance $D\,N$ from a reference solution $W$

\begin{equation}
\mathscr{E}_{D}\left(W\right)=-\frac{1}{N}\ln\mathcal{U}\left(W,D\right)\label{eq:energy_LE}
\end{equation}
with
\begin{equation}
\mathcal{U}\left(W,D\right)=\sum_{\left\{ W^{\prime}\right\} }\mathbb{X}_{\xi}\left(W^{\prime}\right)\delta\left(W^{\prime}\cdot W,N\left(1-2D\right)\right)
\end{equation}
and where $\delta$ is the Kronecker delta symbol. In order to derive
the typical values that the local entropy can take, one needs to compute
the Gibbs measure of the local entropy

\begin{equation}
P_{\mathrm{LE}}\left(W\right)=\frac{1}{Z_{\mathrm{LE}}}\,\exp\left(-y\mathscr{E}_{D}\left(W\right)\right)\label{eq:Gibbs_LE}
\end{equation}
where $y$ has the role of an inverse temperature. For large values
of $y$ this probability measure focuses on the $W$ surrounded by
an exponential number of solutions within a distance $D$. The regions
of high local entropy (HLE) are then described in the regime of large
$y$ and small $D$. In particular, the calculation of the expected
value of the optimal \emph{local entropy} 
\begin{equation}
\mathscr{S}\left(D\right)\equiv\mathscr{E}_{D}^{\textrm{opt}}=\max_{\left\{ W\right\} }\left\{ -\frac{1}{N}\left\langle \ln\mathcal{U}\left(W,D\right)\right\rangle _{\xi}\right\} 
\end{equation}
 shows the existence of extremely dense clusters of solutions up to
values of $\alpha$ close to $\alpha_{c}$ \citep{baldassi_subdominant_2015,baldassi_local_2016,baldassi2016learning,baldassi_unreasonable_2016}.

The probability measure eq.~(\ref{eq:Gibbs_LE}) can be written in
an equivalent form that generalizes to the non zero errors regime,
is analytically simpler to handle, and leads to novel algorithmic
schemes \citep{baldassi_unreasonable_2016}:
\begin{equation}
P_{\mathrm{LE}}\left(W\right)\sim P\left(W;\beta,y,\lambda\right)=Z\left(\beta,y,\lambda\right)^{-1}e^{y\,\Phi\left(W,\beta,\lambda\right)}.\label{eq:prob_large_dev}
\end{equation}
where $\Phi\left(W,\beta,\lambda\right)$ is a ``local free entropy''
potential in which the distance constraint is forced through a Lagrange
multiplier $\lambda$
\begin{equation}
\Phi\left(W,\beta,\lambda\right)=\ln\sum_{\left\{ W^{\prime}\right\} }e^{-\beta{\cal L}_{\mathrm{NE}}\left(W^{\prime}\right)-\lambda\,d\left(W,W^{\prime}\right)}\label{eq:local_free_entropy}
\end{equation}
where $d\left(\cdot,\cdot\right)$ is some monotonically increasing
function of the distance between configurations, defined according
to the type of weights under consideration. In the limit $\beta\to\infty$
and by choosing $\lambda$ so that a given distance is selected, this
expression reduces to eq.~(\ref{eq:Gibbs_LE}).

The crucial property of eq.~(\ref{eq:prob_large_dev}) comes from
the observation that by choosing $y$ to be a non-negative integer,
the partition function can be rewritten as:
\begin{eqnarray}
Z\left(\beta,y,\lambda\right) & = & \sum_{\left\{ W\right\} }e^{y\,\Phi\left(W,\beta,\lambda\right)}\nonumber \\
 & = & \sum_{\left\{ W\right\} }\sum_{\left\{ W^{\prime a}\right\} _{a=1}^{y}}e^{-\beta{\cal L}_{\mathrm{R}}\left(W,W^{\prime a}\right)}\label{eq:part_func}
\end{eqnarray}
where
\begin{equation}
{\cal L}_{\mathrm{R}}\left(W,W^{\prime a}\right)=\sum_{a=1}^{y}{\cal L}_{\mathrm{NE}}\left(W^{\prime a}\right)-\frac{\lambda}{\beta}\sum_{a=1}^{y}d\left(W,W^{\prime a}\right).\label{eq:L_R}
\end{equation}
These are the partition function and the effective loss of $y+1$
interacting real replicas of the system, one of which acts as reference
system ($W)$ while the remaining $y$ ($\left\{ W^{\prime a}\right\} $)
are identical, each being subject to the energy constraint ${\cal L}_{\mathrm{NE}}\left(W^{\prime a}\right)$
and to the interaction term with the reference system. As discussed
in \citet{baldassi_unreasonable_2016}, several algorithmic schemes
can be derived from this framework by minimizing ${\cal L}_{\mathrm{R}}$.
Here we shall also use the above approach to study the existence of
WFMs in continuous models and to design message passing and greedy
learning algorithms driven by the local entropy of the solutions.

\subsection{Two layer networks with continuous weights}

As for the discrete case, we are able to show that in non convex networks
with continuous weights the WFMs exist, are rare and yet accessible
to simple algorithms. In order to perform an analytic study, we consider
the simplest non-trivial two-layer neural network, the committee machine
with non-overlapping receptive fields. It consists of $N$ input units,
one hidden layer with $K$ units and one output unit. The input units
are divided into $K$ disjoint sets of $\tilde{N}=\frac{N}{K}$ units.
Each set is connected to a different hidden unit. The input to the
$\ell$-th hidden unit is given by $x_{\ell}^{\mu}=\frac{1}{\sqrt{\tilde{N}}}\sum_{i=1}^{\tilde{N}}w_{\ell i}\xi_{\ell i}^{\mu}$
where $w_{\ell i}\in\mathbb{R}$ is the connection weight between
the input unit $i$ and the hidden unit $\ell$, and $\xi_{\ell i}^{\mu}$
is the $i$-th input to the $\ell$-th hidden unit. As before, $\mu$
is a pattern index. We study analytically the pure classifier case
in which each unit implements a threshold transfer function and the
loss function is the error loss. Other types of (smooth) functions,
more amenable to numerical simulation, will be also discussed in a
subsequent section. The output of the $\ell$-th hidden unit is given
by

\begin{equation}
\tau_{\ell}^{\mu}=\mathrm{sign}\left(x_{\ell}^{\mu}\right)=\mathrm{sign}\left(\frac{1}{\sqrt{\tilde{N}}}\sum_{i=1}^{\tilde{N}}w_{\ell i}\xi_{\ell i}^{\mu}\right)\label{eq:sign_activation}
\end{equation}
In the second layer all the weights are fixed and equal to one, and
the overall output of the network is simply given by a majority vote
$\sigma_{\mathrm{out}}^{\mu}=\mathrm{sign}\left(\frac{1}{\sqrt{K}}\sum_{\ell}\tau_{\ell}^{\mu}\right).$

As for the binary perceptron, the learning problem consists in mapping
each of the random input patterns $\left(\xi_{\ell i}^{\mu}\right)$
with ($\ell=1,\dots,K$, $i=1,\dots,\tilde{N}$, $\mu=1,\dots,\alpha N$)
onto a randomly chosen output $\sigma^{\mu}.$ Both $\xi_{\ell i}^{\mu}$
and $\sigma^{\mu}$ are independent random variables which take the
values $\pm1$ with equal probability. For a given set of patterns,
the volume of the subspace of the network weights which correctly
classifies the patterns, the so called \emph{version space}, is given
by

\begin{equation}
V=\int\prod_{i\ell}\mathrm{d}w_{\ell i}\prod_{\ell}\delta\left(\sum_{i}w_{\ell i}^{2}-\tilde{N}\right)\prod_{\mu}\Theta\left(\sigma^{\mu}\sigma_{\mathrm{out}}^{\mu}\right)\label{eq:volume}
\end{equation}
where we have imposed a spherical constraint on the weights via a
Dirac $\delta$ in order to keep the volume finite (though exponential
in $N)$. In the case of binary weights the integral would become
a sum over all the $2^{N}$ configurations and the volume would be
the overall number of zero error assignments of the weights.

The committee machine has been studied extensively in the 90s \citep{barkai1992broken,schwarze1992generalization,engel1992storage}.
The capacity of the network can be derived by computing the typical
weight space volume as a function of the number of correctly classified
patterns $\alpha N$, in the large $N$ limit. As for the binary case,
the most probable value of $V$ is obtained by exponentiating the
average of $\log V$, $V_{\mathrm{typical}}\simeq\exp\left(N\left\langle \log V\right\rangle _{\xi}\right)$,
a difficult task which is achieved by the replica method \citep{mezard1987spin,barkai1990statistical}.

For the smallest non-trivial value of $K$, $K=3,$ it has been found
that above $\alpha_{0}\simeq1.76$ the space of solutions changes
abruptly, becoming clustered into multiple components\footnote{Strictly speaking each cluster is composed of a multitude of exponentially
small domains \citep{monasson1995weight}.}. Below $\alpha_{0}$ the geometrical structure is not clustered and
can be described by the simplest version of the replica method, known
as replica symmetric (RS) solution. Above $\alpha_{0}$ the analytical
computation of the typical volume requires a more sophisticated analysis
that properly describes a clustered geometrical structure. This analysis
can be performed by a variational technique which is known in statistical
physics as the replica-symmetry-breaking (RSB) scheme, and the clustered
geometrical structure of the solution space is known as RSB phase.

The capacity of the network, above which perfect classification becomes
impossible, is found to be $\alpha_{c}\simeq$3.02. In the limit of
large $K$ (but still with $\tilde{N}\gg1$), the clustering transition
occurs at a finite number of patterns per weight, $\alpha_{0}\simeq2.95$
\citep{barkai1992broken} whereas the critical capacity grows with
$K$ as $\alpha_{c}\propto\sqrt{\ln K}$ \citep{monasson1995weight}.

\subsection{\label{subsec:WFM}The existence of wide flat minima}

\begin{figure}
\begin{centering}
\includegraphics[width=0.6\columnwidth]{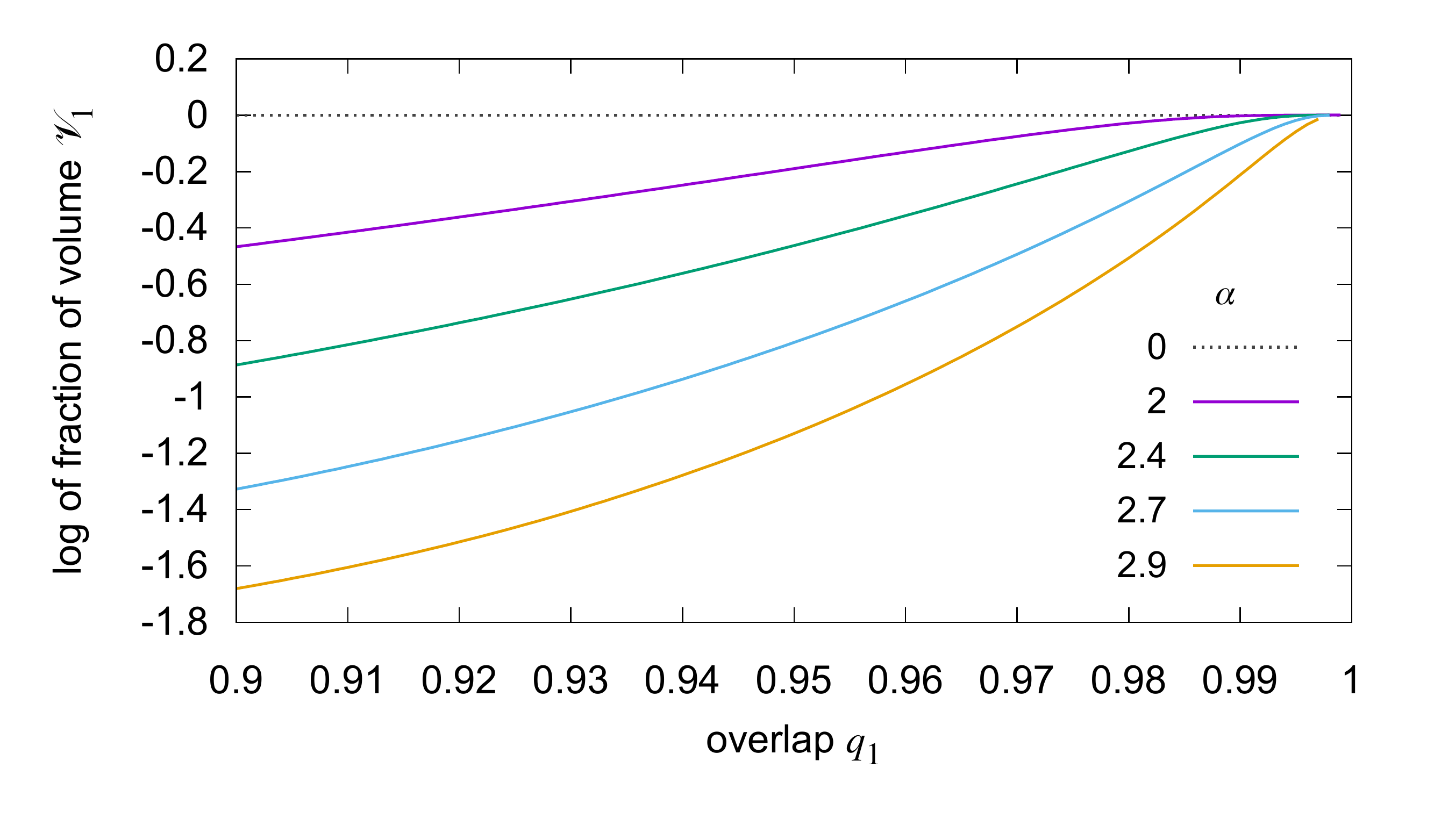}
\par\end{centering}
\begin{centering}
\includegraphics[width=0.6\columnwidth]{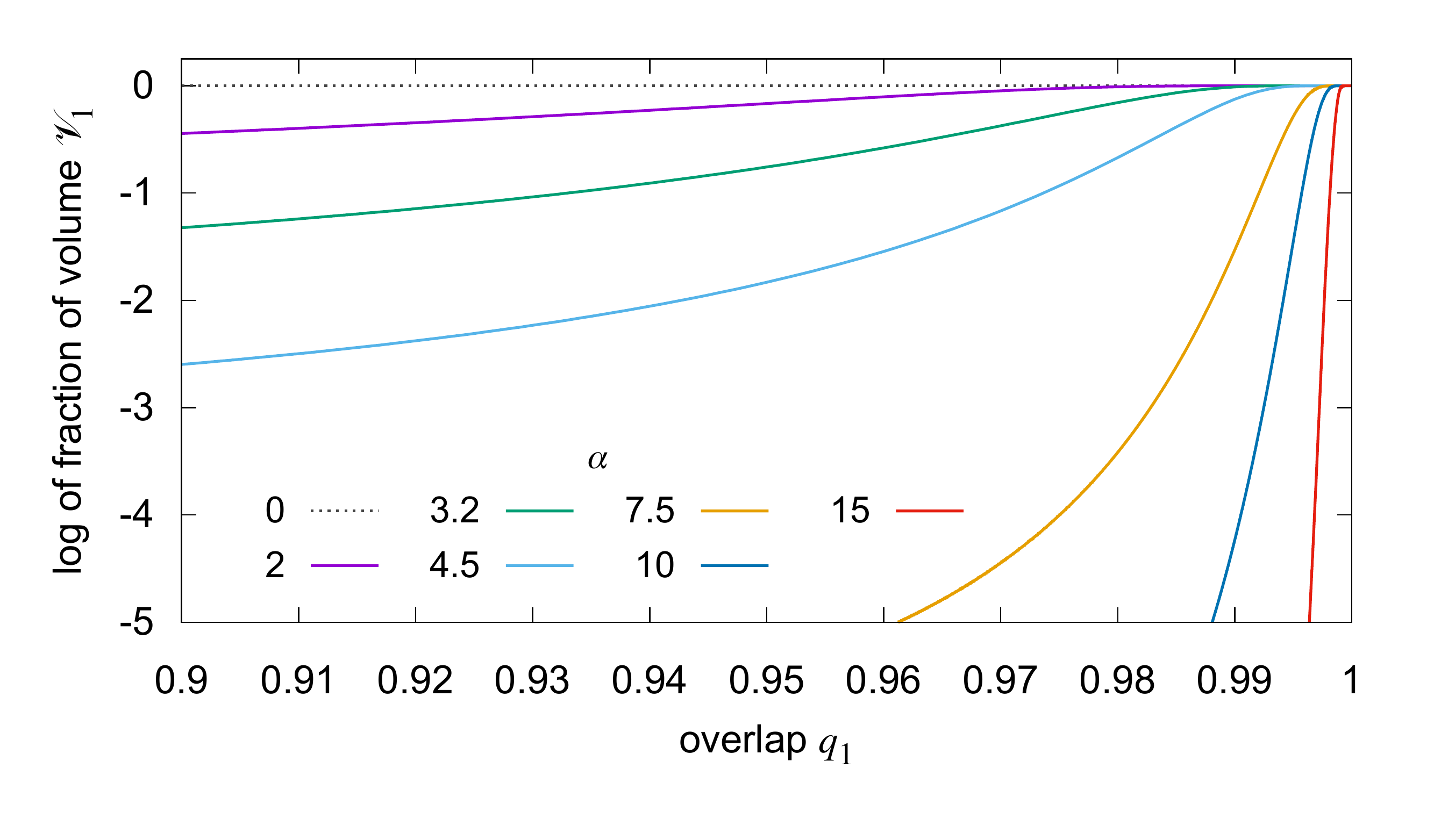}
\par\end{centering}
\caption{\label{fig:treeK3}Normalized local entropy $\mathscr{V}_{1}$ vs
$q_{1}$, i.e.~logarithm of the fraction of configurations of $y$
real replicas at mutual overlap $q_{1}$ in which all replicas have
zero error loss $\mathcal{L}_{\mathrm{NE}}$. The curves are for a
tree-like committee machine trained on $\alpha N$ random patterns,
for various values of $\alpha$, obtained from a replica calculation
in the limit of large $N$ and large number of real replicas $y$.
When the curves approach $0$ as $q_{1}\to1$ it means that nearly
all configurations are minima, thus that the replicas are collectively
exploring a wide minimum (any $q_{1}<1$ implies distances of $O\left(N\right)$
between replicas). Top: the case of $K=3$ units. Bottom: the limiting
case of a large number of hidden units $K$.}
\end{figure}

In order to detect the existence of WFM we use a large deviation measure
which is the continuous version of the measure used in the discrete
case: each configuration of the weights is re-weighted by a local
volume term, analogously to the analysis of section~~\ref{subsec:Local-entropy-binary-perc}.
For the continuous case, however, we adopt a slightly different formalism
which simplifies the analysis. Instead of constraining the set of
$y$ real replicas\footnote{Not to be confused with the virtual replicas of the replica method.}
to be at distance $D$ from a reference weight vector, we can identify
the same WFM regions by constraining them directly to be at a given
mutual overlap: for a given value $q_{1}\in\left[-1,1\right]$, we
impose that $W^{a}\cdot W^{b}=Nq_{1}$ for all pairs of distinct replicas
$a,b$. The overlap $q_{1}$ is bijectively related to the mutual
distance among replicas (which tends to $0$ as $q_{1}\to1$). That,
in turn, determines the distance between each replica and the barycenter
of the group $\frac{1}{y}\sum_{a}W^{a}$, which takes the role that
the reference vector had in the previous treatment. Thus, the regime
of small $D$ corresponds to the regime of $q_{1}$ close to $1$,
and apart from this reparametrization the interpretation of the WFM
volumes is the same. As explained in Appendix~\ref{sec:A-WFM}, the
advantage of this novel technique is that it allows to use directly
the first-step formalism of the RSB scheme (1-RSB). Similarly to the
discrete case, the computation of the maximal WFM volumes leads to
the following results: for $K=3$ and in the large $y$ limit, we
find\footnote{All the details are reported in Appendix~\ref{sec:A-WFM}.}

\begin{align*}
\mathscr{V}\left(q_{1}\right) & =\max_{\left\{ W^{a}\right\} _{a=1}^{y}}\left\{ \frac{1}{Ny}\langle\ln V\left(\left(W^{a}\right)_{a=1}^{y},q_{1}\right)\rangle_{\xi}\right\} \\
 & =G_{S}\left(q_{1}\right)+\alpha G_{E}\left(q_{1}\right)
\end{align*}
with
\begin{eqnarray*}
G_{\mathrm{S}}\left(q_{1}\right) & = & \frac{1}{2}\left[1+\ln2\pi+\ln\left(1-q_{1}\right)\right]\\
G_{\mathrm{E}}\left(q_{1}\right) & = & \int\prod_{\ell=1}^{3}Dv_{\ell}\max_{u_{1},u_{2},u_{3}}\left[-\frac{\sum_{\ell=1}^{3}u_{\ell}^{2}}{2}+\ln\left(\tilde{H}_{1}\tilde{H}_{2}+\tilde{H}_{1}\tilde{H}_{3}+\tilde{H}_{2}\tilde{H}_{3}-2\tilde{H}_{1}\tilde{H}_{2}\tilde{H}_{3}\right)\right]
\end{eqnarray*}
where $\tilde{H}_{\ell}\equiv H\left(\sqrt{\frac{d_{0}}{1-q_{1}}}u_{\ell}+\sqrt{\frac{q_{0}}{1-q_{1}}}v_{\ell}\right)$,
$H\left(x\right)\equiv\int_{x}^{\infty}Dv$, $Dv\equiv dv\frac{1}{\sqrt{2\pi}}e^{-\frac{1}{2}v^{2}}$,
$d_{0}\equiv y\left(q_{1}-q_{0}\right)$ and $q_{0}$ satisfies a
saddle point equation that needs to be solved numerically. $\mathscr{V}\left(q_{1}\right)$
is the logarithm of the volume of the solutions, normalized by $Ny$,
under the spherical constraints on the weights, and with the real
replicas forced to be at a mutual overlap $q_{1}$. In analogy with
the discrete case, we still refer to $\mathscr{V}\left(q_{1}\right)$
as to the \emph{local entropy}. It is composed by the sum of two terms:
the first one, $G_{\mathrm{S}}\left(q_{1}\right)$, corresponds to
the log-volume at $\alpha=0$, where all configurations are solutions
and only the geometric constraints are present. This is an upper bound
for the local entropy. The second term, $\mathscr{V}_{1}\left(q_{1}\right)\equiv\alpha G_{\mathrm{E}}\left(q_{1}\right)$,
is in general negative and it represents the log of the \emph{fraction}
of solutions at overlap $q_{1}$ (among all configurations at that
overlap), and we call it \emph{normalized} local entropy. Following
the interpretation given above, we expect (in analogy to the discrete
case at small $D$, cf.~fig.~\ref{fig:FP}) that in a WFM this fraction
is close to $1$ (i.e. $\mathscr{V}_{1}$ is close to $0$) in an
extended region where the distance between the replicas is small,
i.e. where $q_{1}$ is close to $1$; otherwise, WFMs do not exist
in the model. In fig.\textbf{~}\ref{fig:treeK3} (top panel) we report
the values of $\mathscr{V}_{1}\left(q_{1}\right)$ vs the overlap
$q_{1}$, for different values of $\alpha.$ Indeed, one may observe
that the behavior is qualitatively similar to that of the binary perceptron:
besides the solutions and all the related local minima and saddles
predicted by the standard statistical physics analysis \citep{barkai1992broken,schwarze1992generalization,engel1992storage,monasson1995weight},
there exist absolute minima which are flat at relatively large distances.
Indeed, reaching such wide minima efficiently is non trivial, and
different algorithms can have drastically different behaviors, as
we will discuss in detail in sec\@.~\ref{sec:numerics}.

The case $K=3$ is still relatively close to the simple perceptron,
though the geometrical structure of its minima is already dominated
by non convex features for \textbf{$\alpha>1.76$}. A case which is
closer to more realistic ANNs is $K\gg1$ (but still $N\gg K$), which,
luckily enough, is easier to study analytically. We find:

\begin{align*}
G_{\mathrm{S}}\left(q_{1}\right) & =\frac{1}{2}\left[1+\ln2\pi+\ln\left(1-q_{1}\right)\right]\\
G_{\mathrm{E}}\left(q_{1}\right) & =\!\int\!\!Dv\,\max_{u}\left[-\frac{u^{2}}{2}+\ln H\left(\sqrt{\frac{\Delta q_{1}^{e}}{1-q_{1}^{e}}}u+\frac{q_{0}^{e}}{1-q_{1}^{e}}v\right)\right]
\end{align*}
where $\Delta q_{1}^{e}=q_{1}^{e}-q_{0}^{e}$ with $q_{1}^{e}\equiv1-\frac{2}{\pi}\arccos\left(q_{1}\right)$,
$q_{0}^{e}\equiv1-\frac{2}{\pi}\arccos\left(q_{0}\right)$ and $q_{0}$
is fixed by a saddle point equation. In fig.~\ref{fig:treeK3} (bottom
panel) we observe that indeed WFM still exist for all finite values
of $\alpha$.

The results of the above WFM computation may require small corrections
due to RSB effects, which however are expected to be very tiny due
to the compact nature of the space of solutions at small distances.

A more informative aspect is to study the volumes around the solutions
found by different algorithms. This can be done by the Belief Propagation
method, similarly to the computation of the weight enumerator function
in error correcting codes \citep{di2006weight}.

\subsection{Not all devices are appropriate: the Parity Machine does not display
HLE/WFM regions}

The extent by which a given model exhibits the presence of WFM can
vary (see e.g.~fig.~\ref{fig:treeK3}). A direct comparison of the
local entropy curves on different models in general does not yet have
a well-defined interpretation, although at least for similar architectures
it can still be informative \citet{baldassi2019properties}. On the
other hand, the existence of WFM itself is a structural property.
For neural networks, its origin relies in the threshold sum form of
the nonlinearity characterizing the formal neurons. As a check of
this claim, we can analyze a model which is in some sense complementary,
namely the so called \emph{parity machine.} We take its network structure
to be identical to the committee machine, except for the output unit
which performs the product of the $K$ hidden units instead of taking
a majority vote. While the outputs of the hidden units are still given
by sign activations, eq.~(\ref{eq:sign_activation}), the overall
output of the network reads $\sigma_{\mathrm{out}}^{\mu}=\prod_{\ell=1}^{K}\tau_{\ell}.$
The volume of the weights that correctly classifies a set of patterns
is still given by eq.~(\ref{eq:volume}).

Parity machines are closely related to error correcting codes based
on parity checks. The geometrical structure of the absolute minima
of the error loss function is known \citep{monasson1995weight} to
be composed by multiple regions, each in one to one correspondence
with the internal representations of the patterns. For random patterns
such regions are typically tiny and we expect the WFM to be absent.
Indeed, the computation of the volume proceeds analogously to the
previous case\footnote{It's actually even simpler, see Appendix~\ref{subsec:A-Parity}.},
and it shows that in this case for any distance the volumes of the
minima are always bounded away from the maximal possible volume, i.e.~the
volume one would find for the same distance when no patterns are stored:
the log-ratio of the two volumes is constant and equal to $-\alpha\log\left(2\right)$.
In other words, the minima never become flat, at any distance scale.

\subsection{The connection between Local Entropy and Cross-Entropy}

Given that dense regions of optimal solutions exist in non convex
ANN, at least in the case of independent random patterns, it remains
to be seen which role they play in current models. Starting with the
case of binary weights, and then generalizing the result to more complex
architectures and to continuous weights, we can show that the most
widely used loss function, the so called Cross-Entropy (CE) loss,
focuses precisely on such rare regions (see \citet{baldassi2018role}
for the case of stochastic weights).

\begin{figure}
\begin{centering}
\includegraphics[width=0.6\columnwidth]{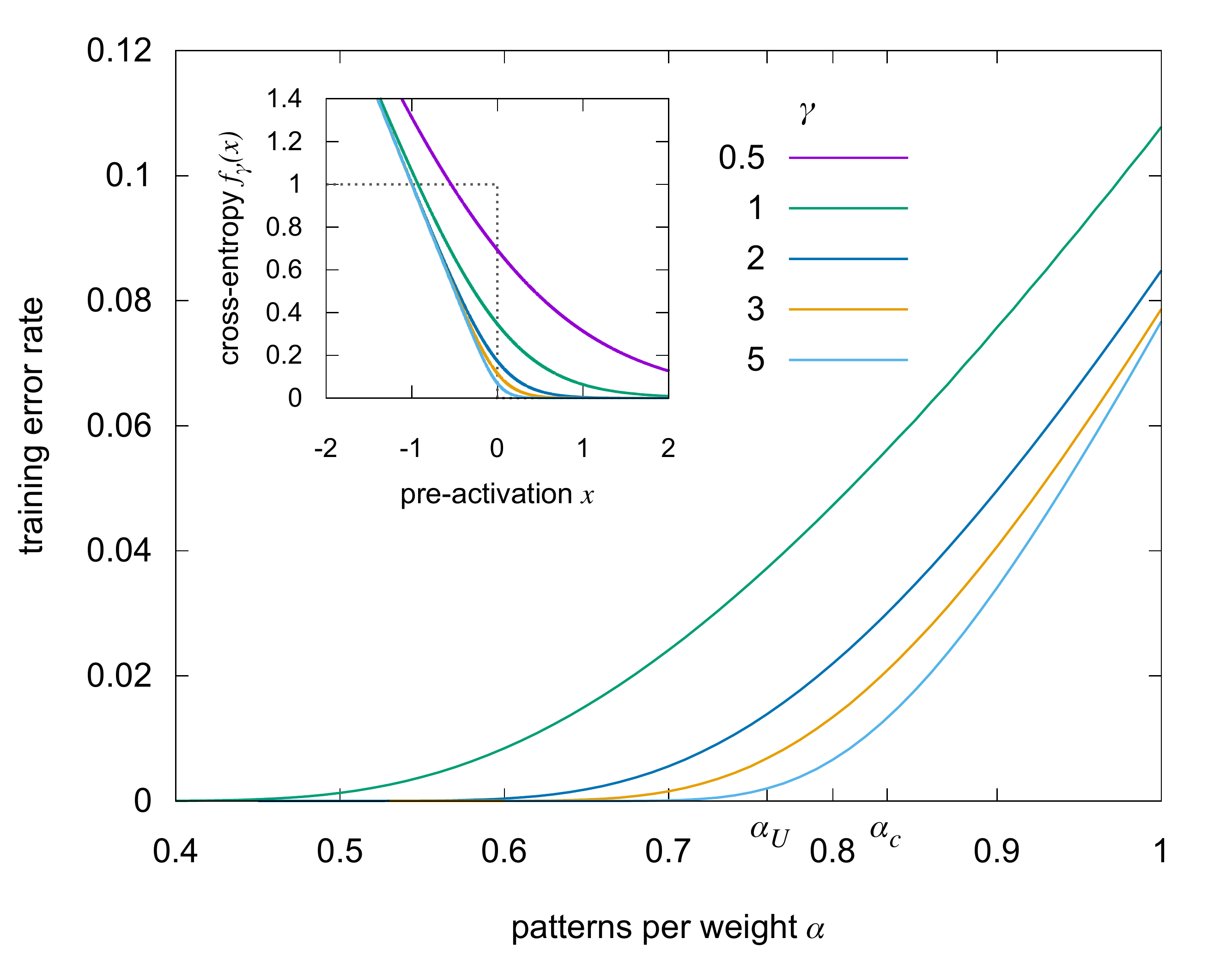}
\par\end{centering}
\caption{\label{fig:errors_vs_alpha}Mean error rate achieved when optimizing
the CE loss in the binary single-layer network, as predicted by the
replica analysis, at various values of $\gamma$ (increasing from
top to bottom). The figure also shows the points $\alpha_{c}\approx0.83$
(up to which solutions exist) and $\alpha_{U}\approx0.76$ (up to
which non-isolated solutions exist). \emph{Inset:} Binary cross-entropy
function $f_{\gamma}\left(x\right)$ for various values of $\gamma$
(increasing from top to bottom). For low values of $\gamma$, the
loss is non-zero even for small positive values of the input, and
thus the minimization procedure tends to favor more robust solutions.
For large values of $\gamma$ the function tends to $\max\left(-x,0\right)$.
The dotted line show the corresponding NE function, which is just
$1$ in case of an error and $0$ otherwise, cf.~eq.~(\ref{eq:L_NE}).}
\end{figure}

For the sake of simplicity, we consider a binary classification task
with one output unit. The CE cost function for each input pattern
reads

\begin{equation}
{\cal L}_{\mathrm{CE}}\left(W\right)=\sum_{\mu=1}^{M}\;f_{\gamma}\left(\frac{\sigma^{\mu}}{\sqrt{\tilde{N}}}\sum_{i=1}^{N}w_{i}\xi_{i}^{\mu}\right)\label{eq:CE_loss}
\end{equation}
 where $f_{\gamma}\left(x\right)=-\frac{x}{2}+\frac{1}{2\gamma}\log\left(2\cosh\left(\gamma x\right)\right)$.
The parameter $\gamma$ allows to control the degree of ``robustness''
of the training, see the inset in figure~\ref{fig:errors_vs_alpha}.
In standard machine learning practice $\gamma$ is simply set to $1$,
but a global rescaling of the weights $W_{i}$ can lead to a basically
equivalent effect. That setting can thus be interpreted as leaving
$\gamma$ as implicit, letting its effective value, and hence the
norm of the weights, to be determined by the initial conditions and
the training algorithm. As we shall see, controlling $\gamma$ explicitly
along the learning process plays a crucial role in finding HLE/WFM
regions.

For the binary case, however, the norm is fixed and thus we must keep
$\gamma$ as an explicit parameter. Note that since $\lim_{\gamma\to\infty}f_{\gamma}\left(x\right)=\max\left(-x,0\right)$
the minima of $\mathcal{L}_{\mathrm{CE}}$ below $\alpha_{c}$ at
large $\gamma$ are the solutions to the training problem, i.e.~they
coincide with those of $\mathcal{L}_{\mathrm{NE}}$.

We proceed by first showing that the minimizers of this loss correspond
to near-zero errors for a wide range of values of $\alpha$, and then
by showing that these minimizers are surrounded by an exponential
number of zero error solutions.

In order to study the probability distribution of the minima of ${\cal L}_{\mathrm{CE}}$
in the large $N$ limit, we need to compute its Gibbs distribution
(in particular, the average of the $\log$ of the partition function,
see eq.~(\ref{eq:Z_typ})) as it has been done for the error loss
$\mathcal{L}_{\mathrm{NE}}$. The procedure follows standard steps
and it is detailed in Appendix~\ref{sec:A-CE}. The method requires
to solve two coupled integral equations as functions of the control
parameters $\alpha$, $\beta$ and $\gamma$. In figure~\ref{fig:errors_vs_alpha}
we show the behavior of the fraction of errors vs the loading $\alpha$,
for various values of $\gamma$. Up to relatively large values of
$\alpha$ the optimum of $\mathcal{L}_{\mathrm{CE}}$ corresponds
to extremely small values of $\mathcal{L}_{\mathrm{NE}}$, virtually
equal to zero for any accessible size $N.$

Having established that by minimizing the the CE one ends up in regions
of perfect classification where the error loss function is essentially
zero, it remains to be understood which type of configurations of
weights are found. Does the CE converge to an isolated point-like
solution in the weight space (such as the typical zero energy configurations
of the error function)\footnote{A quite unlikely fact given that finding isolated solutions is a well
known intractable problem.} or does it converge to the rare regions of high local entropy?

In order to establish the geometrical nature of the typical minima
of the CE loss, we need to compute the average value of $\mathcal{E}_{D}\left(W\right)$
(which tells us how many zero energy configurations of the error loss
function can be found within a given radius $D$ from a given $W$,
see eq.~(\ref{eq:energy_LE})) when $W$ is sampled from the minima
of $\mathcal{L}_{\mathrm{CE}}$. This can be accomplished by a well
known analytical technique \citep{franz1995recipes} which was developed
for the study of the energy landscape in disordered physical systems.
The computation is relatively involved, and here we report only the
final outcome. For the dedicated reader, all the details of the calculation,
which relies on the replica method and includes a double analytic
continuation, can be found in Appendix~\ref{sec:A-CE}. As reported
in figure~\ref{fig:FP}, we find that the typical minima of the CE
loss for small finite $\gamma$ are indeed surrounded by an exponential
number of zero error solutions. In other words, the CE focuses on
HLE regions. 

\begin{figure}
\begin{centering}
\includegraphics[width=0.6\columnwidth]{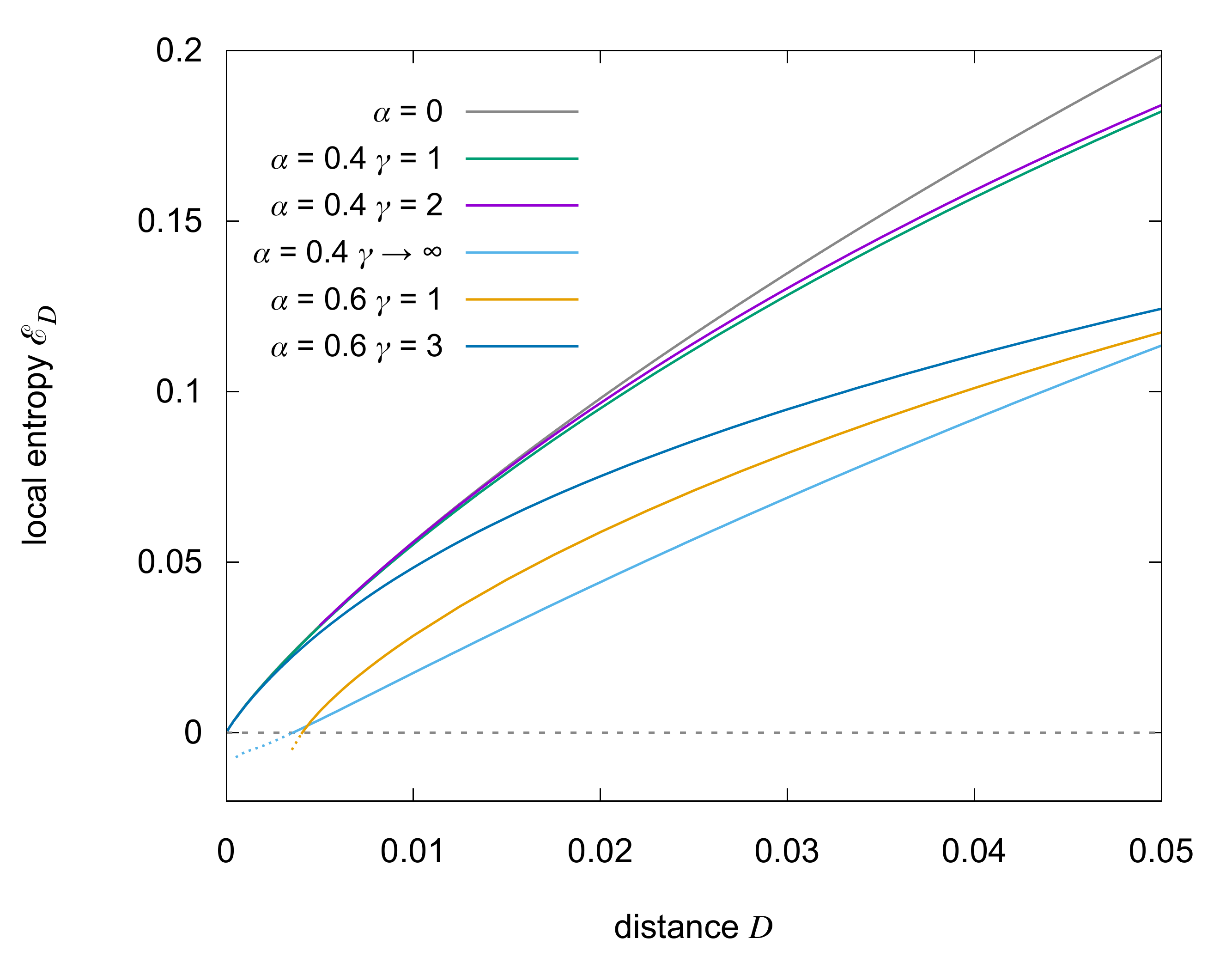}
\par\end{centering}
\caption{\label{fig:FP}Average local entropy around a typical minimum of the
$\mathcal{L}_{\mathrm{CE}}$ loss, for various values of $\alpha$
and $\gamma$. The gray upper curve, corresponding to $\alpha=0$,
is an upper bound since in that case all configurations are solutions.
For $\alpha=0.4$, the two curves with $\gamma=1$ and~$2$ nearly
saturate the upper bound at small distances, revealing the presence
of dense regions of solutions (HLE regions). There is a slight improvement
for $\gamma=2$, but the curve at $\gamma\to\infty$ shows that the
improvement cannot be monotonic: in that limit, the measure is dominated
by isolated solutions. This is reflected by the gap at small $d$
in which the entropy becomes negative, signifying the absence of solutions
in that range. For $\alpha=0.6$ we see that the curves are lower,
as expected. We also see that for $\gamma=1$ there is a gap at small
$d$, and that we need to get to $\gamma=3$ in order to find HLE
regions.}
\end{figure}

\begin{figure}
\begin{centering}
\includegraphics[width=0.6\columnwidth]{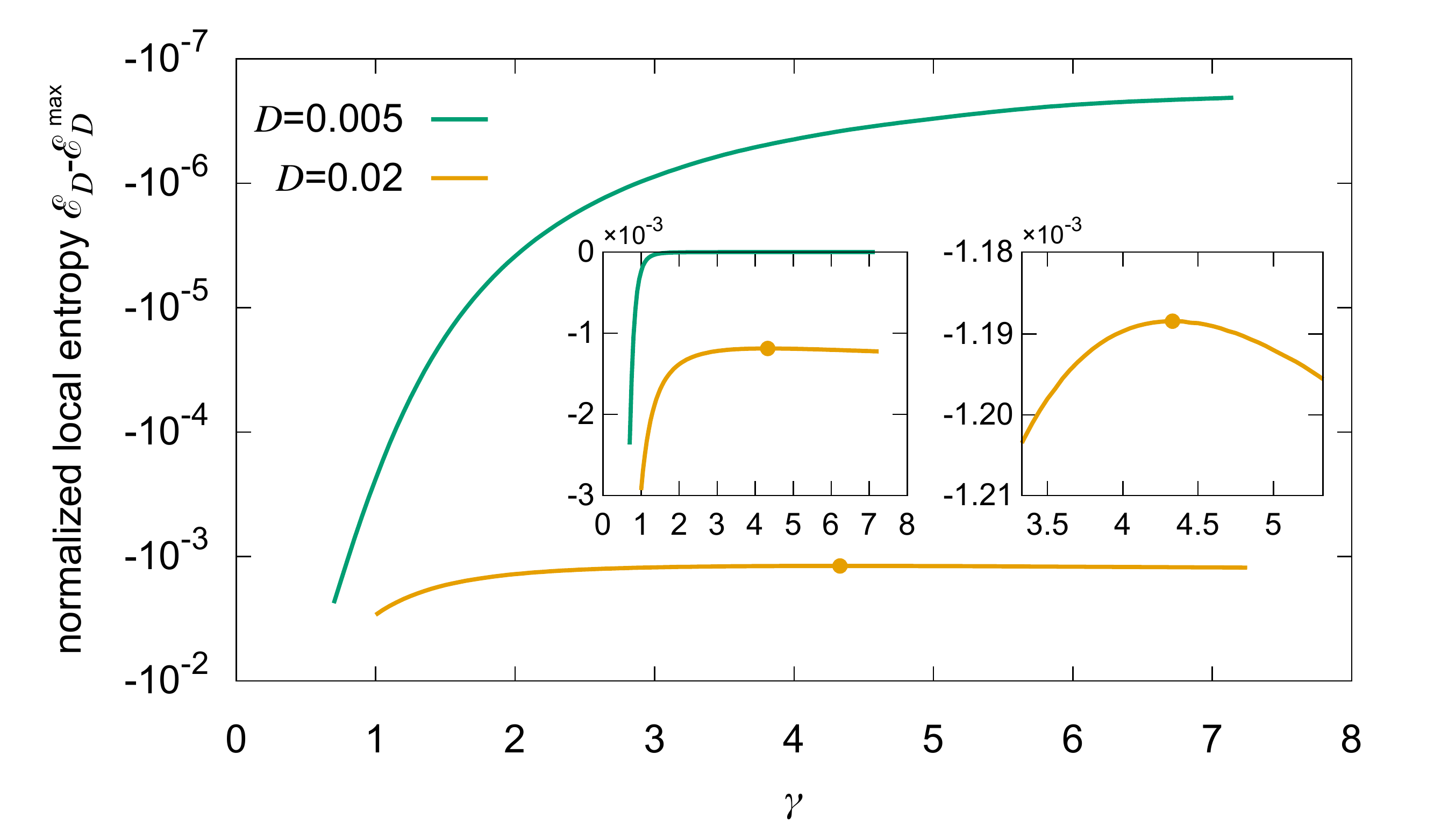}
\par\end{centering}
\caption{\label{fig:optimal_gamma}Normalized local entropy as a function of
$\gamma$, for a binary perceptron trained with the CE loss at $\alpha=0.4$
and two values of the distance $D$. The main plot shows the results
in logarithmic scale, the left inset is the same plot in linear scale.
For $D=0.005$ we could not find the optimal $\gamma$ due to numerical
issues; for $D=0.02$ the optimum is shown with a circle, although
the curve looks rather flat in an extended region: the right inset
shows an enlargement that makes the appearance of a maximum evident.}
\end{figure}

It is clear from the figure that $\gamma$ needs to be sufficiently
large for this phenomenon to occur. On the other hand, in the limit
$\gamma\to\infty$ the space of solutions is again dominated by the
isolated ones; we thus expect the existence of an optimal value of
$\gamma$, depending on the parameters $\alpha$ and $D$. We set
$\alpha=0.4$ and used two values of $D$, $0.005$ and $0.02$, and
measured the normalized local entropy $\mathcal{E}_{D}-\mathcal{E}_{D}^{\max}$
where $\mathcal{E}_{D}^{\max}=-D\log D-\left(1-D\right)\log\left(1-D\right)$
is the upper bound corresponding to the case $\alpha=0$ (gray curve
in fig.~\ref{fig:FP}). The results are shown in fig.~\ref{fig:optimal_gamma}.
Numerical issues prevented us from reaching the optimal $\gamma$
at $D=0.005$ (the main plot, in log scale, shows that the curve is
still growing), and the left inset (same data as the main plot, but
not in log scale) shows that there is a whole region where the local
entropy is extremely close to optimal (what we would call a dense
region). At a larger distance, $D=0.02$, we could find the optimum
(denoted with a dot) at a lower local entropy (consistently with fig.~\ref{fig:FP}),
but it is also clear that the ``good'' region of $\gamma$ is rather
large since the curve is very flat. In the limit of $\gamma\to\infty$
the curves would tend to $-0.054$ and $-0.028$ for the $D=0.02$
and $D=0.005$ cases, respectively.

The physical significance of these ``optimal $\gamma$ regions''
and their relation with the local geometry of the landscape is not
obvious. The main reason for this is that a fine description of the
local geometry of the HLE regions is still an open problem -- circumstantial
evidence from theoretical considerations and numerical simulations
suggests that they are rather complex structures \citep{baldassi_subdominant_2015},
possibly with a multifractal nature. At the present time, this is
pure speculation. On the practical side, however, a local search algorithm
that explores the landscape at low $\gamma$ and gradually increases
it should eventually reach this ``good $\gamma$'' plateau that
leads it toward a HLE region; further increasing $\gamma$ would then
be unlikely to drive it out, unless very strong drift terms are present
\citep{baldassi_unreasonable_2016}.

As an algorithmic check we have verified that while a Simulated Annealing
approach gets stuck at very high energies when trying to minimize
the error loss function, the very same algorithm with the CE loss
is indeed successful up to relatively high values of $\alpha$, with
just slightly worse performance compared to an analogous procedure
based on local entropy \citep{baldassi_local_2016}. In other words,
the CE loss on single-layer networks is a computationally cheap and
reasonably good proxy for the LE loss. These analytical results extend
straightforwardly to two layer networks with binary weights. The study
of continuous weight models can be performed resorting to the Belief
Propagation method.

\section{Belief Propagation and focusing Belief Propagation\label{sec:BP}}

Belief Propagation (BP), also known as sum-product, is an iterative
message-passing algorithm for statistical inference. When applied
to the problem of training a committee machine with a given set of
input-output patterns, it can be used to obtain, at convergence, useful
information on the probability distribution, over the weights of the
network, induced by the Gibbs measure. In particular, it allows to
compute the marginals of the weights as well as their entropy, which
in the zero-temperature regime is simply the logarithm of the volume
of the solutions, eq.~(\ref{eq:volume}), rescaled by the number
of variables $N$. The results are approximate, but (with high probability)
they approach the correct value in the limit of large $N$ in the
case of random uncorrelated inputs, at least in the replica-symmetric
phase of the space of the parameters. Due to the concentration property,
in this limit the macroscopic properties of any given problem (such
as the entropy) tend to converge to a common limiting case, and therefore
a limited amount of experiments with a few samples is sufficient to
describe very well the entire statistical ensemble.

We have used BP to study the case of the zero-temperature tree-like
committee machine with continuous weights and $K=3$. We have mostly
used $N=999$, which turns out to be large enough to produce results
in quite good agreement with the replica theory analysis. The implementation
can be made efficient by encoding each message with only two quantities
(see Appendix~\ref{sec:A-BP}). As mentioned above, this algorithm
works well in the replica-symmetric phase, which for our case means
when $\alpha\le\alpha_{0}\approx1.76$. Above this value, the (vanilla)
algorithm doesn't converge at all.

However, BP can be employed to perform additional analyses as well.
In particular, it can be modified rather straightforwardly to explore
and describe the region surrounding any given configuration, as it
allows to compute the local entropy (i.e.~the log-volume of the solutions)
for any given distance and any reference configuration (this is a
general technique, the details for our case are reported in Appendix~\ref{subsec:A-FranzParisi}).
The convergence issues are generally much less severe in this case.
Even in the RSB phase, if the reference configuration is a solution
in a wide minimum, the structure is locally replica-symmetric, and
therefore the algorithm converges and provides accurate results, at
least up to a value of the distance where other unconnected regions
of the solutions space come into consideration. In our tests, the
only other issue arose occasionally at very small distances, where
convergence is instead prevented by the loss of accuracy stemming
from finite size effects and limited numerical precision.

Additionally, the standard BP algorithm can be modified and transformed
into a (very effective) solver. There are several ways to do this,
most of which are purely heuristic. However, it was shown in \citet{baldassi_unreasonable_2016}
that adding a particular set of self-interactions to the weight variables
could approximately but effectively describe the replicated system
of eq.~(\ref{eq:part_func}): in other words, this technique can
be used to analyze the local-entropy landscape instead of the Gibbs
one. By using a sufficiently large number of replicas $y$ (we generally
used $y=10)$ and following an annealing protocol in the coupling
parameter $\lambda$ (starting from a low value and making it diverge)
this algorithm focuses on the maximally dense regions of solutions,
thus ending up in wide flat minima. For these reasons, the algorithm
was called ``focusing-BP'' (fBP). The implementation closely follows
that of \citet{baldassi_unreasonable_2016} (complete details are
provided in Appendix~\ref{subsec:A-fBP}). Our tests -- detailed
below -- show that this algorithm is the best solver (by a wide margin)
among the several alternatives that we tried in terms of robustness
of the minima found (and thus of generalization properties, as also
discussed below). Moreover, it also achieves the highest capacity,
nearly reaching the critical capacity where all solutions disappear.

\section{Entropic Least Action Learning}

Least Action Learning (LAL) \citep{mitchison1989bounds} is a heuristic
greedy algorithm that was designed to extend the well-known Perceptron
algorithm to the case of committee-machines with a single binary output
and sign activation functions. It takes one parameter, the learning
rate $\eta$. In its original version, patterns are presented randomly
one at a time, and at most one hidden unit is affected at a time.
In case of correct output, nothing is done, while in case of error
the hidden unit, among those with a wrong output, whose pre-activation
was closest to the threshold (and is thus the easiest to fix) is selected,
and the standard perceptron learning rule (with rate $\eta$) is applied
to it. In our tests we simply extended it to work in mini-batches,
to make it more directly comparable with stochastic-gradient-based
algorithms: for a given mini-batch, we first compute all the pre-activations
and the outputs for all patterns at the current value of the weights,
then we apply the LAL learning rule for each pattern in turn.

This algorithm proves to be surprisingly effective at finding minima
of the NE loss very quickly: in the random patterns case, its algorithmic
capacity is higher than gradient-based variants and almost as high
as fBP, and it requires comparatively few epochs. It is also computationally
very fast, owing to its simplicity. However, as we show in the sec.~\ref{sec:numerics},
it finds solutions that are much narrower compared to those of other
algorithms.

In order to drive LAL toward WFM regions, we add a local-entropy component
to it, by applying the technique described in \citet{baldassi_unreasonable_2016}
(see eq.~(\ref{eq:L_R})): we run $y$ replicas of the system in
parallel and we couple them with an elastic interaction. The resulting
algorithm, that we call entropic-LAL (eLAL) can be described as follows.
We initialize $y$ replicas randomly with weights $W^{a}$ and compute
their average $\tilde{W}$. We present mini-batches independently
to each replica, using different permutations of the dataset for each
of them. At each mini-batch, we apply the LAL learning rule. Then,
each replica is pushed toward the group average with some strength
proportional to a parameter $\lambda$: more precisely, we add a term
$\lambda\eta\left(\tilde{W}-W^{a}\right)$ to each of the weight vectors
$W^{a}$. After this update, we recompute the average $\tilde{W}$.
At each epoch, we increase the interaction strength $\lambda$. The
algorithm stops when the replicas have collapsed to a single configuration.

This simple scheme proves rather effective at enhancing the wideness
of the minima found while still being computationally efficient and
converging quickly, as we show in the sec.~\ref{sec:numerics}. We
show tests performed with $y=20$, but the entropic enhancement is
gradual, already providing improvements with very few replicas and
progressing at least up to $y=100$, which is the maximum value that
we have tested. Its main limitation is, of course, that it is tailored
to committee machines and it is unclear how to extend it to general
architectures. On the other hand, the effectiveness of the replication
scheme seems very promising from the point of view of improving the
quality of the minima found by greedy and efficient heuristics.

\section{\label{sec:numerics}Numerical studies}

We conclude our study by comparing numerically the curvature, the
wideness of the minima and the generalization error found by different
approaches. We consider two main scenarios: one, directly comparable
with the theoretical calculations, where a tree committee machine
with $K=9$ is trained over random binary patterns, and a second one,
which allows us to estimate the generalization capabilities, where
a fully-connected committee machine with $K=9$ is trained on a subset
of the Fashion-MNIST dataset \citep{fashionmnist}. The choice of
using $K=9$ instead of $3$ is intended to enhance the potential
robustness effect that the CE loss can have over NE on such architectures
(see the inset of fig.~\ref{fig:errors_vs_alpha}): for $K=3$, a
correctly classified pattern already requires $2$ out for $3$ units
to give the correct answer, and there is not much room for improvement
at the level of the pre-activation of the output unit. On the other
hand, since we study networks with a number of inputs of the order
of $10^{3}$, an even larger value of $K$ would either make $N/K$
too small in the tree-like case (exacerbating numerical issues for
the BP algorithms and straying too far from the theoretical analysis)
or make the computation of the Hessians too onerous for the fully-connected
case (each Hessian requiring the computation of $\left(NK\right)^{2}$
terms).

We compare several training algorithms with different settings (see
Appendix~\ref{subsec:A-Numerical}): stochastic GD with the CE loss
(ceSGD); least-action learning (LAL) and its entropic version (eLAL);
focusing BP (fBP). Of these, the non-gradient based ones (LAL, eLAL
and fBP) can be directly used with the sign activation functions (\ref{eq:sign_activation})
and the NE loss. On the other hand, ceSGD requires a smooth loss landscape,
therefore we used $\tanh$ activations, adding a gradually-diverging
parameter $\beta$ in their argument, since $\lim_{\beta\to\infty}\tanh\left(\beta x\right)=\mathrm{sign}\left(x\right)$.
The $\gamma$ parameter of the CE loss (\ref{eq:CE_loss}) was also
increased gradually. As in the theoretical computation, we also constrained
the weights of each hidden unit of the network to be normalized. The
NE loss with sign activations is invariant under renormalization of
each unit's weights, whereas the CE loss with $\tanh$ activations
is not. In the latter case, the parameters $\beta$ and $\gamma$
can be directly interpreted as the norm of the weights, since they
just multiply the pre-activations of the units. In a more standard
approach, the norm would be controlled by the initial choice of the
weights and be driven by the SGD algorithm automatically. In our tests
instead we have controlled these parameters explicitly, which allows
us to demonstrate the effect of different schedules. In particular,
we show (for both the random and the Fashion-MNIST scenarios) that
slowing down the growth of the norm with ceSGD makes a significant
difference in the quality of the minima that are reached. We do this
by using two separate settings for ceSGD, a ``fast'' and a ``slow''
one. In ceSGD-fast both $\beta$ and $\gamma$ are large from the
onset and grow quickly, whereas in ceSGD-slow they start from small
values and grow more slowly (requiring much more epochs for convergence).

In all cases -- for uniformity of comparison, simplicity, and consistency
with the theoretical analysis -- we consider scenarios in which the
training error (i.e.~the NE loss) gets to zero. This is, by definition,
the stopping condition for the LAL algorithm. We also used this as
a stopping criterion for ceSGD in the ``fast'' setting. For the
other algorithms, the stopping criterion was based on reaching a sufficiently
small loss (ceSGD in the ``slow'' setting), or the collapse of the
replicas (eLAL and fBP).

The analysis of the quality of the results was mainly based on the
study of the local loss landscape at the solutions. On one hand, we
computed the normalized local entropy using BP as described in a sec.~\ref{subsec:A-FranzParisi},
which provides a description of the NE landscape. On the other hand,
we also computed the spectrum of the eigenvalues of a smoothed-out
version of the NE loss, namely the mean square error loss computed
on networks with $\tanh$ activations. This loss depends on the parameters
$\beta$ of the activations: we set $\beta$ to be as small as possible
(maximizing the smoothing and thereby measuring features of the landscape
at a large scale) under the constraint that all the solutions under
consideration were still corresponding to zero error (to prevent degrading
the performance). For the Fashion-MNIST case, we also measured the
generalization error of each solution, and the robustness to input
noise.

\begin{figure}
\begin{centering}
\includegraphics[width=0.6\columnwidth]{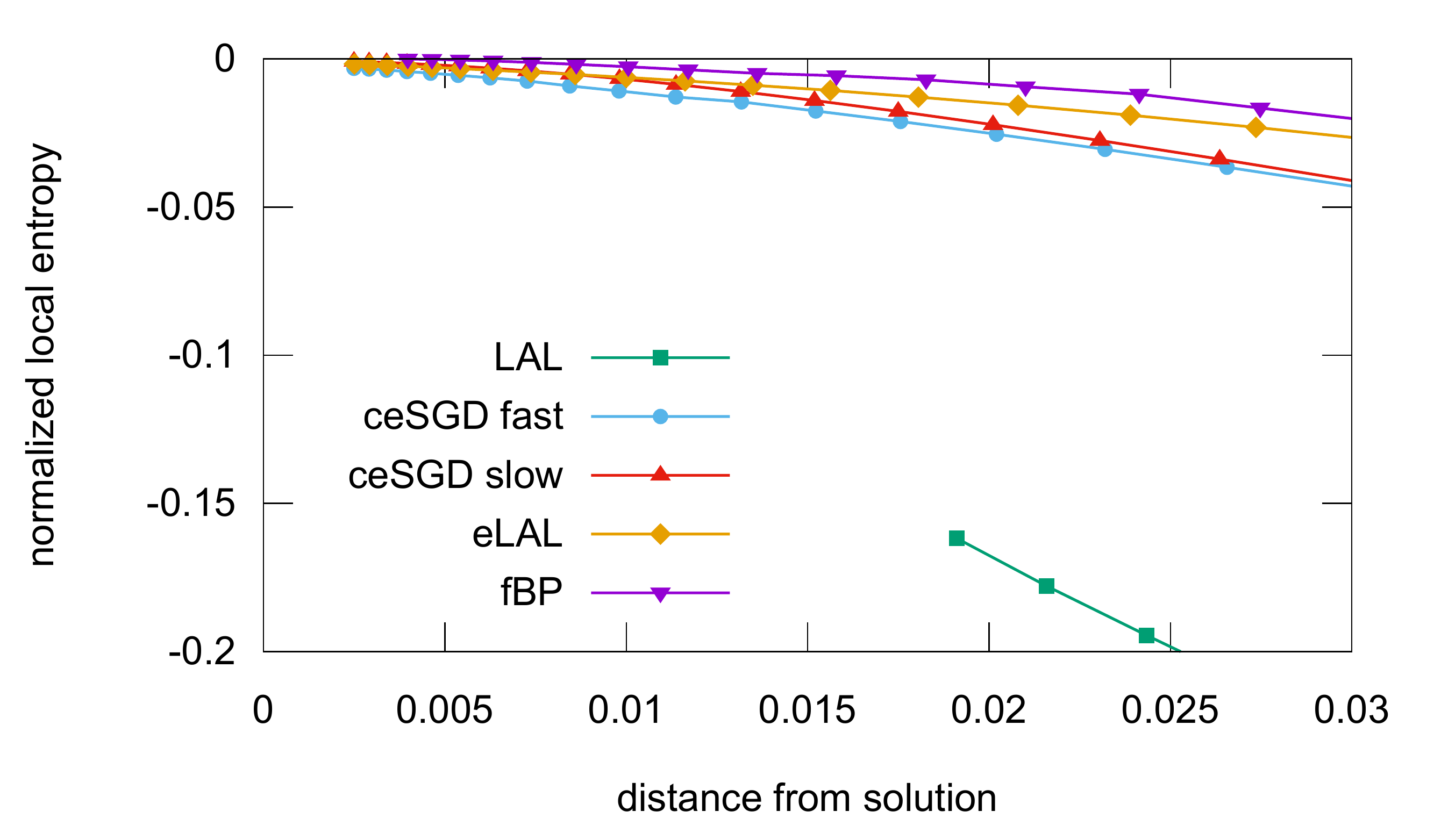}\caption{\label{fig:wef}Normalized local entropy as a function of the distance
from a reference solution, on a tree-like committee machine with $K=9$
and $N=999$, trained on $1000$ random patterns. The results were
obtained with the BP algorithm, by averaging over $10$ samples. Numerical
issues (mainly due to the approximations used) prevented BP from converging
at small distances for the LAL algorithm, and additionally they slightly
affect the results at very small distances. Qualitatively, though,
higher curves correspond to larger local entropies and thus wider
minima.}
\par\end{centering}
\end{figure}

\begin{figure}
\centering{}\includegraphics[width=0.6\columnwidth]{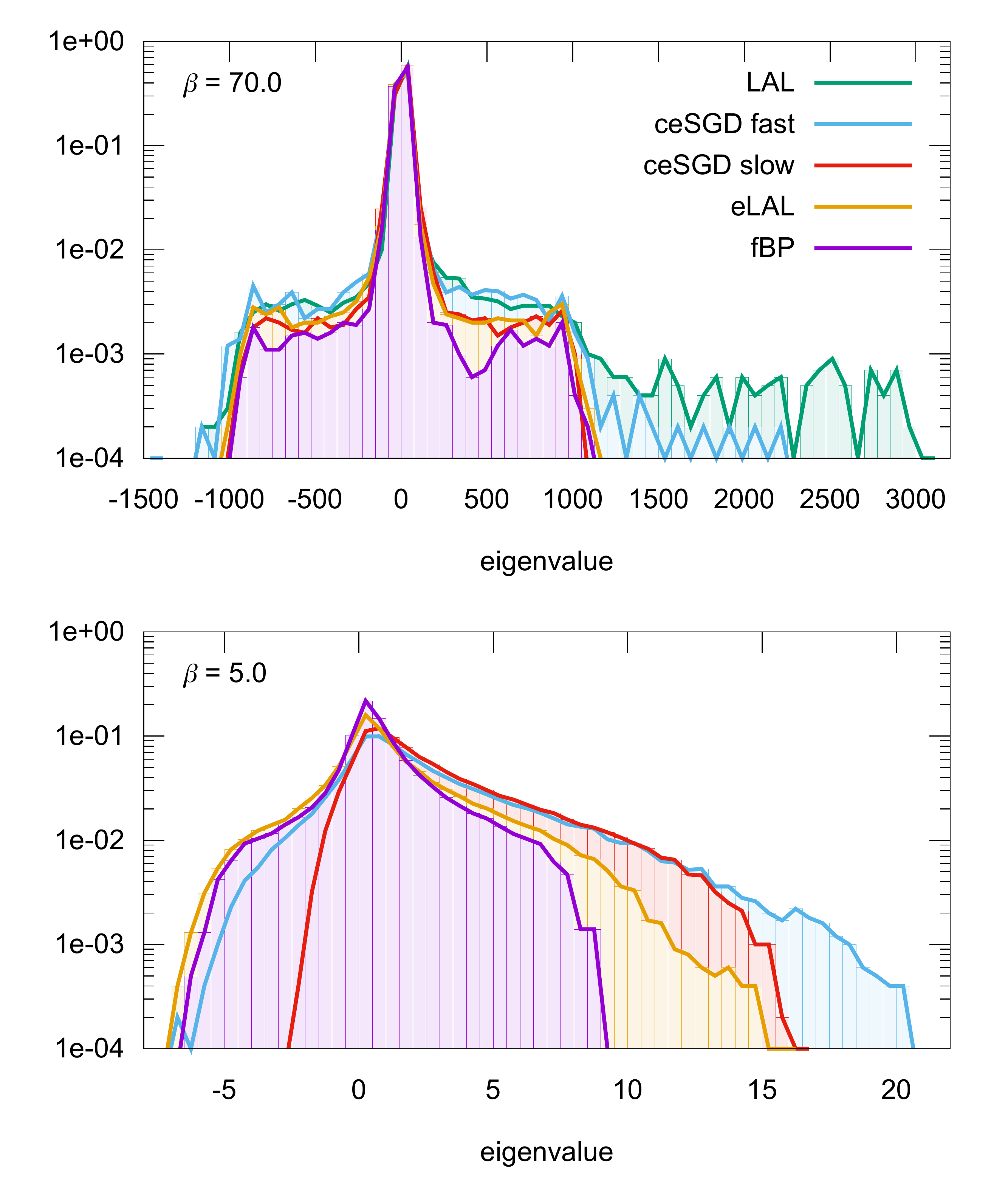}\caption{\label{fig:hess_rand}Spectra of the Hessian for the same solutions
of fig.~\ref{fig:wef}, for various algorithms. The spectra are directly
comparable since they are all computed on the same loss function (MSE;
using CE doesn't change the results qualitatively) and the networks
are normalized. The top panel shows the results with the parameter
$\beta$ of the activation functions set to a value such that all
solutions of all algorithms are still valid; this value is exclusively
determined by the LAL algorithm, and the bottom panel shows the results
for a much lower value of $\beta$ that can be used when removing
the LAL solutions, where differences between ceSGD-slow, eLAL and
fBP that were not visible al higher $\beta$ can emerge (the spectrum
of LAL would still be the widest by far even at this $\beta$).}
\end{figure}

In the random patterns scenario we set $N=999$ and $\alpha=1$, and
tested $10$ samples (the same for all the algorithms). The results
are presented in figs.~\ref{fig:wef} and \ref{fig:hess_rand}. The
two analyses allow to rank the algorithms (for the Hessians we can
use the maximum eigenvalue as a reasonable metric) and their results
are in agreement. As expected, fBP systematically finds very dense
regions of solutions, qualitatively compatible with the theoretical
analysis (cf. fig.~\ref{fig:wef} with fig.~\ref{fig:treeK3}) and
corresponding to the narrowest spectra of the Hessian at all $\beta$;
the other algorithms follow roughly in the order eLAL, ceSGD-slow,
ceSGD-fast, LAL. The latter is a very efficient solver for this model,
but it finds solutions in very narrow regions. On the other hand,
the same algorithm performed in parallel on a set of interacting replicas
is still efficient but much better at discovering WFMs. These results
are for $y=20$ replicas in eLAL, but our tests show that $y=10$
would be sufficient to match ceSGD-slow and that $y=100$ would further
improve the results and get closer to fBP. Overall, the results of
the random pattern case confirm the existence of WFMs in continuous
networks, and suggest that a (properly normalized) Hessian spectrum
can be used as a proxy for detecting whether an algorithm has found
a WFM region.

\begin{figure}
\begin{centering}
\includegraphics[width=0.6\columnwidth]{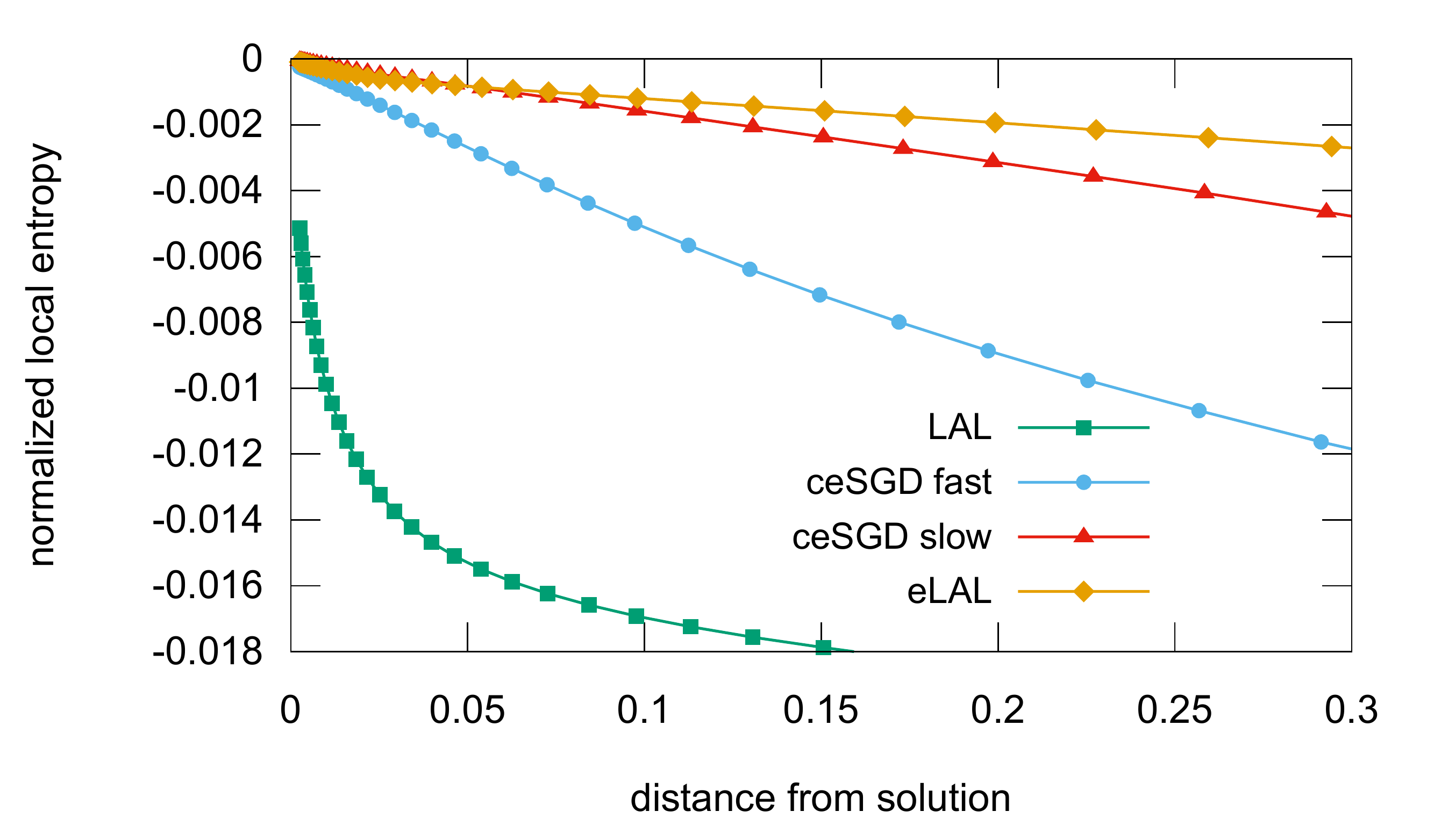}
\par\end{centering}
\centering{}\includegraphics[width=0.6\columnwidth]{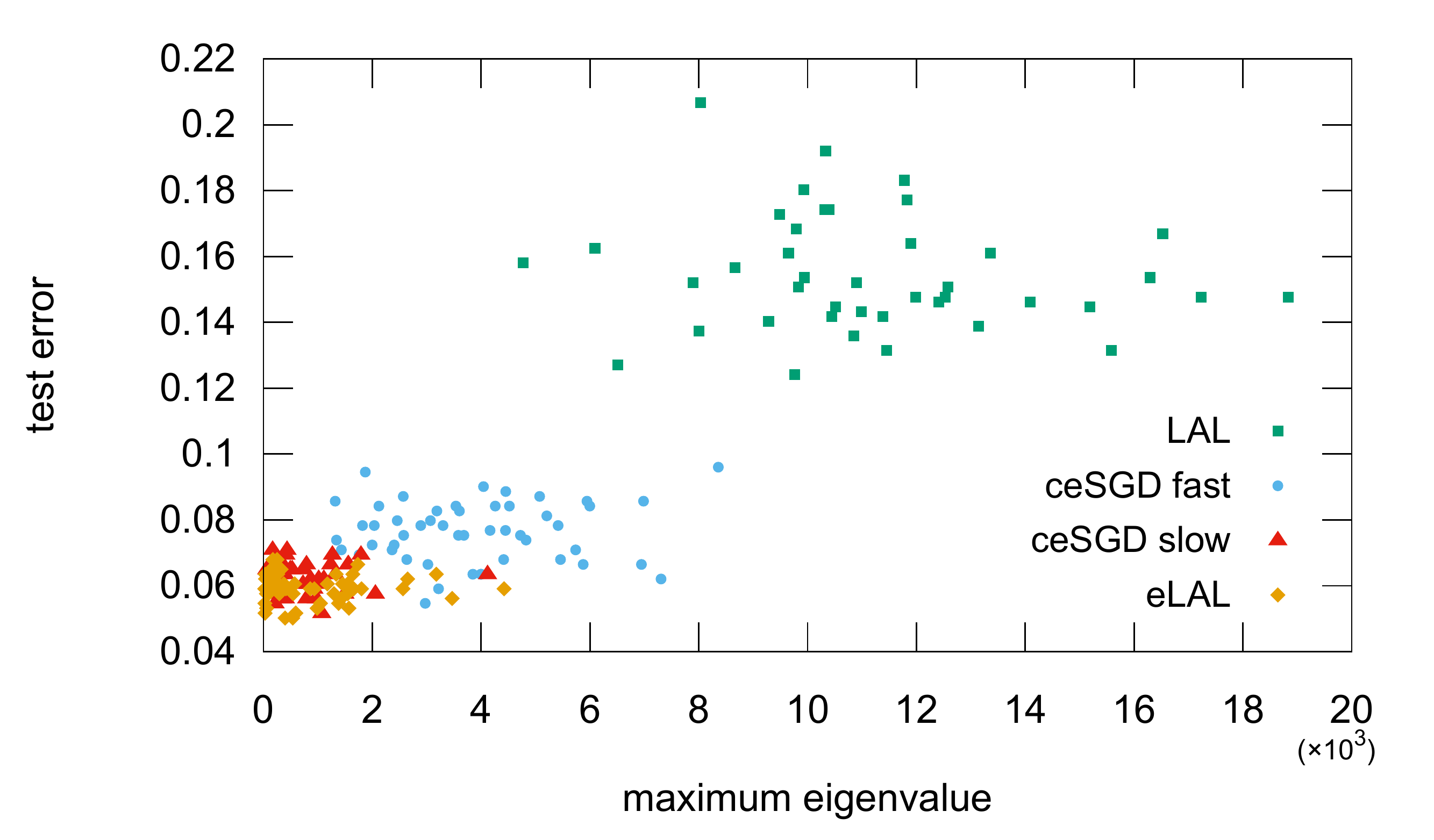}\caption{\label{fig:fashion}Experiments on a subset of Fashion-MNIST. Top:
average normalized local entropies; Bottom: test error vs maximum
eigenvalue of the Hessian spectrum at $\beta=90$. The maximum eigenvalue
is correlated to the generalization (WFMs tend to generalize better),
and the quality of the minima varies between algorithms.}
\end{figure}

We then studied the performance of ceSGD (fast and slow settings),
LAL and eLAL on a small fully-connected network which learns to discriminate
between two classes of the Fashion-MNIST dataset (we chose the classes
Dress and Coat, which are rather challenging to tell apart but also
sufficiently different to offer the opportunity to generalize even
with a small simple network trained on very few examples). We trained
our networks on a small subset of the available examples ($500$ patterns;
binarized to $\pm1$ by using the median of each image as a threshold
on the original grayscale inputs; we filtered both the training and
test sets to only use images in which the median was between $0.25$
and $0.75$ as to avoid too-bright or too-dark images and make the
data more uniform and more challenging). This setting is rather close
to the one which we could study analytically, except for the patterns
statistics and the use of fully-connected rather than tree-like layers,
and it is small enough to permit computing the full spectrum of the
Hessian. On the other hand, it poses a difficult task in terms of
inference (even though finding solutions with zero training error
is not hard), which allowed us to compare the results of the analysis
of the loss landscape with the generalization capabilities on the
test set. Each algorithm was ran $50$ times. The results are shown
in fig.~\ref{fig:fashion}, and they are analogous to those for the
random patterns case, but in this setting we can also observe that
indeed WFMs tend to generalize better. Also, while we could not run
fBP on this data due to the correlations present in the inputs and
to numerical problems related to the fully-connected architecture,
which hamper convergence, it is still the case that ceSGD can find
WFMs if the norms are controlled and increased slowly enough, and
that we can significantly improve the (very quick and greedy) LAL
algorithm by replicating it, i.e.~by effectively adding a local-entropic
component.

We also performed an additional batch of tests on a randomized version
of the Fashion-MNIST dataset, in which the inputs were reshuffled
across samples on a pixel-by-pixel basis (such that each sample only
retained each individual pixel bias while the correlations across
pixels were lost). This allowed us to bridge the two scenarios and
directly compare the local volumes in the presence or absence of features
of the data that can lead to proper generalization. We kept the settings
of each algorithm as close as possible to those for the Fashion-MNIST
tests. Qualitatively, the results were quite similar to the ones on
the original data except for a slight degradation of the performance
of eLAL compared to ceSGD. Quantitatively, we observed that the randomized
version was more challenging and generally resulted in slightly smaller
volumes. Additional measures comparing the robustness to the presence
of noise in the input (which measures overfitting and thus can be
conceived as being a precursor of generalization) confirm the general
picture. The detailed procedures and results are reported in Appendix~\ref{sec:A-rand-FMNIST}.

\section*{Conclusions and future directions}

In this paper, we have generalized the local entropy theory to continuous
weights and we have shown that WFM exists in non convex neural systems.
We have also shown that the CE loss spontaneously focuses on WFM.
On the algorithmic side we have derived and designed novel algorithmic
schemes, either greedy (very fast) or message passing, which are driven
by the local entropy measure. Moreover, we have shown numerically
that ceSGD can be made to converge in WFM by an appropriate cooling
procedure of the parameter which controls the norm of the weights.
Our findings are in agreement with recent results showing that ReLU
transfer functions also help the learning dynamics to focus on WFM
\citep{baldassi2019properties}. Future work will be aimed at extending
our methods to multiple layers, trying to reach a unified framework
for current DNN models. This is a quite challenging task which has
the potential to reveal the role that WFM play for generalization
in different data regimes and how that can be connected to the many
layer architectures of DNN.
\begin{acknowledgments}
CB and RZ acknowledge ONR Grant N00014-17-1-2569. We thank Leon Bottou
for discussions.
\end{acknowledgments}

\appendix
\makeatletter
\renewcommand{\thefigure}{A\@arabic\c@figure}
\makeatletter
\setcounter{figure}{0}  

\section{\label{sec:A-HLE-RSB}High Local Entropy states from the 1-RSB formalism}

Given a system described by a vector of discrete variables $W$ with
an associated energy function $E\left(W\right)$, the Boltzmann equilibrium
distribution at inverse temperature $\beta$ reads 
\begin{equation}
P\left(W;\beta\right)=\frac{1}{Z\left(\beta\right)}e^{-\beta E\left(W\right)}
\end{equation}
where the normalization factor $Z$ is given by the partition function
\begin{equation}
Z\left(\beta\right)=\sum_{W}e^{-\beta E\left(W\right)}
\end{equation}

In the limit $\beta\to\infty$, the distribution is just a flat measure
over the ground states of the energy; we can denote the ground state
energy as $E^{\star}=\min_{W}E\left(W\right)$ and the characteristic
function over the ground states as
\begin{equation}
\mathbb{X}\left(W\right)=\begin{cases}
1 & \mathrm{if}\;E\left(W\right)=E^{\star}\\
0 & \mathrm{otherwise}
\end{cases}
\end{equation}
such that $Z\left(\infty\right)=\sum_{W}\mathbb{X}\left(W\right)$
and $\log Z\left(\infty\right)$ gives the entropy of the ground states.

In \citet{baldassi_subdominant_2015}, we introduced a large-deviation
measure with a modified energy function in which each configuration
is reweighted by a ``local entropy'' term. There, we only considered
the $\beta\to\infty$ limit and defined the local entropy as the number
of ground states at a certain normalized distance $D$ from a reference
configuration $\tilde{W}$:
\begin{equation}
\mathcal{S}\left(\tilde{W},D\right)=\log\mathcal{N}\left(\tilde{W},D\right)=\log\sum_{W}\mathbb{X}\left(W\right)\delta\left(d\left(W,\tilde{W}\right)-ND\right)\label{eq:local_entropy}
\end{equation}
where $d\left(\cdot,\cdot\right)$ is a suitably defined distance
function and $\delta\left(\cdot\right)$ is the Kronecker delta. With
this definition, we can define a modified partition function as follows:
\begin{equation}
Z\left(\infty,y,D\right)=\lim_{\beta\to\infty}\sum_{\tilde{W}}e^{-\beta E\left(\tilde{W}\right)+y\mathcal{S}\left(\tilde{W},D\right)}\label{eq:Z_large_dev_tilde}
\end{equation}
which (up to irrelevant constant factors) coincides with:
\begin{equation}
Z\left(\infty,y,D\right)=\sum_{\tilde{W}}\mathbb{X}\left(\tilde{W}\right)\mathcal{N}\left(\tilde{W},D\right)^{y}
\end{equation}
This approach can be shown to be strictly related to the 1-step replica-symmetry-broken
(1RSB) formalism of the bare energetic problem. First, let us define
a local free entropy at any given inverse temperature $\beta^{\prime}$
(i.e.~a generalization of eq.~(\ref{eq:local_entropy})), and use
a soft-constraint on the distance through a Lagrange multiplier $\lambda$:
\begin{equation}
\phi\left(\tilde{W},\lambda,\beta^{\prime}\right)=\log\sum_{W}e^{-\beta^{\prime}E\left(W\right)-\frac{\lambda}{2}d\left(W,\tilde{W}\right)}
\end{equation}
Note that the use of a Lagrange multiplier is mostly convenient in
order to make the relation with the 1RSB description evident. It is
only equivalent to using a hard constraint for the distance in the
thermodynamic limit, and depending on the convexity properties of
the function $\phi$, but we will generally ignore these issues for
the time being and come back to them later.

We can then rewrite the large-deviation partition function eq.~(\ref{eq:Z_large_dev_tilde})
in this more general case as:
\begin{equation}
Z\left(\beta,\beta^{\prime},y,\lambda\right)=\sum_{\tilde{W}}e^{-\beta E\left(\tilde{W}\right)+y\phi\left(\tilde{W},\lambda,\beta^{\prime}\right)}\label{eq:Z_nontraced}
\end{equation}

Let us now consider the case in which $y\in\mathbb{N}$: this allows
us, by simple algebraic manipulations, to rewrite the partition function
introducing a sum over all the configurations of $y$ replicas of
the system:
\begin{equation}
Z\left(\beta,\beta^{\prime},y,\lambda\right)=\sum_{\tilde{W},\left\{ W^{a}\right\} }e^{-\beta E\left(\tilde{W}\right)-\beta^{\prime}\sum_{a=1}^{y}E\left(W^{a}\right)-\frac{\lambda}{2}\sum_{a=1}^{y}d\left(W^{a},\tilde{W}\right)}
\end{equation}

This partition function describes a system of $y+1$ interacting \emph{real}
replicas with an interaction which is mediated by the reference configuration
$\tilde{W}$. However we can isolate the sum over the configurations
of $\tilde{W}$ to obtain a system of $y$ interacting real replicas.
In the special case $\beta=0$ we obtain:
\begin{equation}
Z\left(0,\beta^{\prime},y,\lambda\right)=\sum_{\tilde{W},\left\{ W^{a}\right\} }e^{-\beta^{\prime}\sum_{a=1}^{y}E\left(W^{a}\right)+\log\sum_{\tilde{W}}\exp\left(-\frac{\lambda}{2}\sum_{a=1}^{y}d\left(W^{a},\tilde{W}\right)\right)}\label{eq:Z_traced}
\end{equation}

We have stressed the fact that the replicas are real to avoid the
confusion with the \emph{virtual} replicas used for the ``replica
trick'': here, we are not replicating the system virtually in order
to compute a free entropy in the limit of zero replicas: instead,
we are describing a system of $y$ identical interacting objects.
The general case of $y\in\mathbb{R}$ can be obtained by analytic
continuation once an expression for all integer $y$ is found.

This description is highly reminiscent of -- in fact, almost identical
to -- the derivation of the ergodicity-breaking scheme used in \citet{monasson1995structural}:
there, an auxiliary symmetry breaking field is introduced (having
the same role of $\tilde{W}$ in our notation); then, a free energy
expression is introduced in which the role of the energy is taken
by a ``local free entropy'' (the analogous of eq.~(\ref{eq:local_entropy})
for general $\beta$), after which the system is replicated $y$ times
and the auxiliary field $\tilde{W}$ is traced out, leading to a system
of $y$ real replicas with an effective pairwise interaction. Finally,
the limit of vanishing interaction ($\lambda\to0^{+}$) is taken in
order to derive the equilibrium description. When this system is studied
in the replica-symmetric (RS) Ansatz, it results in the 1RSB description
of the original system, with $y$ having the role of the Parisi parameter
(usually denoted by $m$). Indeed, in this limit of vanishing interaction
and for $\beta=0$, our equations reduce to the 1RSB case as well.

Therefore, apart from minor differences, the main point of discrepancy
between that analysis and our approach is that we don't restrict ourselves
to the equilibrium distribution. Instead, we explore the whole range
of values of $\lambda$. In this context, we also have no reason to
restrict ourselves to the range $y\in\left[0,1\right]$, as it is
usually done in order to give a physical interpretation to the 1RSB
solution; to the contrary, we are (mostly) interested in the limit
of large $y$, in which only the configurations of maximal local entropy
are described.

The relationship between our analysis and the usual 1RSB case can
be made even more direct, leading to an alternative -- although with
very similar results -- large deviations analysis: consider, instead
of eq.~(\ref{eq:Z_traced}), a partition function in which the interaction
among the replicas is pairwise (without the reference configuration
$\tilde{W}$) and the constraint on the distance is hard (introduced
via a Dirac delta function):
\begin{equation}
Z_{1RSB}\left(\beta^{\prime},y,D\right)=\sum_{\left\{ W^{a}\right\} }e^{-\beta^{\prime}\sum_{a=1}^{y}E\left(W^{a}\right)}\prod_{a>b}\delta\left(d\left(W^{a},W^{b}\right)-ND\right)\label{eq:Z_1RSB}
\end{equation}

Suppose then that we study the average free entropy $\left\langle \log Z_{1RSB}\left(\beta^{\prime},y,\gamma\right)\right\rangle $
(where $\left\langle \cdot\right\rangle $ represents the average
over the quenched parameters, if any) in the context of replica theory.
Then, we will have $n$ virtual replicas of the whole system, and
since each system has $y$ real replicas we end up with $ny$ total
replicas. Let's use indices $c$, $d$ for the virtual replicas and
$a$, $b$ for the real ones, such that a configuration will now have
two indices, e.g.~$W^{ca}$. Suppose that we manage to manipulate
the expression such that it becomes a function, among other order
parameters, of the overlaps $q^{ca,db}=\frac{1}{N}\left\langle W^{ca},W^{db}\right\rangle $,
where $\left\langle \cdot,\cdot\right\rangle $ represents some inner
product, and that the distance function $d\left(\cdot,\cdot\right)$
can be expressed in terms of those. Then, as usual, we would introduce
auxiliary integrals
\begin{equation}
\int\prod_{\left(ca,db\right)}\left(N\mathrm{d}q^{ca,db}\right)\prod_{\left(ca,db\right)}\delta\left(Nq^{ca,db}-\left\langle W^{ca},W^{db}\right\rangle \right)
\end{equation}
Using this, we can rewrite the interaction term. Say that $d\left(W,W^{\prime}\right)=\left\langle W,W\right\rangle +\left\langle W^{\prime},W^{\prime}\right\rangle -2\left\langle W,W^{\prime}\right\rangle $,
then:
\begin{equation}
\prod_{c}\prod_{a>b}\delta\left(d\left(W^{ca},W^{cb}\right)-ND\right)=\prod_{c}\prod_{a>b}\delta\left(N\left(q^{ca,ca}+q^{cb,cb}-2q^{ca,cb}-D\right)\right)\label{eq:interaction_as_overlaps}
\end{equation}
By assuming replica symmetry, we seek a saddle point with this structure:
\begin{eqnarray}
q^{ca,ca} & = & Q\nonumber \\
q^{ca,cb} & = & q_{1}\quad\left(a\ne b\right)\\
q^{ca,db} & = & q_{0}\quad\left(c\ne d\right)\nonumber 
\end{eqnarray}
with $Q\ge q_{1}\ge q_{0}$. The interaction term eq.~(\ref{eq:interaction_as_overlaps})
becomes:
\begin{equation}
\prod_{c}\prod_{a>b}\delta\left(N\left(q^{ca,ca}+q^{cb,cb}-2q^{ca,cb}-D\right)\right)=\delta\left(2N\left(Q-q_{1}-D\right)\right)
\end{equation}

Therefore, the external parameter $D$ eliminates a degree of freedom
in the solution to the saddle point equations for the overlaps. The
final step in the replica calculation would have the form
\begin{eqnarray}
\left\langle \log Z_{1RSB}\left(\beta^{\prime},y,D\right)\right\rangle  & = & \phi_{1RSB}\left(\beta^{\prime},y,Q,q_{1},q_{0},\dots\right)\delta\left(Q-q_{1}-D\right)\nonumber \\
 & = & \phi_{1RSB}\left(\beta^{\prime},y,Q,Q-D,q_{0},\dots\right)
\end{eqnarray}
where $\phi_{1RSB}$ is the expression which would have been derived
in an equilibrium computation without the interaction term, the dots
in the argument represent extra order parameters, and the order parameters
are fixed by the saddle point equations
\begin{eqnarray}
\partial_{Q}\phi_{1RSB}\left(\beta^{\prime},y,Q,Q-D,q_{0},\dots\right) & = & 0\nonumber \\
\partial_{q_{0}}\phi_{1RSB}\left(\beta^{\prime},y,Q,Q-D,q_{0},\dots\right) & = & 0\\
\vdots\nonumber 
\end{eqnarray}

Thus, the difference with respect to the usual 1RSB computation is
that the equation for finding the extremum over $q_{1}$ is removed,
and the one for finding the extremum over $Q$ is modified. Maximizing
over $D$, by solving for $\partial_{D}\phi=0$, is then equivalent
to the usual 1RSB description (equivalent to the case $\lambda\to0$
in the soft-constraint case):
\begin{equation}
Z_{1RSB}\left(\beta^{\prime},y\right)=\max_{D}Z_{1RSB}\left(\beta^{\prime},y,D\right)
\end{equation}

In the common case where $Q$ is fixed (e.g.~if the variables $W$
are discrete, or constraints on the norm are introduced) then this
representation fixes $q_{1}$; it is clear then that our large deviations
analysis (the alternative one of eq.~(\ref{eq:Z_1RSB})) is simply
derived by fixing $q_{1}$ as an external parameter, and thus omitting
the saddle point equation $\partial_{q_{1}}\phi_{1RSB}=0$. Note that
this wouldn't make physical sense in the standard derivation of the
1RSB equations, since in that context $q_{1}$ is only introduced
as an overlap between virtual replicas when choosing an Ansatz for
the solutions of the saddle point equations; our derivation is only
physically meaningful when describing a system of real interacting
replicas or, in the case of the original derivation from eq.~(\ref{eq:Z_large_dev_tilde}),
a system with a modified energy function.

\section{\label{sec:A-CE}Cross-Entropy minima, errors and high local entropy
configurations }

\subsection{Cross-Entropy loss ground states}

In order to study analytically the properties of the minima of the
CE loss function in the case of i.i.d. random patterns, the key obstacle
is to compute the normalization factor of the Gibbs measure, the partition
function $Z$. Once this is done one has access to the the Gibbs measure
which concentrates on the minima of the loss in the $\beta\to\infty$
limit.

$Z$ is an exponentially fluctuating random variable and in order
to find its most probable values we need to average its logarithm,
a complicated task which we perform by the replica method. Once this
is done, the typical value of $Z$ can be recovered by $Z_{\text{typ}}\simeq\exp\left(N\left\langle \log Z\right\rangle _{\xi}\right)$,
where $\left\langle \cdot\right\rangle _{\xi}$ stands for the average
over the random patterns. 

We refer to \citet{engel-vandenbroek} for a thorough review of the
replica method. Here we just remind the reader that the replica method
is an analytic continuation technique which allows in some cases (mean-field
models) to compute the expectation of the logarithm of the partition
function from the knowledge of its integer moments. The starting point
is the following small $n$ expansion

\[
Z^{n}=1+n\log Z+O\left(n^{2}\right)
\]
This identity may be averaged over the random patterns and gives the
average of the $\log$ from the averaged $n$-th power of the partition
function

\[
\left\langle \log Z\right\rangle _{\xi}=\min_{n\to0}\frac{\left\langle Z^{n}\right\rangle _{\xi}-1}{n}
\]
The idea of the replica method is to restrict to integer $n$ and
to take the analytic continuation $n\to0$

\[
\left\langle Z^{n}\right\rangle _{\xi}=\prod_{a=1}^{n}\left\langle Z_{a}\right\rangle _{\xi}=\sum_{\left\{ W^{1},\dots,W^{n}\right\} }\left\langle e^{-\beta\sum_{a=1}^{n}E\left(W^{a}\right)}\right\rangle _{\xi}
\]
We have $n$ replicas of the initial model. The random patterns in
the expression of the energy disappear once the average has been carried
out. Eventually one computes the partition function of an effective
system of $nN$ variables with a non random energy function resulting
from the average. The result may be written formally as 

\[
\left\langle Z^{n}\right\rangle _{\xi}=\exp\left(NF\left(n\right)\right)
\]
where $F$ is the expression resulting from the sum over all configurations.
Once the small $n$ limit is taken, the final expression can be estimated
analytically by means of the saddle-point method given that $N$ is
assumed to be large.

In the case of our problem we have

\[
Z=\sum_{\left\{ w_{i}=\pm1\right\} }\exp\left(-\beta\sum_{\mu=1}^{M}\;f\left(\frac{1}{\sqrt{N}}\sum_{i=1}^{N}w_{i}\xi_{i}^{\mu}\right)\right)
\]
Following the replica approach, we need to compute

\[
\left\langle Z^{n}\right\rangle =\left\langle \int\prod_{i,a}\mathrm{d}\mu\left(w_{i}^{a}\right)\prod_{\mu,a}\exp\left(-\beta f\left(\frac{1}{\sqrt{N}}\sum_{i=1}^{N}w_{i}^{a}\xi_{i}^{\mu}\right)\right)\right\rangle _{\xi}
\]
where the integration measure is just over the binary values of the
weights. By enforcing $x^{\mu}=\frac{1}{\sqrt{N}}\sum_{i=1}^{N}w_{i}\xi_{i}^{\mu}$
through a delta function, we can linearize the dependence on the randomness
of the patterns and perform the average as follows:

\begin{align*}
\left\langle Z^{n}\right\rangle  & =\left\langle \int\prod_{i,a}\mathrm{d}\mu\left(w_{i}^{a}\right)\int\prod_{a,\mu}\frac{\mathrm{d}x^{a\mu}\mathrm{d}\hat{x}^{a\mu}}{2\pi}\prod_{a,\mu}\exp\left(-\beta f\left(x^{a\mu}\right)\right)\prod_{a,\mu}\exp\left(i\hat{x}^{a\mu}x^{a\mu}-i\hat{x}^{a\mu}\sum_{i=1}^{N}\frac{w_{i}^{a}\xi_{i}^{\mu}}{\sqrt{N}}\right)\right\rangle _{\xi}=\\
= & \int\prod_{i,a}\mathrm{d}\mu\left(w_{i}^{a}\right)\int\prod_{a,\mu}\frac{\mathrm{d}x^{a\mu}\mathrm{d}\hat{x}^{a\mu}}{2\pi}\prod_{a\mu}\exp\left(-\beta f\left(x^{a\mu}\right)\right)\prod_{a,\mu}\exp\left(i\hat{x}^{a\mu}x^{a\mu}\right)\exp\left(-\frac{1}{2N}\sum_{ab}\hat{x}^{a\mu}\hat{x}^{b\mu}\sum_{i}\frac{w_{i}^{a}w_{i}^{b}}{N}\right)=\\
= & \int\prod_{i,a}\mathrm{d}\mu\left(w_{i}^{a}\right)\int\prod_{a,\mu}\frac{\mathrm{d}x^{a\mu}\mathrm{d}\hat{x}^{a\mu}}{2\pi}\int\prod_{a>b}\frac{\mathrm{d}q^{ab}\mathrm{d}\hat{q}^{ab}}{2\pi}\prod_{a>b}e^{-Nq^{ab}\hat{q}^{ab}}\prod_{a>b}e^{\hat{q}^{ab}\sum_{i}w_{i}^{a}w_{i}^{b}}\prod_{a\mu}\exp\left(-\beta f\left(x^{a\mu}\right)\right)\times\\
\; & \times\prod_{a,\mu}\exp\left(i\hat{x}^{a\mu}x^{a\mu}\right)\exp\left(-\frac{1}{2}\sum_{ab}\hat{x}^{a\mu}\hat{x}^{b\mu}q^{ab}\right)=\\
= & \int\prod_{a>b}\frac{\mathrm{d}q^{ab}\mathrm{d}\hat{q}^{ab}N}{2\pi}\prod_{a>b}e^{-Nq^{ab}\hat{q}^{ab}}\left(\int\prod_{i,a}\mathrm{d}\mu\left(w_{i}^{a}\right)\prod_{a>b}e^{\hat{q}^{ab}\sum_{i}w_{i}^{a}w_{i}^{b}}\right)\times\\
\; & \times\left(\int\prod_{a}\frac{\mathrm{d}x^{a}\mathrm{d}\hat{x}^{a}}{2\pi}\prod_{a}\exp\left(-\beta\;f\left(x^{a}\right)\right)\prod_{a}\exp\left(i\hat{x}^{a}x^{a}\right)\exp\left(-\frac{1}{2}\sum_{ab}\hat{x}^{a}\hat{x}^{b}q^{ab}\right)\right)^{\alpha N}
\end{align*}
where we have used the delta functions to introduce the order parameters
$q^{ab}$ and $\hat{q}^{ab}$. In order to write the multiple integrals
in a form which can be evaluated by saddle point, we restrict to the
replica symmetric assumption $q^{ab}=q$ and $\hat{q}^{qb}=\hat{q}$
, and perform few simplifications.

First we sum over the weights:

\begin{align*}
\int\prod_{i,a}\mathrm{d}\mu\left(w_{i}^{a}\right)\prod_{a>b}e^{\hat{q}\sum_{i}w_{i}^{a}w_{i}^{b}} & =\int\prod_{i,a}\mathrm{d}\mu\left(w_{i}^{a}\right)\prod_{i}e^{\frac{\hat{q}}{2}\left(\sum_{a}w_{i}^{a}\right)^{2}}e^{-\frac{\hat{q}}{2}Nn}=\\
=\, & e^{-\frac{\hat{q}}{2}Nn}\left(\sum_{w=\pm1}e^{\frac{\hat{q}}{2}\left(\sum_{a}w^{a}\right)^{2}}\right)^{N}=e^{-\frac{\hat{q}}{2}Nn}\left(\int Du\sum_{w=\pm1}e^{\sqrt{\hat{q}}u\sum_{a}w^{a}}\right)^{N}=\\
\simeq & \exp Nn\left[-\frac{\hat{q}}{2}+\int Du\log\left(2\cosh\sqrt{\hat{q}}u\right)\right]
\end{align*}

Second, we simplify the terms which are raised to the power $\alpha N$:

\begin{eqnarray*}
 &  & \int\prod_{a}\frac{\mathrm{d}x^{a}\mathrm{d}\hat{x}^{a}}{2\pi}\prod_{a}\exp\left(-\beta\;f\left(x^{a}\right)\right)\prod_{a}\exp\left(i\hat{x}^{a}x^{a}\right)\exp\left(-\frac{1}{2}\sum_{ab}\hat{x}^{a}\hat{x}^{b}q\right)=\\
 &  & \quad=\int\prod_{a}\frac{\mathrm{d}x^{a}\mathrm{d}\hat{x}^{a}}{2\pi}\prod_{a}\exp\left(-\beta\;f\left(x^{a}\right)\right)\prod_{a}\exp\left(i\hat{x}^{a}x^{a}\right)\exp\left(-\frac{1}{2}\left(1-q\right)\sum_{a}\left(\hat{x}^{a}\right)^{2}-\frac{q}{2}\left(\sum_{a}\hat{x}^{a}\right)^{2}\right)\\
 &  & \quad=\int Du\,\left(\int\frac{\mathrm{d}x\mathrm{d}\hat{x}}{2\pi}e^{-\beta\;f(x)}e^{i\hat{x}x}\exp\left(-\frac{1}{2}\left(1-q\right)\hat{x}^{2}+iu\sqrt{q}\hat{x}\right)\right)^{n}\\
 &  & \quad=\int Du\left(\int\frac{\mathrm{d}x}{\sqrt{2\pi}}e^{-\beta\;f(x)}\frac{\exp\left(-\frac{(x+u\sqrt{q})^{2}}{2(1-q)}\right)}{\sqrt{1-q}}\right)^{n}
\end{eqnarray*}
Finally we can write the saddle point expression for the replicated
partition function:

\[
\left\langle Z^{n}\right\rangle \simeq\exp\left[Nn\left(\frac{q\hat{q}}{2}-\frac{\hat{q}}{2}+\int Du\log\left(2\cosh\sqrt{\hat{q}}u\right)+\alpha\int Du\log\int\exp\left(-\beta f\left(x\sqrt{1-q}+u\sqrt{q}\right)\right)\right)\right]=\exp\left(NnG\right)
\]
where it is useful to write the action $G$ as the sum of three terms 

\[
G=\frac{\hat{q}}{2}\left(q-1\right)+G_{S}+\alpha G_{E}
\]
The entropic contribution $G_{S}$ reads

\[
G_{s}=\int Du\log\left(2\cosh\sqrt{\hat{q}}u\right)
\]
and the energetic one $G_{E}$

\[
G_{E}=\int Du\log\int\exp\left(-\beta f\left(x\sqrt{1-q}+u\sqrt{q}\right)\right)
\]
The replicated partition function can then be computed in the limit
$N\to\infty$ by solving the saddle point equations $\frac{\partial G}{\partial\hat{q}}=0$
and $\frac{\partial G}{\partial q}=0$. The derivatives of $G_{S}$and
$G_{E}$ read

\[
\frac{\partial G_{S}}{\partial\hat{q}}=\int Du\frac{\int Dx\:e^{-\beta f\left(x\sqrt{1-q)}+u\sqrt{q}\right)}\left[-\frac{\beta}{2}f^{\prime}\left(x\sqrt{1-q)}+u\sqrt{q}\right)\left(\frac{u}{\sqrt{q}}-\frac{x}{\sqrt{1-q}}\right)\right]}{\int Dx\:e^{-\beta f\left(x\sqrt{1-q}+u\sqrt{q}\right)}}
\]

and 

\[
\frac{\partial G_{E}}{\partial q}=\int Du\frac{u}{2\sqrt{\hat{q}}}\tanh\left(u\sqrt{\hat{q}}\right)=\frac{1}{2}\left(1-\int Du\tanh^{2}\left(u\sqrt{\hat{q}}\right)\right)
\]
Setting to zero these derivatives we get the saddle point equations
for $q$ and $\hat{q}$

\begin{align*}
q & =\int Du\tanh^{2}\left(u\sqrt{\hat{q}}\right)\\
\hat{q} & =-\alpha\beta\int Du\frac{\int Dx\:e^{-\beta f\left(x\sqrt{1-q}+u\sqrt{q}\right)}\left[-\frac{\beta}{2}f^{\prime}\left(x\sqrt{1-q}+u\sqrt{q}\right)\left(\frac{u}{\sqrt{q}}-\frac{x}{\sqrt{1-q}}\right)\right]}{\int Dx\:e^{-\beta f\left(x\sqrt{1-q}+u\sqrt{q}\right)}}=\\
\; & =-\frac{\alpha}{\sqrt{1-q}}\int Du\frac{\int Dx\:e^{-\beta f\left(x\sqrt{1-q}+u\sqrt{q}\right)}\left[\frac{1}{\sqrt{1-q}}+x\left(\frac{u}{\sqrt{q}}-\frac{x}{\sqrt{1-q}}\right)\right]}{\int Dx\:e^{-\beta f\left(x\sqrt{1-q}+u\sqrt{q}\right)}}
\end{align*}
In the limit of large $\beta$ we need to rescale the order parameters
to obtain finite quantities. By setting $q=1-\frac{\delta q}{\beta}$,
we find for the last equation

\[
\hat{q}=\frac{\alpha\beta^{2}}{\delta q}\int Du\;\left[\text{argmax}_{x}\left(-\frac{x^{2}}{2}-\log\left(1+\exp\left(-2\gamma\left(x\sqrt{\delta q}+u\right)\right)\right)\right)\right]^{2}
\]
Once the saddle point equations are solved numerically, we can compute
the minimum energy (minimum loss) and the entropy at low temperature.
We have:

\[
E=-\frac{\partial G}{\partial\beta}=-\alpha\frac{\partial G_{E}}{\partial\beta}=\alpha\int Du\frac{\int Dx\:e^{-\beta f\left(x\sqrt{1-q}+u\sqrt{q}\right)}f\left(x\sqrt{1-q}+u\sqrt{q}\right)}{\int Dx\:e^{-\beta f\left(x\sqrt{1-q}+u\sqrt{q}\right)}}
\]
In the limit of large $\beta$, with $q=1-\frac{\delta q}{\beta}$,
we find

\[
E=\alpha\int Du\,f\left(x^{*}\left(u\right)\right)
\]
where

\[
x^{*}\left(u\right)\equiv\text{argmax}_{x}\left(-\frac{x^{2}}{2}-\log\left(1+\exp\left(-2\gamma\left(x\sqrt{\delta q}+u\right)\right)\right)\right)
\]
We can compute the entropy using the relation $S=G+\beta E$ . 

In figure~\ref{fig:errors_vs_alpha} we show the behavior of the
energy vs the loading $\alpha$. As one may observe, up to relatively
large values of $\alpha$ the energy is extremely small, virtually
equal zero for any accessible size $N.$

Having established that by minimizing the the cross-entropy one ends
up in regions of perfect classification where the error loss function
is zero, it remains to be understood which type of configurations
of weights are found. Does the CE converge to a typical zero energy
configuration of the error function (i.e.~an isolated point-like
solution in the weight space) or does it converge to the rare regions
of high local entropy?

The answer to this question is that the CE does in fact focus on the
HLE subspaces. 

\subsection{The local entropy around CE ground states}

In order to show this analytically we need to be able to count how
many zero error configurations exist in a region close to a typical
minima of the CE loss. Following \citet{franz1995recipes}, this computation
can be done by averaging with the CE Gibbs measure the entropy of
the Number of Errors loss function. 

Let's call $f$ and $g$ the cross-entropy and error loss functions
per pattern, respectively. We need to evaluate the most probable value
of 

\[
\log Z_{FP}=\log\frac{\int\prod_{i}\mathrm{d}\mu\left(w_{i}\right)\prod_{\mu}\exp\left(-\beta f\left(\sum_{i}\frac{w_{i}\xi_{i}^{\mu}}{\sqrt{N}}\right)\right)\log\left[\int\prod_{i}\mathrm{d}\mu\left(v_{i}\right)\prod_{\mu}\exp\left(-\beta^{\prime}g\left(\sum_{i}\frac{v_{i}\xi_{i}^{\mu}}{\sqrt{N}}\right)\right)\delta\left(pN-\sum_{i}w_{i}v_{i}\right)\right]}{\int\prod_{i}\mathrm{d}\mu\left(w_{i}\right)\prod_{\mu}\exp\left(-\beta f\left(\sum_{i}\frac{w_{i}\xi_{i}^{\mu}}{\sqrt{N}}\right)\right)}
\]
which can be computed by replica approach as we have done for $Z$,
yielding the local entropy $\mathcal{E}_{D}=\left\langle \log Z_{FP}\right\rangle $.
In the above expression $pN$ is the constrained overlap between the
different minima, which is trivially related to the Hamming distance
$DN$ by $D=\frac{1}{2}\left(1-p\right)$. We will need to perform
twice the replica trick, one to extract the most probable value of
$\log Z_{FP}$ (index $n)$ and one to linearize the $\log$ inside
the integral (index $r$), 

\begin{align*}
 & \left\langle Z_{FP}^{n,r}\right\rangle =\\
 & \left\langle \int\prod_{ai}\mathrm{d}\mu\left(w_{i}^{a}\right)\prod_{\mu a}\exp\left(-\beta f\left(\sum_{i}\frac{w_{i}^{a}\xi_{i}^{\mu}}{\sqrt{N}}\right)\right)\times\right.\\
\; & \left.\times\frac{1}{r}\left(\int\prod_{ci}\mathrm{d}\mu\left(v_{i}\right)\prod_{\mu c}\exp\left(-\beta^{\prime}g\left(\sum_{i}\frac{v_{i}^{c}\xi_{i}^{\mu}}{\sqrt{N}}\right)\right)\delta\left(pN-\sum_{i}w_{i}^{a=1}v_{i}^{c}\right)-1\right)\right\rangle _{\xi}
\end{align*}
This quantity is computed for $n,r$ integer ($a,b=1\dots,n$ and
$c,d,=1,\dots,r$) and eventually the analytic continuation $n,r\to0$
is taken. We will do this under the replica symmetric (RS) assumption
which for small distances \textbf{$D$ }is expected to by exact. For
the sake of completeness, we report hereafter all the main steps of
the calculation.

We need to compute

\begin{eqnarray*}
\left\langle Z_{FP}^{n,r}\right\rangle  & = & \int\prod_{ai}\mathrm{d}\mu\left(w_{i}^{a}\right)\int\prod_{a\mu}\frac{\mathrm{d}x^{a\mu}\mathrm{d}\hat{x}^{a\mu}}{2\pi}\prod_{\mu a}\exp\left(-\beta f\left(x^{a\mu}\right)\right)\prod_{a,\mu}e^{i\hat{x}^{a\mu}x^{a\mu}}\times\\
 &  & \times\int\prod_{ci}\mathrm{d}\mu\left(v_{i}^{c}\right)\int\prod_{c\mu}\frac{\mathrm{d}y^{c\mu}\mathrm{d}\hat{y}^{c\mu}}{2\pi}\prod\exp\left(-\beta^{\prime}g\left(y^{c\mu}\right)\right)\times\\
 &  & \times\prod_{\mu}\left\langle e^{i\hat{x}^{a\mu}\sum_{ia}\frac{w_{i}^{a}\xi_{i}^{\mu}}{\sqrt{N}}+i\hat{y}^{c\mu}\sum_{ic}\frac{v_{i}^{c}\xi_{i}^{\mu}}{\sqrt{N}}}\right\rangle _{\xi}\prod_{c}\delta\left(pN-\sum_{i}w_{i}^{a=1}v_{i}^{c}\right)
\end{eqnarray*}
The average over the patterns is factorized and can be easily performed.
Upon expanding the results for large $N$, the term in the brackets
reads

\begin{align*}
\exp & \left(-\frac{1}{2N}\sum_{i}\left(\sum_{a}w_{i}^{a}\hat{x}^{a\mu}+\sum_{c}v_{i}^{c}\hat{y}^{c\mu}\right)^{2}\right)=\\
= & \exp\left(-\frac{1}{2}\sum_{ab}\hat{x}^{a\mu}\hat{x}^{b\mu}\frac{1}{N}\sum_{i}w_{i}^{a}w_{i}^{b}-\frac{1}{2}\sum_{cd}\hat{x}^{c\mu}\hat{x}^{d\mu}\frac{1}{N}\sum_{i}v_{i}^{c}v_{i}^{d}-\sum_{ac}\hat{x}^{a\mu}\hat{y}^{c\mu}\frac{1}{N}\sum_{i}w_{i}^{a}v_{i}^{c}\right)
\end{align*}
Introducing the order parameters corresponding to the different overlaps
we find for the total expression 

\begin{eqnarray*}
\left\langle Z_{FP}^{n,r}\right\rangle  & = & \int\prod_{a>b}\frac{\mathrm{d}q^{ab}\mathrm{d}\hat{q}^{ab}N}{2\pi}\int\prod_{c>d}\frac{\mathrm{d}s^{cd}\mathrm{d}\hat{s}^{cd}N}{2\pi}\prod_{a>b}e^{-N\hat{q}^{ab}q^{ab}}\prod_{c>d}e^{-N\hat{s}^{cd}s^{cd}}\int\prod_{c}\frac{\mathrm{d}\hat{p}^{c}N}{2\pi}\prod_{c}e^{-N\hat{p}^{c}p}\times\\
 &  & \times\int\prod_{a>1,c}\frac{\mathrm{d}t^{ac}\mathrm{d}\hat{t}^{ac}}{2\pi}N\prod_{a>1,c}e^{-N\hat{t}^{ac}t^{ac}}\times\\
 &  & \times\int\prod_{ai}\mathrm{d}\mu\left(w_{i}^{a}\right)\int\prod_{ci}\mathrm{d}\mu\left(v_{i}^{c}\right)\prod_{a>b}e^{\hat{q}^{ab}\sum_{i}w_{i}^{a}w_{i}^{b}}\prod_{c>d}e^{\hat{s}^{cd}\sum_{i}v_{i}^{c}v_{i}^{d}}\prod_{c}e^{\hat{p}^{c}\sum_{i}w_{i}^{a=1}v_{i}^{c}}\prod_{a>1,c}e^{\hat{t}^{ac}\sum_{i}w_{i}^{a}v_{i}^{c}}\times\\
 &  & \times\left(\int\prod_{a}\frac{\mathrm{d}x^{a}\mathrm{d}\hat{x}^{a}}{2\pi}\int\prod_{c}\frac{\mathrm{d}y^{c}\mathrm{d}\hat{y}^{c}}{2\pi}\prod_{a}e^{-\beta f\left(x^{a}\right)}\prod_{c}e^{-\beta^{\prime}g\left(x^{c}\right)}\prod_{a}e^{ix^{a}\hat{x}^{a}}\times\right.\\
 &  & \left.\times\prod_{c}e^{iy^{c}\hat{y}^{c}}e^{-\frac{1}{2}\sum_{ab}\hat{x}^{a}\hat{x}^{b}q^{ab}-\frac{1}{2}\sum_{cd}\hat{y}^{c}\hat{y}^{d}s^{cd}-\sum_{c}\hat{x}^{1}\hat{y}^{c}p-\sum_{a>1,c}\hat{x}^{a}\hat{y}^{c}t^{ac}}\right)^{\alpha N}
\end{eqnarray*}
In order to proceed, we search the solutions of the saddle point equations
in the RS subspace, $q^{ab}=q$, $\hat{q}^{ab}=\hat{q}$, $s^{ab}=s$,
etc. The various factors can be simplified as follows:

\begin{align*}
\prod_{i}e^{\sum_{a>b}\hat{q}^{ab}w_{i}^{a}w_{i}^{b}} & =\prod_{i}e^{\frac{\hat{q}}{2}\left[\left(\sum_{a}w_{i}^{a}\right)^{2}-\sum_{a}\left(w_{i}^{a}\right)^{2}\right]}=e^{-\frac{Nn\hat{q}}{2}}\prod_{i}e^{\frac{\hat{q}}{2}\left(\sum_{a}w_{i}^{a}\right)^{2}},\\
\prod_{i}e^{\sum_{c>d}\hat{s}^{ab}v_{i}^{c}v_{i}^{d}} & =e^{-\frac{Nr\hat{s}}{2}}\prod_{i}e^{\frac{\hat{s}}{2}\left(\sum_{c}v_{i}^{a}\right)^{2}},\\
\prod_{i}e^{\sum_{c}\hat{p}^{c}w_{i}^{a=1}v_{i}^{c}} & =\prod_{i}e^{\hat{p}\sum_{c}w_{i}^{a=1}v_{i}^{c}},\\
\prod_{i}e^{\sum_{a>1,c}\hat{t}^{ac}w_{i}^{a}v_{i}^{d}} & =\prod_{i}e^{\hat{t}\left(\sum_{a>1}w_{i}^{a}\right)\left(\sum_{c}v_{i}^{c}\right)}=\prod_{i}e^{\hat{t}\left(\sum_{a}w_{i}^{a}\right)\left(\sum_{c}v_{i}^{c}\right)-\hat{t}w_{i}^{a=1}\left(\sum_{c}v_{i}^{c}\right)},
\end{align*}
and 
\begin{align*}
-\sum_{a>b}\hat{q}^{ab}q^{ab} & =-\frac{n\left(n-1\right)}{2}\hat{q}q\simeq\frac{n}{2}\hat{q}q,\\
-\sum_{c>d}\hat{s}^{cd}s^{cd} & =-\frac{r\left(r-1\right)}{2}\hat{s}s\simeq\frac{r}{2}\hat{s}s,\\
-\sum_{c}\hat{p}^{c}p & =-\hat{rpp},\\
-\sum_{a>1,c}\hat{t}^{ac}t^{ac} & =-\left(n-1\right)r\hat{t}t\simeq r\hat{t}t.
\end{align*}

A series of further simplifications are needed in order to write the
$\left\langle Z_{FP}^{n,r}\right\rangle $ in the appropriate saddle
point form. The terms containing the integrals over $w$ and $v$
become factorized 

\begin{eqnarray*}
 &  & \left(\int\prod_{a}\mathrm{d}\mu\left(w^{a}\right)\int\prod_{c}\mathrm{d}\mu\left(v^{c}\right)\,e^{-\frac{n}{2}\hat{q}-\frac{r}{2}\hat{s}}\int Dz\,e^{z\sqrt{\hat{q}-\hat{t}}\sum_{a}w^{a}}\int Du\,e^{u\sqrt{\hat{s}-\hat{t}}\sum_{c}v^{c}}\int Dx\,e^{x\sqrt{\hat{t}}\left(\sum_{a}w^{a}+\sum_{c}v^{c}\right)}\right.\times\\
 &  & \left.\times e^{\left(\hat{p}-\hat{t}\right)w^{a=1}\sum_{c}v^{c}}\right)^{N}=\\
 &  & =\left(\int DzDuDx\left(\int\mathrm{d}\mu\left(w\right)e^{-\frac{\hat{q}}{2}+w\left(z\sqrt{\hat{q}-\hat{t}}+x\sqrt{\hat{t}}\right)}\right)^{n-1}\times\right.\\
 &  & \left.\times\int\mathrm{d}\mu\left(w^{a=1}\right)e^{-\frac{\hat{q}}{2}+w^{a=1}\left(z\sqrt{\hat{q}-\hat{t}}+x\sqrt{\hat{t}}\right)}\left(\int\mathrm{d}\mu\left(v\right)\,e^{\left(u\sqrt{\hat{s}-\hat{t}}+x\sqrt{\hat{t}}\right)v+\left(\hat{p}-\hat{t}\right)w^{a=1}v}\right)^{r}\right)^{N}
\end{eqnarray*}
where we have kept the notation $w^{a=1}$ just for the sake of clarity.
Being an integration variable we now drop it. By summing over $w$
and v and with some straightforward change of variables we get

\begin{align*}
\int & DzDuDx\frac{\sum_{w=\pm1}e^{w\left(z\sqrt{\hat{q}-\hat{t}}+x\sqrt{\hat{t}}\right)}\left(-\frac{\hat{s}}{2}+\log2\cosh\left(\left(u\sqrt{\hat{s}-\hat{t}}+x\sqrt{\hat{t}}\right)v+\left(\hat{p}-\hat{t}\right)w\right)\right)}{\sum_{w=\pm1}e^{w\left(z\sqrt{\hat{q}-\hat{t}}+x\sqrt{\hat{t}}\right)}}\\
= & -\frac{\hat{s}}{2}+\int Dz\frac{\sum_{w=\pm1}e^{wz\sqrt{\hat{q}}}\left(\int DuDx\log\cosh\left(u\sqrt{\hat{s}-\hat{t}}+\sqrt{\hat{t}}\left(\sqrt{\frac{\hat{t}}{\hat{q}}}z+\sqrt{\frac{\hat{q}-\hat{t}}{\hat{q}}}x\right)+\left(\hat{p}-\hat{t}\right)w\right)\right)}{\sum_{w=\pm1}e^{wz\sqrt{\hat{q}}}}=\\
= & -\frac{\hat{s}}{2}+\int Dz\frac{\sum_{w=\pm1}e^{wz\sqrt{\hat{q}}}\int D\phi\log\cosh\left(\phi\sqrt{\frac{(\hat{s}-\hat{t})\hat{q}+(\hat{q}-\hat{t})\hat{t}}{\hat{q}}}+\frac{\hat{t}}{\sqrt{\hat{q}}}z+\left(\hat{p}-\hat{t}\right)w\right)}{2\cosh z\sqrt{\hat{q}}}=
\end{align*}
For the integral containing the dependence on $f$ and $g$ we find
similar simplifications.

\begin{align*}
\int & DzDh\left(\int\frac{\mathrm{d}x\mathrm{d}\hat{x}}{2\pi}\,e^{-\beta f(x)-\frac{1-q}{2}\hat{x}^{2}+i\hat{x}\left(z\sqrt{q-t}+h\sqrt{t}+x\right)}\right)^{n-1}\times\\
\times & \left(\int\prod_{c}\frac{\mathrm{d}y^{c}\mathrm{d}\hat{y}^{c}}{2\pi}\int\frac{dxd\hat{x}}{2\pi}\,e^{-\beta f(x)-\frac{1-q}{2}\hat{x}^{2}+i\hat{x}\left(z\sqrt{q-t}+h\sqrt{t}+x+i\left(p-t\right)\sum_{c}\hat{y}^{c}\right)}\times\right.\\
\; & \left.\times\prod_{c}e^{-\beta^{\prime}g\left(y^{c}\right)-\frac{1-s}{2}\sum_{c}\left(\hat{y}^{c}\right)^{2}-\frac{s-t}{2}\left(\sum_{c}\hat{y}^{c}\right)^{2}+i\sum_{c}\hat{y}^{c}\left(h\sqrt{t}+y^{c}\right)}\right)=\\
= & \int DzDh\frac{1}{Z_{0}}\int\prod_{c}\frac{\mathrm{d}y^{c}\mathrm{d}\hat{y}^{c}}{2\pi}\int\frac{\mathrm{d}x}{\sqrt{2\pi}}\frac{e^{-\beta f\left(x\right)}}{\sqrt{1-q}}e^{-\frac{1}{2(1-q)}\left[\left(z\sqrt{q-t}+h\sqrt{t}+x\right)^{2}-\left(p-t\right)^{2}\left(\sum_{c}\hat{y}^{c}\right)^{2}+2i\left(z\sqrt{q-t}+h\sqrt{t}+x\right)\left(p-t\right)\sum_{c}\hat{y}^{c}\right]}\times\\
\; & \times\prod_{c}e^{-\beta^{\prime}g\left(y^{c}\right)-\frac{1-s}{2}\sum_{c}\left(\hat{y}^{c}\right)^{2}-\frac{s-t}{2}\left(\sum_{c}\hat{y}^{c}\right)^{2}+i\sum_{c}\hat{y}^{c}\left(h\sqrt{t}+y^{c}\right)}=\dots
\end{align*}
where we have used the notation

\[
Z_{0}\equiv\int\frac{\mathrm{d}x}{\sqrt{2\pi}}\frac{e^{-\beta f\left(x\right)}}{\sqrt{1-q}}e^{-\frac{1}{2\left(1-q\right)}\left(z\sqrt{q-t}+h\sqrt{t}+x\right)^{2}}
\]

Continuing the computation, we can linearize the terms in $\hat{y}^{c}$
with an auxiliary integral, factor the terms with the index $c$,
take the $r\to0$ limit and perform explicitly two integrals: 
\begin{align*}
\dots= & \int DzDh\frac{1}{Z_{0}}\int Du\int\frac{\mathrm{d}x}{\sqrt{2\pi}}\frac{e^{-\beta f\left(x\right)}}{\sqrt{1-q}}e^{-\frac{1}{2(1-q)}\left(z\sqrt{q-t}+h\sqrt{t}+x\right)^{2}}\times\\
\times & \left(\int\frac{\mathrm{d}y\mathrm{d}\hat{y}}{2\pi}e^{-\beta^{\prime}g\left(y\right)-\frac{1-s}{2}\hat{y}^{2}+i\hat{y}\left(\sqrt{s-t-\frac{\left(p-t\right)^{2}}{1-q}}u+h\sqrt{t}+y-\frac{\left(z\sqrt{q-t}+h\sqrt{t}+x\right)\left(p-t\right)}{1-q}\right)}\right)^{r}=\\
\\
= & \int DzDh\frac{1}{Z_{0}}\int Du\int\frac{dx}{\sqrt{2\pi}}\frac{e^{-\beta f\left(x\right)}}{\sqrt{1-q}}e^{-\frac{1}{2\left\langle 1-q\right\rangle }\left(z\sqrt{q}+x\right)^{2}}\times\\
\times & \;\log\int\frac{\mathrm{d}y}{\sqrt{2\pi}}\frac{e^{-\beta^{\prime}g\left(y\right)}}{\sqrt{1-s}}e^{-\frac{1}{\sqrt{2\left(1-s\right)}}\left[u\sqrt{s-t-\frac{\left(p-t\right)^{2}}{1-q}}+\sqrt{t}\left(z\sqrt{\frac{t}{q}}+h\sqrt{\frac{q-t}{q}}\right)+y-\frac{z\sqrt{q}(p-t)}{1-q}\right]^{2}}=\\
\\
= & \int Dz\frac{1}{Z_{0}^{\prime}}\int\frac{\mathrm{d}x}{\sqrt{2\pi}}\frac{e^{-\beta f\left(x\right)}}{\sqrt{1-q}}e^{-\frac{1}{2\left(1-q\right)}\left(z\sqrt{q}+x\right)^{2}}\times\\
\times & \int Du\,\log\int\frac{\mathrm{d}y}{\sqrt{2\pi}}\frac{e^{-\beta^{\prime}g\left(y\right)}}{\sqrt{1-s}}e^{-\frac{1}{\sqrt{2\left(1-s\right)}}\left[u\sqrt{s-t-\frac{\left(p-t\right)^{2}}{1-q}-\frac{t\left(q-t\right)}{q}}+\frac{t}{\sqrt{q}}z-\frac{z\sqrt{q}\left(p-t\right)}{1-q}+y\right]^{2}}
\end{align*}
where
\[
Z_{0}^{\prime}\equiv\int\frac{\mathrm{d}x}{\sqrt{2\pi}}\frac{e^{-\beta f\left(x\right)}}{\sqrt{1-q}}e^{-\frac{1}{2\left(1-q\right)}\left(z\sqrt{q}+x\right)^{2}}
\]

The local entropy $\mathcal{E}_{D}=\left\langle \log Z_{FP}\right\rangle $
is nothing but the total exponent for the saddle point equations,
which after some additional changes of variables can eventually be
written as

\[
\mathcal{E}_{D}=\frac{1}{2}\hat{s}s-\hat{p}p+t\hat{t}-\frac{\hat{s}}{2}+G_{S}+\alpha G_{E}
\]
where

\[
G_{S}=\int Dz\frac{\sum_{w=\pm1}e^{wz\sqrt{\hat{q}}}\int D\phi\log\left[2\cosh\left(\left(\sqrt{\hat{s}-\frac{\hat{t}^{2}}{q}}\right)\phi+\frac{\hat{t}}{\sqrt{\hat{q}}}z+\left(\hat{p}-\hat{t}\right)w\right)\right]}{2\cosh\left(z\sqrt{\hat{q}}\right)}
\]
and

\[
G_{E}=\int Dz\frac{\int Dx\,e^{-\beta f\left(x\sqrt{1-q}+z\sqrt{q}\right)}\int Du\log[\int Dye^{-\beta^{\prime}g\left(y\sqrt{1-s}+\frac{t}{\sqrt{q}}z+\frac{p-t}{\sqrt{1-q}}x+u\sqrt{\omega}\right)}]}{\int Dx\,e^{-\beta f\left(x\sqrt{1-q}+z\sqrt{q}\right)}}
\]
where we have defined $\omega=s-t-\frac{\left(p-t\right)^{2}}{1-q}+\frac{t\left(q-t\right)}{q}$.

If we now take the limit $\beta^{\prime}\to\infty$ and plug in the
expression for the error loss function \textbf{$g\left(x\right)=\Theta\left(-x\right)$},
we can eliminate one integral
\begin{align*}
\lim_{\beta^{\prime}\to\infty}\int & Dy\,e^{-\beta^{\prime}\Theta\left(-\left(y\sqrt{1-s}+\frac{t}{\sqrt{q}}z+\frac{p-t}{\sqrt{1-q}}x+w\sqrt{\omega}\right)\right)}=\\
= & \int Dy\:\Theta\left(y\sqrt{1-s}+\frac{t}{\sqrt{q}}z+\frac{p-t}{\sqrt{1-q}}x+w\sqrt{\omega}\right)=H\left(-\frac{\frac{t}{\sqrt{q}}z+\frac{p-t}{\sqrt{1-q}}x+u\sqrt{\omega}}{\sqrt{1-s}}\right)
\end{align*}
and the expression of $G_{E}$ simplifies to

\[
G_{E}=\int Dz\frac{\int Dx\,e^{-\beta f\left(x\sqrt{1-q}+z\sqrt{q}\right)}\int Du\log H\left(-\frac{\frac{t}{\sqrt{q}}z+\frac{p-t}{\sqrt{1-q}}x+u\sqrt{\omega}}{\sqrt{1-s}}\right)}{\int Dx\,e^{-\beta f\left(x\sqrt{1-q}+z\sqrt{q}\right)}}
\]

In order to compute the entropy for a given distance $D=\left(1-p\right)/2$,
we need to solve the saddle point equations with respect to $s,\hat{s},t,\hat{t},\hat{p}$
with the values of $q$ and $\hat{q}$ obtained by solving the equations
for the CE loss function. The saddle point equations can be written
as

\begin{align*}
0 & =\frac{\partial\mathcal{E}_{D}}{\partial\hat{s}}=\frac{1}{2}\left(s-1\right)+\frac{\partial G_{S}}{\partial\hat{s}}\\
0 & =\frac{\partial\mathcal{E}_{D}}{\partial\hat{p}}=-p+\frac{\partial G_{S}}{\partial\hat{p}}\\
0 & =\frac{\partial\mathcal{E}_{D}}{\partial\hat{t}}=t+\frac{\partial G_{S}}{\partial\hat{t}}\\
0 & =\frac{\partial\mathcal{E}_{D}}{\partial s}=\frac{\hat{s}}{2}+\alpha\frac{\partial G_{E}}{\partial s}\\
0 & =\frac{\partial\mathcal{E}_{D}}{\partial t}=t+\alpha\frac{\partial G_{E}}{\partial t}
\end{align*}
with

\begin{align*}
\frac{\partial G_{S}}{\partial\hat{s}} & =\int Dz\frac{\sum_{w=\pm1}e^{wz\sqrt{\hat{q}}}\int D\phi\tanh\left(\phi\sqrt{\hat{s}-\frac{\hat{t}^{2}}{\hat{q}}}+z\frac{\hat{t}}{\sqrt{\hat{q}}}+\left(\hat{p}-\hat{t}\right)w\right)\frac{\phi}{2\sqrt{\hat{s}-\frac{\hat{t}^{2}}{\hat{q}}}}}{2\cosh z\sqrt{\hat{q}}},\\
\frac{\partial G_{S}}{\partial\hat{p}} & =\int Dz\frac{\sum_{w=\pm1}e^{wz\sqrt{\hat{q}}}\int D\phi\tanh\left(\phi\sqrt{\hat{s}-\frac{\hat{t}^{2}}{\hat{q}}}+z\frac{\hat{t}}{\sqrt{\hat{q}}}+\left(\hat{p}-\hat{t}\right)w\right)w}{2\cosh z\sqrt{\hat{q}}},\\
\frac{\partial G_{S}}{\partial\hat{t}} & =\int Dz\frac{\sum_{w=\pm1}e^{wz\sqrt{\hat{q}}}\int D\phi\tanh\left(\phi\sqrt{\hat{s}-\frac{\hat{t}^{2}}{\hat{q}}}+z\frac{\hat{t}}{\sqrt{\hat{q}}}+\left(\hat{p}-\hat{t}\right)w\right)}{2\cosh z\sqrt{\hat{q}}}\left[-\frac{\hat{t}\phi}{\hat{q}\sqrt{\hat{s}-\frac{\hat{t}^{2}}{\hat{q}}}}+z\frac{\hat{t}}{\sqrt{\hat{q}}}-w\right],\\
\frac{\partial G_{E}}{\partial s} & =\int Dz\frac{\int Dx\,e^{-\beta f\left(x\sqrt{1-q}+z\sqrt{q}\right)}\int Du\,{\cal G}\left(\frac{\frac{t}{\sqrt{q}}z+\frac{p-t}{\sqrt{1-q}}x+u\sqrt{\omega}}{\sqrt{1-s}}\right)\left[\frac{u}{2\sqrt{1-s}\sqrt{\omega}}+\frac{\frac{t}{\sqrt{q}}+x\frac{p-t}{\sqrt{1-q}}+u\sqrt{\omega}}{2\left(1-s\right)^{3/2}}\right]}{\int Dx\,e^{-\beta f\left(x\sqrt{1-q}+z\sqrt{q}\right)}},\\
\frac{\partial G_{E}}{\partial t} & =\int Dz\frac{\int Dx\,e^{-\beta f\left(x\sqrt{1-q}+z\sqrt{q}\right)}\int Du\,{\cal G}\left(\frac{\frac{t}{\sqrt{q}}z+\frac{p-t}{\sqrt{1-q}}x+u\sqrt{\omega}}{\sqrt{1-s}}\right)\left[-\frac{x}{\sqrt{1-q}}+\frac{z}{\sqrt{q}}+u\frac{1}{\sqrt{\omega}}\frac{pq-t}{q\left(1-q\right)}\right]\frac{1}{\sqrt{1-s}}}{\int Dx\,e^{-\beta f\left(x\sqrt{1-q}+z\sqrt{q}\right)}}
\end{align*}
where we have defined ${\cal G}\left(-x\right)\equiv\frac{\partial}{\partial x}\log H\left[-x\right]$.

The results are reported in fig.\textbf{~}\ref{fig:FP}. We may observe
that the minima of the CE are indeed surrounded by an exponential
number of zero error solutions. In other words, the CE focuses on
HLE regions.

\subsection{RS stability and zero entropy condition}

In order to corroborate the validity of the RS solution, we need to
check two necessary conditions: the entropy of the CE model is positive
and the replica symmetric solution is stable. In other words we can
focus on values of the parameters $\alpha,\gamma,\beta$ such that
both conditions are met.

Following \citet{engel-vandenbroek}, the stability is verified if
the following condition is met

\[
\alpha\gamma_{E}\gamma_{S}<1
\]
where $\gamma_{E}$ and $\gamma_{S}$ are the two eigenvalues of the
Hessian matrix computed at the RS saddle point. We find

\[
\gamma_{S}=\int Dz\left[1-\tanh(z\sqrt{\hat{q}})\right]^{2}
\]
and 

\[
\gamma_{E}=\frac{1}{\left(1-q\right)^{2}}\int Dz\left[1-\left(\left\langle x^{2}\right\rangle _{z}-\left\langle x\right\rangle _{z}^{2}\right)\right]^{2}
\]
where the averages can be expressed through the quantity

\[
\overline{\tilde{x}^{k}\left(z\right)}=\frac{\int\frac{\mathrm{d}x\mathrm{d}\tilde{x}}{2\pi}\,\tilde{x}^{k}\,e^{-\frac{1}{2}\left(1-q\right)\tilde{x}^{2}+i\tilde{x}\left(z\sqrt{q}+x\right)}e^{-\beta f\left(x\right)}}{\int\frac{\mathrm{d}x\mathrm{d}\tilde{x}}{2\pi}e^{-\frac{1}{2}\left(1-q\right)\tilde{x}^{2}+i\tilde{x}\left(z\sqrt{q}+x\right)}e^{-\beta f\left(x\right)}}
\]
One finds:

\[
\overline{\tilde{x}\left(z\right)}=\frac{i}{1-q}\frac{\int\frac{\mathrm{d}x}{\sqrt{2\pi}}e^{-\frac{\left(z\sqrt{q}+x\right)^{2}}{2\left(1-q\right)}}\left(z\sqrt{q}+x\right)e^{-\beta f\left(x\right)}}{\int\frac{\mathrm{d}x}{\sqrt{2\pi}}e^{-\frac{\left(z\sqrt{q}+x\right)^{2}}{2\left(1-q\right)}}e}=\frac{i}{\sqrt{1-q}}\frac{\int Dx\,x\,e^{-\beta f\left(z\sqrt{q}+x\sqrt{1-q}\right)}}{\int Dxe^{-\beta f\left(z\sqrt{q}+x\sqrt{1-q}\right)}}=\frac{i}{\sqrt{1-q}}\left\langle x\right\rangle _{z}
\]
and similarly

\[
\overline{\tilde{x}^{2}\left(z\right)}=\frac{1}{\left(1-q\right)^{2}}\frac{\int Dx\,\left(1-q-\left(1-q\right)x^{2}\right)\,e^{-\beta f\left(z\sqrt{q}+x\sqrt{1-q}\right)}}{\int Dxe^{-\beta f\left(z\sqrt{q}+x\sqrt{1-q}\right)}}=\frac{1}{1-q}-\frac{\left\langle x^{2}\right\rangle _{z}}{\left(1-q\right)}
\]

For each $\alpha,$we can thus identify the values of $\gamma$ and
$\beta$ for which both the entropy is positive and the solution is
stable. In particular, $\beta$ can be chosen to be quite large, corresponding
to energies which are extremely small (RSB is expected to have relatively
minor effects at zero temperature).

\section{\label{sec:A-WFM}Wide Flat Minima for the continuous case}

Following \citet{barkai1992broken} and the technique described in
sec.~\ref{sec:A-HLE-RSB} we may analyze the existence of WFM by
studying the 1-RSB saddle point equations, with $q_{1}$ and $y$
(usually called $m$ in the 1-RSB context) used as control parameters. 

The computation of the average of $\left\langle \log V\right\rangle _{\xi}$
over the patterns by the replica methods leads to the following saddle
point expression in the large $N$ limit

\[
\frac{1}{N}\left\langle \log V\right\rangle _{\xi}=\mathrm{extr}_{q_{\ell}^{ab},\hat{q}_{\ell}^{ab},E_{\ell}^{a}}G\left(\left\{ q_{\ell}^{ab},\hat{\,q}_{\ell}^{ab},\,E_{\ell}^{a}\right\} \right)
\]
where

\[
G\left(\left\{ q_{\ell}^{ab},\hat{\,q}_{\ell}^{ab},\,E_{\ell}^{a}\right\} \right)=G_{S}\left(\left\{ q_{\ell}^{ab},\hat{\,q}_{\ell}^{ab},\,E_{\ell}^{a}\right\} \right)+\alpha G_{E}\left(\left\{ q_{\ell}^{ab}\right\} \right).
\]
Given that the distribution of the input patterns is the same for
each hidden unit, averages are expected to be independent of $\ell$
and the dependency on $\ell$ of the order parameters can be dropped
$\left\{ q_{\ell}^{ab},\hat{\,q}_{\ell}^{ab},\,E_{\ell}^{a}\right\} \to\left\{ q^{ab},\hat{\,q}^{ab},\,E^{a}\right\} $.
In 1-RSB scheme, once the conjugate order parameters $\left\{ \hat{q}^{ab},\,E^{a}\right\} $
are integrated out, the expressions for $G_{S}$ and $G_{E}$ read

\begin{eqnarray*}
G_{S}\left(q_{0},q_{1},y\right) & = & \frac{1}{2}\left[\frac{1+\left(y-1\right)\Delta q_{1}}{1-q_{1}+y\Delta q_{1}}+\ln2\pi+\left(1-\frac{1}{y}\right)\ln\left(1-q_{1}\right)+\frac{1}{y}\ln\left(1-q_{1}+y\Delta q_{1}\right)\right]\\
G_{E}\left(q_{0},q_{1},y\right) & = & \frac{1}{y}\int\prod_{\ell=1}^{K}Dv_{\ell}\ln\left(\int\prod_{\ell=1}^{K}Du_{\ell}\left(\Sigma_{\left(K\right)}\right)^{y}\right)
\end{eqnarray*}
where $\Delta q_{1}=q_{1}-q_{0}$ and $\Sigma_{\left(K\right)}$ is
a complicated function of the order parameters which for $K=3$ reads

\[
\Sigma_{\left(3\right)}=H_{1}H_{2}+H_{1}H_{3}+H_{2}H_{3}-2H_{1}H_{2}H_{3}
\]
with

\[
H_{\ell}=H\left[\sqrt{\frac{\Delta q_{1}}{1-q_{1}}}u_{\ell}+\sqrt{\frac{q_{0}}{1-q_{1}}}v_{\ell}\right].
\]

For our WFM computation, $q_{1}$ is the constrained overlap between
the weight vectors of the $m$ real replicas. $q_{0}$ is the only
parameter for which we have to solve the saddle point equation. In
order to look for the WFM of maximum volume we are interested in the
large $y$ limit. In this case the expressions simplify substantially

\begin{align*}
G_{S}\left(q_{1}\right) & =\frac{1}{2}\left[1+\ln2\pi+\ln\left(1-q_{1}\right)\right]\\
G_{E}\left(q_{0},q_{1}\right) & =\int\prod_{\ell=1}^{K}Dv_{\ell}max_{u_{1,},u_{2},u_{3}}\left[-\frac{\sum_{\ell=1}^{3}u_{\ell}^{2}}{2}+\log\left(\tilde{H}_{1}\tilde{H}_{2}+\tilde{H}_{1}\tilde{H}_{3}+\tilde{H}_{2}\tilde{H}_{3}-2\tilde{H}_{1}\tilde{H}_{2}\tilde{H}_{3}\right)\right]
\end{align*}
where $\tilde{H}_{\ell}\equiv H\left[\sqrt{\frac{d_{0}}{1-q_{1}}}u_{\ell}+\sqrt{\frac{q_{0}}{1-q_{1}}}v_{\ell}\right]$
and $d_{0}\equiv y\Delta q_{1}$. The pre-factor $\frac{1}{y}$ of
$G_{E}$ has been eliminated by a change of variables $u_{\ell}^{\prime}=\frac{u_{\ell}}{\sqrt{y}}$
(with $u^{\prime}$ then renamed $u$).

Notice that $G_{S}\left(q_{1}\right)$ corresponds to the volume at
$\alpha=0$, i.e.~to the volume of the weight space just under the
spherical constraints, with the real replicas forced to be at an overlap
$q_{1}$. If WFM exist for positive $\alpha$, we expect to observe
the normalized local entropy $\mathscr{V}_{1}\left(q_{1}\right)=\alpha G_{E}\left(q_{1}\right)$
to approach $0$ for $q_{1}$ sufficiently close to one.

In fig.\textbf{~}\ref{fig:treeK3} (top panel) we report the values
of the WFM volumes vs the overlap $q_{1},$ for different values of
$\alpha.$ Indeed one may observe that the behavior is qualitatively
similar to that of the binary perceptron, WFM exist deep into the
RSB region, i.e. besides the RSB states and all the related local
minima and saddles, there exist absolute minima which are flat at
large distances. We mention that evaluating the $\max$ inside the
integral is a quite challenging task as the function to be maximized
may present multiple maxima. We have tackled this problem by an appropriate
sampling technique.

The case $K=3$ is still relatively close to the perceptron, though
the geometrical structure of its minima is already dominated by non
convex RSB features for \textbf{$\alpha>1.76$}. A case which is closer
to more realistic NN is $K\gg1$, which, luckily enough, is easier
to study analytically \citep{barkai1992broken}.

The 1-RSB expression for $G$, simplifies to

\begin{align*}
G_{S}\left(q_{0},q_{1},y\right) & =\frac{1}{2}\left[\frac{1+\left(y-1\right)\Delta q_{1}}{1-q_{1}+y\Delta q_{1}}+\ln2\pi+\left(1-\frac{1}{y}\right)\ln\left(1-q_{1}\right)+\frac{1}{y}\ln\left(1-q_{1}+y\Delta q_{1}\right)\right]\\
G_{E}\left(q_{0},q_{1},y\right) & =\frac{1}{y}\int Dv\,\ln\int Du\:\left(H\left[\sqrt{\frac{\Delta q_{1\mathrm{eff}}}{1-q_{1\mathrm{eff}}}}u+\sqrt{\frac{q_{0\mathrm{eff}}}{1-q_{1\mathrm{eff}}}}v\right]\right)^{y}
\end{align*}
where $q_{0\mathrm{eff}}=1-\frac{2}{\pi}\arccos q_{0}$, $q_{1\mathrm{eff}}=1-\frac{2}{\pi}\arccos q_{1}$and
$\Delta q_{1\mathrm{eff}}=q_{1\mathrm{eff}}-q_{0\mathrm{eff}}$. While
the critical capacity diverges with $K$ as $\sqrt{\ln K}$ \citep{monasson1995weight},
the value of $\alpha$ at which RSB sets in and the landscape of the
minima becomes non trivial remains finite, $\alpha_{RSB}\simeq2.95$.

As we have done for the $K=3,$ we study the large $y$ limit. We
find

\begin{align*}
G_{S}\left(q_{1}\right) & =\frac{1}{2}\left[1+\ln2\pi+\ln\left(1-q_{1}\right)\right]\\
G_{E}\left(q_{0},q_{1}\right) & =\int Dv\;\max_{u}\left\{ -\frac{u^{2}}{2}+\log\left[H\left(\sqrt{\frac{\Delta q_{1}^{e}}{1-q_{1}^{e}}}u+\frac{q_{0}^{e}}{1-q_{1}^{e}}v\right)\right]\right\} 
\end{align*}
In fig.\textbf{~}\ref{fig:treeK3} (bottom panel) we observe that
WFM are indeed still present.

In order to check the validity of the WFM computation one should check
for the stability of the solutions of the saddle point equations by
a stability analysis or a 2-RSB computation. For numerical reasons
this is a quite difficult task. We thus decided to follow a different
path which has also algorithmic interest, namely to study the problem
by Belief Propagation.

\subsection{Large deviation analysis of the Parity Machine\label{subsec:A-Parity}}

Here we report the results of the large deviation analysis on the
so-called parity machine. Its network structure is identical to the
committee machine, except for the output unit which performs the product
of the $K$ hidden units instead of taking a majority vote. The outputs
of the hidden units are still given by sign activations. Thus, the
overall output of the network reads:

\[
\sigma_{\mathrm{out}}^{\mu}=\prod_{\ell=1}^{K}\tau_{\ell}.
\]
For a given set of patterns, the volume of the weights which correctly
classifies the patterns is then given by

\[
V=\int\prod_{i\ell}\mathrm{d}w_{\ell i}\prod_{\ell}\delta\left(\sum_{i}w_{\ell i}^{2}-\tilde{N}\right)\prod_{\mu}\Theta\left(\sigma^{\mu}\prod_{\ell=1}^{K}\tau_{\ell}^{\mu}\right).
\]

The computation proceeds as for the committee machine case, until
we find the following expressions for the 1-RSB volume:

\begin{align*}
G_{S}\left(q_{1}\right) & =\frac{1}{2}\left[\frac{1+\left(y-1\right)\Delta q_{1}}{1-q_{1}+y\Delta q_{1}}+\ln2\pi+\left(\frac{y-1}{y}\right)\ln\left(1-q_{1}\right)+\frac{1}{y}\ln\left(1-q_{1}+y\Delta q_{1}\right)\right]\\
G_{E}\left(q_{1}\right) & =\frac{1}{y}\int\prod_{\ell=1}^{K}Dv_{\ell}\ln\left[\int\prod_{\ell=1}^{K}Du_{\ell}\left(\sum_{\left\{ \tau_{\ell}\right\} }\prod_{\ell}H_{\ell}\left(\tau_{\ell}\omega_{\ell}\right)\Theta\left(\prod_{\ell=1}^{K}\tau_{\ell}\right)\right)^{y}\right]
\end{align*}
where $\Delta q_{1}=q_{1}-q_{0}$ and where $\omega_{\ell}=\sqrt{\frac{\Delta q_{1}}{1-q_{1}}}u_{\ell}+\sqrt{\frac{q_{0}}{1-q_{1}}}v_{\ell}$
and $q_{0}$ is fixed by a saddle point equation. The sum over the
internal states be computed for general $K,$ leading to the following
final expression for $G_{E}$:
\[
G_{E}\left(q_{1}\right)=\frac{1}{y}\int\prod_{\ell=1}^{K}Dv_{\ell}\ln\left[\int\prod_{\ell=1}^{K}Du_{\ell}\frac{1}{2^{y}}\left(1+\zeta_{K}\left(\left\{ \omega_{\ell}\right\} \right)\right)^{y}\right]
\]
where $\zeta_{K}\left(\left\{ \omega_{\ell}\right\} \right)=\left(-1\right)^{K}\prod_{\ell=1}^{K}\left(1-2H_{\ell}\left(\omega_{\ell}\right)\right)$.
In the large $K$ limit, $\zeta_{K}\left(\left\{ \omega_{\ell}\right\} \right)$
converges rapidly to zero and the expression for $G_{E}$ simplifies.
We can thus compute the volume by optimizing over $q_{0}$ for arbitrary
$y$ and $q_{1}$, and compare it to the volume that one would find
for the same distance when no patterns are stored: the log-ratio of
the two volumes is constant and equal to $-\alpha\log\left(2\right)$.
This shows that the minima never become flat, at any distance scale.

\section{\label{sec:A-BP}Belief Propagation on a Tree-like committee machine
with continuous weights: equations, local volume around solutions
and algorithms for the replicated network}

\subsection{BP equations for the committee machine}

We can use Belief Propagation (see e.g.~\citet{krzakala2016statistical,braunstein2006learning,baldassi_unreasonable_2016})
to study the space of the solutions of a tree-like committee machine
with continuous weights and random inputs (the outputs can either
be random or generated from a rule). The messages in this case are
probability density distributions over $\mathbb{R}$, and we will
need to ensure normalization by using an additional constraints over
the norm of the weights vectors.

The basic factor graph is thus composed of $N$ continuous variable
nodes $x_{i}$, divided into $K$ groups of $N/K$ variables each;
for each pattern, we will have $K$ factor nodes, each one involving
one group of $N/K$ variables and an auxiliary binary output variable
$\tau_{ka}$ (with two indices, one for the hidden unit and one for
the pattern), and another factor node connected to the $\tau$ variables
and controlling the final output. We will enforce the normalization
of the weights by adding an extra field for each variable $x_{i}$,
as explained below.

As a general notation scheme, we will use the letter $h$ to denote
messages from variable nodes to factor nodes, the letter $u$ for
messages from factor nodes to variable nodes, and the letter $m$
for non-cavity marginals over variables.

Let's then call $h_{ki\to ka}\left(x_{ki}\right)$ the cavity message
from variable $ki$ (representing a weight with hidden unit index
$k$$\in\left\{ 1,\dots,K\right\} $, weight index $i\in\left\{ 1,\dots,N/K\right\} $,
whose value is $x_{ki}$) to the factor node $ka$ (representing the
part of an input pattern $a$ which involves the hidden unit $k$).
The BP equation reads:
\begin{equation}
h_{ki\to ka}\left(x_{ki}\right)\propto u_{\mathrm{n}}\left(x_{ki}\right)\prod_{b\ne a}u_{kb\to ki}\left(x_{ki}\right)\label{eq:BP_h}
\end{equation}
where $u_{kb\to ki}\left(x_{ki}\right)$ represents a message from
another pattern node $kb$ to the variable $ki$, while $u_{\mathrm{n}}\left(x_{ki}\right)$
is an external field enforcing the normalization constraint, which
is the same for all variables:
\begin{equation}
u_{\mathrm{n}}\left(x\right)=e^{-\frac{\chi}{2}x^{2}}\label{eq:u_n}
\end{equation}
This method of enforcing normalization is equivalent to using a Dirac
delta in the limit of large $N$, but it's easier to implement in
the BP algorithm. In order to set the parameter $\chi$ we will need
to evaluate the norm of the weights at convergence (detailed below)
and adjust $\chi$ until the norm matches the desired value, $\sum_{ki}x_{ki}^{2}=N$.

The other messages read:
\begin{eqnarray}
 &  & u_{ka\to ki}\left(x_{ki}\right)=\label{eq:BP_u}\\
 &  & \sum_{\tau_{ka}=\pm1}h_{ka}\left(\tau_{ka}\right)\int\prod_{j\ne i}h_{kj\to ka}\left(x_{kj}\right)\mathrm{d}x_{kj}\;\Theta\left(\tau_{ka}\left(\sum_{j\ne i}x_{kj}\xi_{kj}^{ka}+x_{ki}\xi_{ki}^{ka}\right)\right)\nonumber 
\end{eqnarray}
where the message $h_{ka}$ goes from the auxiliary output variable
$\tau_{ka}$ to node $ka$. Notice that the messages $h$ are assumed
to be normalized, but the messages $u$ aren't, because the integral
of expression~(\ref{eq:BP_u}) might diverge.

In the limit of large $N$, eq.~(\ref{eq:BP_u}) can be approximated
using the central limit theorem, since we're assuming that the input
pattern entries are random i.i.d. variables. Therefore, we don't need
the full distributions $h_{kj\to ka}$ to perform the integral, only
their first and second moments. For simplicity of implementation,
though, instead of the moments it is actually more convenient to use
the inverse of the variance and the mean rescaled by the variance,
which we will denote with $\zeta$ and $\tilde{\mu}$, respectively.
We will use the following notation:
\begin{eqnarray}
\zeta_{ki\to ka} & = & \left(\left\langle x_{ki}^{2}\right\rangle _{h_{ki\to ka}}-\left\langle x_{ki}\right\rangle _{h_{ki\to ka}}^{2}\right)^{-1}\label{eq:zeta}\\
\tilde{\mu}_{ki\to ka} & = & \zeta_{ki\to ka}\,\left\langle x_{ki}\right\rangle _{h_{ki\to ka}}\label{eq:mutilde}
\end{eqnarray}

With these, we can compute the two auxiliary quantities
\begin{eqnarray}
c_{ka\to ki} & = & \sum_{j\ne i}\frac{\tilde{\mu}_{kj\to ka}}{\zeta_{kj\to ka}}\xi_{kj}^{ka}\label{eq:c}\\
v_{ka\to ki} & = & \sum_{j\ne i}\zeta_{kj\to ka}^{-1}\left(\xi_{kj}^{ka}\right)^{2}\label{eq:v}
\end{eqnarray}
which can be simplified by noting that $\left(\xi_{kj}^{ka}\right)^{2}=1$.
We can now rewrite eq.~(\ref{eq:BP_u}) using the central limit theorem:
\begin{eqnarray}
u_{ka\to ki}\left(x_{ki}\right) & = & \sum_{\tau_{ka}=\pm1}h_{ka}\left(\tau_{ka}\right)\int Dz\;\Theta\left(\tau_{ka}\left(c_{ka\to ki}+z\sqrt{v_{ka\to ki}}+x_{ki}\xi_{ki}^{ka}\right)\right)\nonumber \\
 & = & \frac{1}{2}\left(1+h_{ka}\mathrm{erf}\left(\frac{c_{ka\to ki}+x_{ki}\xi_{ki}^{ka}}{\sqrt{2v_{ka\to ki}}}\right)\right)
\end{eqnarray}
where in the last line we have introduced the shorthand notation 
\begin{eqnarray*}
h_{ka} & \equiv & h_{ka}\left(+1\right)-h_{ka}\left(-1\right)
\end{eqnarray*}
because we can represent any message over a binary variable with a
single quantity (in this case, the ``magnetization'' often employed
in spin glass literature) and therefore we abuse the notation and
identify that message with a single parameter. The context and the
presence or absence of the argument will suffice to disambiguate the
notation.

In the limit of large $N$, we observe that $c_{ka\to ki}=O\left(\sqrt{N}\right)$
and $v_{ka\to ki}=O\left(N\right)$, and therefore $\frac{c_{ka\to ki}}{\sqrt{v_{ka\to ki}}}=O\left(1\right)$,
but $\frac{x_{ki}\xi_{ki}^{ka}}{\sqrt{v_{ka\to ki}}}=O\left(\frac{1}{\sqrt{N}}\right)$
since we are assuming $x_{ki}=O\left(1\right)$ due to the normalization
constraint. We can therefore expand $u_{ka\to ki}$ to the second
order, obtaining (up to an irrelevant factor):
\begin{equation}
u_{ka\to ki}\left(x_{ki}\right)\propto\text{1+\ensuremath{U_{ka\to ki}x_{ki}}-\ensuremath{\frac{1}{2}\left(V_{ka\to ki}-U_{ka\to ki}^{2}\right)x_{ki}^{2}}}
\end{equation}
where
\begin{eqnarray}
U_{ka\to ki} & = & \frac{\xi_{ki}^{ka}}{\sqrt{v_{ka\to ki}}}\frac{2h_{ka}\frac{1}{\sqrt{2\pi}}\exp\left(-\frac{1}{2}\frac{c_{ka\to ki}^{2}}{v_{ka\to ki}}\right)}{1+h_{ka}\text{erf}\left(\frac{c_{ka\to ki}}{\sqrt{2v_{ka\to ki}}}\right)}\label{eq:U}\\
V_{ka\to ki} & = & \frac{\xi_{ki}^{ka}}{\sqrt{v_{ka\to ki}}}\frac{c_{ka\to ki}}{\sqrt{v_{ka\to ki}}}U_{ka\to ki}+U_{ka\to ki}^{2}\label{eq:V}
\end{eqnarray}
Now we can substitute this into eq.~(\ref{eq:BP_h}):
\[
h_{ki\to ka}\left(x_{ki}\right)=\frac{1}{z_{ki\to ka}}e^{-\frac{\chi}{2}x_{ki}^{2}}\prod_{b\ne a}\left(\text{1+\ensuremath{U_{kb\to ki}x_{ki}}-\ensuremath{\frac{1}{2}\left(V_{ka\to ki}-U_{ka\to ki}^{2}\right)x_{ki}^{2}}}\right)
\]
where $z_{ki\to ka}$ is a normalization constant. Expanding this
to the second order we finally obtain:
\begin{equation}
h_{ki\to ka}\left(x_{ki}\right)=\frac{1}{z_{ki\to ka}}e^{-\frac{1}{2}\left(\chi+\sum_{b\ne a}V_{kb\to ki}\right)x_{ki}^{2}+\sum_{b\ne a}U_{kb\to ki}x_{ki}}
\end{equation}
and thus we obtain the following simple expression for the parameters
of the distribution eqs.~(\ref{eq:zeta})-(\ref{eq:mutilde}):
\begin{eqnarray}
\zeta_{ki\to ka} & = & \chi+\sum_{b\ne a}V_{kb\to ki}\label{eq:zeta_2}\\
\tilde{\mu}_{ki\to ka} & = & \sum_{b\ne a}U_{kb\to ki}\label{eq:mutilde_2}
\end{eqnarray}
Also, the normalization constant is:
\begin{equation}
z_{ki\to ka}=\frac{\sqrt{2\pi}}{\sqrt{\lambda_{ki\to ka}}}e^{\frac{1}{2}\frac{\text{\ensuremath{\tilde{\mu}_{ki\to ka}}}^{2}}{\zeta_{ki\to ka}}}
\end{equation}
One can immediately write the corresponding formulas for the non-cavity
parameters of the marginal $m_{ki}\left(x_{ki}\right)$:
\begin{eqnarray}
\zeta_{ki} & = & \chi+\sum_{b}V_{kb\to ki}\\
\tilde{\mu}_{ki} & = & \sum_{b}U_{kb\to ki}\\
z_{ki} & = & \frac{\sqrt{2\pi}}{\sqrt{\lambda_{ki}}}e^{\frac{1}{2}\frac{\text{\ensuremath{\tilde{\mu}_{ki}}}^{2}}{\zeta_{ki}}}
\end{eqnarray}
This form of the equations shows that we can, in fact, parametrize
all the distributions as Gaussian, and also that the updates in the
equations can be performed efficiently for each node by keeping in
memory the non-cavity versions of the parameters $\zeta$ and $\tilde{\mu}$
and the parameters $U$ and $V$, such that updating $\zeta,\tilde{\mu}$
is only a matter of subtracting the previous value of $U,V$ and adding
the new one.

We can also note that the computation of $U,V$ in eqs.~(\ref{eq:U})-(\ref{eq:V})
can be made computationally more efficient by writing $c_{ka\to ki}$
and $v_{ka\to ki}$ in terms of their non-cavity counterparts

\begin{eqnarray}
c_{ka} & = & \sum_{j}\frac{\tilde{\mu}_{kj\to ka}}{\zeta_{kj\to ka}}\xi_{kj}^{ka}\label{eq:c-noncav}\\
v_{ka} & = & \sum_{j}\zeta_{kj\to ka}\left(\xi_{kj}^{ka}\right)^{2}\label{eq:v-noncav}
\end{eqnarray}
and expanding. However, in practice this further approximation does
not work well at large values of $\alpha$ when using the Focusing-BP
protocol (detailed below, sec.~\ref{subsec:A-fBP}) and at moderate
values of $N$: by avoiding it, we can find solutions at slightly
larger values of $\alpha$ (e.g.~we can reach $\alpha=2.7$ instead
of $2.6$ at $N=999$, $K=3$). Relatedly, it is also slightly problematic
at large values of the polarization field $\lambda$ when exploring
the space of configurations (see below, sec.~\ref{subsec:A-FranzParisi}).
In general terms, the problems arise when the fields become very polarized
and the normalization constants become very small. Therefore, in our
implementation this approximation is optional.

In order to set the normalization parameter $\chi$, we simply compute
the following quantity at the end of each iteration
\begin{equation}
\frac{1}{N}\sum_{ki}\left\langle x_{ki}^{2}\right\rangle =\frac{1}{N}\sum_{ki}\left(\zeta_{ki}^{-1}+\zeta_{ki}^{-2}\tilde{\mu}_{ki}^{2}\right)\label{eq:q}
\end{equation}
and adjust $\chi$ so that this becomes $1$. Of course, we also add
the criterion that this adjustment needs to be sufficiently small
in evaluating whether the BP equations have converged.

The remaining BP equations involve the nodes connecting the auxiliary
variables $\tau_{ka}$ and enforcing the desired outputs from the
committee:
\begin{eqnarray}
u_{ka}\left(\tau_{ka}\right) & = & \frac{1}{2}\left(1+\tau_{ka}\mathrm{erf}\left(\frac{c_{ka}}{\sqrt{2v_{ka}}}\right)\right)\label{eq:u_ka}\\
h_{ka}\left(\tau_{ka}\right) & = & \sum_{\left\{ \tau_{la}\right\} _{l\ne k}}\prod_{l\ne k}u_{la}\left(\tau_{la}\right)\Theta\left(\sigma_{a}\left(\sum_{l\ne k}\tau_{la}+\tau_{ka}\right)\right)\label{eq:h_ka}
\end{eqnarray}
where $\sigma_{a}$ is the desired output for pattern $a$. For small
$K$, we compute $h_{ka}$ exactly (i.e. without using the central
limit theorem), which can be done in $O\left(K^{2}\right)$ time.

Equations (\ref{eq:c}), (\ref{eq:v}), (\ref{eq:U}), (\ref{eq:V}),
(\ref{eq:zeta_2}), (\ref{eq:mutilde_2}), (\ref{eq:c-noncav}), (\ref{eq:v-noncav}),
(\ref{eq:u_ka}) and (\ref{eq:h_ka}), together with the normalization
requirement obtained through eq.~(\ref{eq:q}), form the full set
of BP equations.

The free entropy (also sometimes known as action) of the system can
be computed from the usual BP formulas. We get:
\begin{equation}
\phi=\frac{1}{N}\left(\sum_{ki}f_{ki}-\sum_{ka}f_{ka}-\sum_{a}f_{a}\right)\label{eq:free_entropy}
\end{equation}
where:
\begin{eqnarray}
f_{ki} & = & \log z_{ki}\\
f_{ka} & = & \sum_{i}\left(\log z_{ki\to ka}-\log z_{ki}\right)\\
f_{a} & = & -\log\left(1+\frac{1}{2}\sigma_{a}\mathrm{erf}\left(\frac{\sum_{k}h_{ka}}{\sqrt{2\sum_{k}\left(1-h_{ka}^{2}\right)}}\right)\right)
\end{eqnarray}
From this, we can compute the entropy by simply accounting for the
energetic contribution introduced by the normalization constraint.
We also shift it by subtracting the log-volume of the normalized sphere,
so that its value is upper bounded by $0$
\begin{equation}
S=\phi+\frac{\chi}{2}-\log\left(\sqrt{2\pi e}\right)\label{eq:entropy}
\end{equation}

\subsection{Exploring the space of solutions around a given configuration\label{subsec:A-FranzParisi}}

Given a particular configuration $\tilde{W}$, which we assume normalized
as $\sum_{ki}\tilde{w}_{ki}^{2}=N$, we are interested in exploring
the space of solutions at a given distance $D$ from it (for a suitable
definition of the distance). Analogously to the norm, in the limit
of large $N$ we can control the distance by just adding an extra
field to each node (putting it as an extra factor in eq.~(\ref{eq:BP_h})),
of the form:
\[
u_{D}\left(x_{ki}\right)=e^{-\frac{\lambda}{2}\left(x_{ki}-\tilde{w}_{ki}\right)^{2}}
\]

By varying the auxiliary parameter $\lambda$ between $0$ and $\infty$
we can restrict ourselves to smaller and smaller regions around $\tilde{w}$.
Adding this extra field in practice just amounts at adding two terms
$\lambda\tilde{w}_{ki}$ and $\lambda$ to the expressions of $\tilde{\mu}_{ki\to ka}$
and $\zeta_{ki\to ka}$, respectively (eqs.~(\ref{eq:zeta_2})-(\ref{eq:mutilde_2})).

For convenience, we actually abuse the terminology and use a squared
distance in our definition:
\begin{equation}
d\left(x,\tilde{w}\right)=\frac{1}{2N}\left\Vert x-\tilde{w}\right\Vert _{2}^{2}
\end{equation}
which can be computed from the BP messages at convergence, as follows.
First define the auxiliary quantities (representing the cavity variance
and mean of each variable without the $u_{D}$ field):

\begin{eqnarray}
r_{ki\to D} & = & \left\langle x_{ki}^{2}\right\rangle _{h_{ki\to D}}-\left\langle x_{ki}\right\rangle _{h_{ki\to D}}^{2}=\left(\zeta_{ki}-\lambda\right)^{-1}\\
\mu_{ki\to D} & = & \left\langle x_{ki}\right\rangle _{h_{ki\to D}}=\frac{\tilde{\mu}_{ki}-\lambda\tilde{w}_{ki}}{r_{ki\to D}}
\end{eqnarray}
Then the expression of the average distance reads:
\begin{eqnarray}
\left\langle d\left(x,\tilde{W}\right)\right\rangle  & = & \frac{1}{N}\sum_{ki}\frac{\int\mathrm{d}x_{ki}\:u_{D}\left(x_{ki}\right)h_{ki\to D}\left(x_{ki}\right)\frac{1}{2}\left(x_{ki}-\tilde{w}_{ki}\right)^{2}}{\int\mathrm{d}x_{ki}\:u_{D}\left(x_{ki}\right)h_{ki\to D}\left(x_{ki}\right)}\nonumber \\
 & = & \frac{1}{N}\sum_{ki}\frac{1}{2\lambda}\left(\frac{r_{ki\to D}}{r_{ki\to D}+\lambda^{-1}}+\frac{1}{\gamma}\left(\frac{\mu_{ki\to D}-\tilde{w}_{ki}}{r_{ki\to D}+\lambda^{-1}}\right)^{2}\right)\label{eq:distp}
\end{eqnarray}
The free entropy of the system can be written as before, eq.~(\ref{eq:free_entropy}),
but now in the computation of the entropy we also need to account
for the energy of the extra field:
\begin{equation}
S=\phi+\frac{\chi+\lambda}{2}-\frac{\lambda}{N}\sum_{ki}\tilde{w}_{ki}\frac{\tilde{\mu}_{ki}}{\lambda_{ki}}-\log\left(\sqrt{2\pi e}\right)\label{eq:local_entropy-1}
\end{equation}

By varying $\lambda$ and using eqs.~(\ref{eq:distp}) and~(\ref{eq:local_entropy-1}),
we can obtain a plot of the local entropy as a function of the distance
around any given configuration $\tilde{W}$, as long as the BP equations
converge. As a general rule of thumb, the equations don't converge
when $\lambda$ is too low in the 1-RSB phase, and when $\lambda$
is too large and $\tilde{W}$ is not a solution. The equations do
converge however even in the 1-RSB phase for large enough $\lambda$
if $\tilde{W}$ is a solution, which can be understood as the external
field breaking the symmetry. When $\tilde{W}$ is not a solution,
going to the limit $\lambda\to\infty$ eventually restricts the BP
equations to a region of the configuration space without solutions,
leading to non-normalizable messages (this could of course be amended
e.g. by just working at non-zero temperature). Besides these situations,
other numerical problems may arise under certain circumstances when
$N$ is not large enough, due to the approximations in the messages.

As a consistency check, it can be verified numerically that formula~(\ref{eq:local_entropy-1})
yields the expected result for the $\alpha=0$ case, $S=\log\left(\sqrt{D\left(2-D\right)}\right)$.

\subsection{Focusing-BP\label{subsec:A-fBP}}

In order to implement the Focusing-BP protocol we need to add a new
type of node to each variable. These nodes do not directly represent
energy terms in the usual sense, and therefore they are not factor
nodes; rather, their role is that of effectively representing an interaction
of $y$ identical replicas of the original system with an extra auxiliary
configuration $x^{\star}$. The derivation (in a discrete setting)
can be found in \citet{baldassi_unreasonable_2016}.

We denote as $u_{\star\to ki}\left(x_{ki}\right)$ the new extra field
to be multiplied in eq.~(\ref{eq:BP_h}). This field, like all others
over the $x_{ki}$ variables, is parametrized by two quantities $U_{\star\to ki}$
and $V_{\star\to ki}$ which get added to $\tilde{\mu}_{ki}$ and
$\zeta_{ki}$ (eqs.~(\ref{eq:zeta_2}) and~(\ref{eq:mutilde_2})).The
update equation is rather involved, but it can be simplified by breaking
it down in steps and adopting a new notation for messages composition,
leading to this expression:
\begin{equation}
u_{\star\to ki}=\left(\left(\left(h_{ki\to\star}*\lambda\right)\uparrow\left(y-1\right)\right)\otimes u_{\mathrm{n}^{\star}}\right)*\lambda\label{eq:u_star}
\end{equation}
where the notation is as follows:
\begin{itemize}
\item $u=h*\lambda$ represents (intuitively speaking) the effect of passing
a message $h$ through a Gaussian interaction with strength $\lambda$.
In formulas:
\[
u\left(\tilde{x}\right)=\int\mathrm{d}x\,e^{-\frac{\lambda}{2}\left(\tilde{x}-x\right)^{2}}h\left(x\right)
\]
\item $h=u_{1}\otimes u_{2}$ represents the ``composition'' of two messages,
in formulas:
\[
h\left(x\right)\propto u_{1}\left(x\right)u_{2}\left(x\right)
\]
which, using the internal representation in terms of rescaled mean
$\tilde{\mu}$ and inverse variance $\zeta$ for the $h$, and the
corresponding quantities for the $u_{1/2}$ fields, simply translates
to:
\begin{eqnarray*}
\tilde{\mu} & = & U_{1}+U_{2}\\
\zeta & = & V_{1}+V_{2}
\end{eqnarray*}
\item $\hat{u}=u\uparrow y$ with integer $y$ represents the composition
of $u$ with itself (i.e.~$\hat{u}=u\otimes u\otimes\dots\otimes u$)
$y$ times. This can be trivially extended to non-integer $y$ and
yields:
\begin{eqnarray*}
\hat{U} & = & yU\\
\hat{V} & = & yV
\end{eqnarray*}
\item $u_{\mathrm{n}^{\star}}$ is a normalization field, required for normalization
of the $x^{\star}$ variables analogously to the $u_{\mathrm{n}}$
field for the $x$ variables; analogously to that field, it is parametrized
with a single parameter $\chi^{\star}$ and can be represented as
$U_{\mathrm{n}^{\star}}=0$, $V_{\mathrm{n}^{\star}}=\chi^{\star}$
(cf. equations~(\ref{eq:BP_h}), (\ref{eq:u_n}) and~(\ref{eq:zeta_2})).
\item $h_{ki\to\star}$ is the cavity field from a variable $x_{ki}$ to
the corresponding variable $x_{ki}^{\star}$, which can be defined
from the relation $m_{ki}=h_{ki\to\star}\otimes u_{\star\to ki}$.
\end{itemize}
Therefore formula (\ref{eq:u_star}) can be intuitively understood
as follows: the cavity messages coming from the replicated variable
nodes ($h_{ki\to\star})$ are passed through their interaction (with
strength $\lambda)$ with the corresponding variable $x_{ki}^{\star}$;
the resulting messages are composed together (there are $y-1$ identical
messages) and with the normalization field $u_{\mathrm{n}^{\star}}$;
the resulting cavity message $h_{\star\to ki}$ is again passed through
the interaction $\lambda$ to yield the field $u_{\star\to ki}$.

The parameter $\chi^{\star}$ must be set to a value that normalizes
the $x^{\star}$ variables; the norm of the $x^{\star}$ variables
can be obtained from their marginals with the same formula used for
the $x$, eq.~(\ref{eq:q}); the marginals can be obtained with this
formula:
\[
m_{\star}=\left(\left(h_{ki\to\star}*\lambda\right)\uparrow y\right)\otimes u_{\mathrm{n}^{\star}}
\]

\section{Numerical experiments details\label{subsec:A-Numerical}}

Here we provide the details and the settings used for the numerical
experiments reported in sec.~\ref{sec:numerics}.

In all the experiments and for all algorithms except fBP we have used
a mini-batch size of $100$. The mini-batches were generated by randomly
shuffling the datasets and splitting them at each epoch. For eLAL,
the permutations were performed independently for each replica. Also,
for all algorithms except fBP the the weights were initialized from
a uniform distribution and then normalized for each unit. The learning
rate $\eta$ was kept fixed throughout the training. The parameters
$\gamma$ and $\beta$ for the ceSGD algorithm were initialized at
some values $\gamma_{0}$, $\beta_{0}$ and multiplied by $1+\gamma_{1}$,
$1+\beta_{1}$ after each epoch. Analogously, the parameter $\lambda$
for the eLAL algorithm was initialized to $\lambda_{0}$ and multiplied
by $1+\lambda_{1}$ after each epoch. The parameter $\lambda$ for
the fBP algorithm ranged in all cases between $0.5$ and $30$ with
an exponential schedule divided into $30$ steps; at each step, the
algorithm was run until convergence or at most $200$ iterations.
We used $y=20$ for eLAL and $y=10$ for fBP. The stopping criterion
for ceSGD-fast was that a solution (0 errors with $\beta=\infty$)
was found; for ceSGD-slow, that the CE loss reached $10^{-7}$; for
eLAL, that the sum of the squared distances between each replica and
the average replica $\tilde{W}$ reached $10^{-7}$. We also report
here the average and standard deviation of the number of epochs $\overline{T}$
for each algorithm.

\emph{Parameters for the case of random patterns.} ceSGD-fast: $\eta=10^{-2}$,
$\gamma_{0}=3$, $\beta_{0}=1$, $\gamma_{1}=1$, $\beta_{1}=10^{-3}$
($\overline{T}=770\pm150$). ceSGD-slow: $\eta=3\cdot10^{-3}$, $\gamma_{0}=0.1$,
$\beta_{0}=0.5$, $\gamma_{1}=4\cdot10^{-4}$, $\beta_{1}=2\cdot10^{-4}$
($\overline{T}=\left(1.298\pm0.007\right)\cdot10^{4}$). LAL: $\eta=5\cdot10^{-3}$
($\overline{T}=76\pm15$). eLAL: $\eta=10^{-2}$, $\lambda_{0}=0.5$,
$\lambda_{1}=10^{-4}$ ($\overline{T}=861\pm315$).

\emph{Parameters for the Fashion-MNIST experiments.} ceSGD-fast: $\eta=2\cdot10^{-4}$,
$\gamma_{0}=5$, $\beta_{0}=2$, $\gamma_{1}=1$, $\beta_{1}=10^{-4}$
($\overline{T}=460\pm334$). ceSGD-slow: $\eta=3\cdot10^{-5}$, $\gamma_{0}=0.5$,
$\beta_{0}=0.5$, $\gamma_{1}=10^{-3}$, $\beta_{1}=10^{-3}$ ($\overline{T}=\left(3.57\pm0.05\right)\cdot10^{3}$).
LAL: $\eta=10^{-4}$ ($\overline{T}=61\pm21$). eLAL: $\eta=2\cdot10^{-3}$,
$\lambda_{0}=30$, $\lambda_{1}=5\cdot10^{-3}$ ($\overline{T}=190\pm40$).

\section{Experiments with randomized Fashion-MNIST\label{sec:A-rand-FMNIST}}

Our protocol to produce a randomized versions of the Fashion-MNIST
dataset was as follows. Starting with the binarized, two-class dataset
described in the main text, new input patterns were derived from the
original ones as $\xi_{i}^{\mu\prime}=\xi_{i}^{\pi_{i}\left(\mu\right)}$
where $\pi_{i}\left(\cdot\right)$ is a random permutation, different
for each $i$ (i.e. we shuffled each pixel across samples). In this
way, each pixel $i$ has the same bias as in the original dataset,
$\mathbb{E_{\mu}}\left[\xi_{i}^{\prime}\right]=\mathbb{E}_{\mu}\left[\xi_{i}\right]$,
but the (connected) correlations are destroyed, $\mathbb{E}_{\mu}\left[\xi_{i}^{\prime}\xi_{j}^{\prime}\right]\approx\mathbb{E}_{\mu}\left[\xi_{i}^{\prime}\right]\mathbb{E}_{\mu}\left[\xi_{j}^{\prime}\right]$,
and furthermore the patterns $\xi^{\mu\prime}$ no longer carry information
about the target label $\sigma^{\mu}$, so that no generalization
is possible. As a result, the randomized patterns carry more information
per pixel that needs to be stored by the device, which in turn can
be expected to make the learning problem harder.

We produced $5$ such shuffled datasets and performed $10$ tests
on each with the same algorithms as for the Fashion-MNIST tests, using
the same parameters when possible. The exception was eLAL, which we
had to tweak slightly to avoid divergencies: we used $\eta=5\cdot10^{-4}$,
$\lambda_{0}=20$, $\lambda_{1}=1\cdot10^{-2}$.

\begin{figure}
\begin{centering}
\includegraphics[width=0.6\columnwidth]{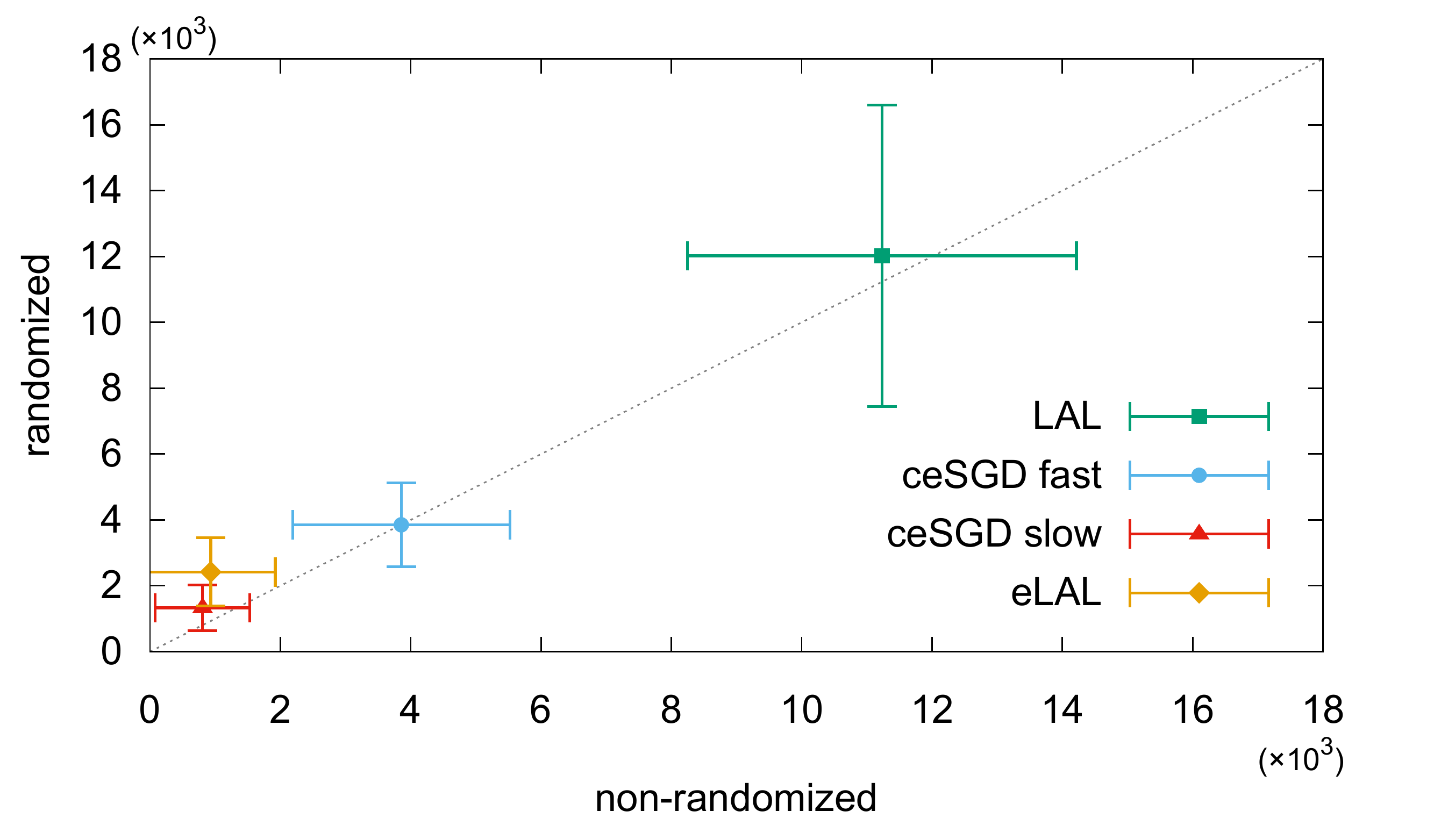}
\par\end{centering}
\caption{\label{fig:A-maxev_plain_vs_rand}Comparison of the distributions
of the maximum eigenvalues (means and stadard deviations) between
the original Fashion-MNIST dataset and its randomized version, for
various algorithms. The ``non-randomized'' values use the same data
as the clouds of points of fig.~\ref{fig:fashion}, bottom panel).}
\end{figure}

We directly compared the results with those obtained on the original
dataset. Fig.~\ref{fig:A-maxev_plain_vs_rand} shows that the maximum
eigenvalues (cf.~fig.~\ref{fig:fashion}, bottom panel) hardly change
between the two tests, with only a slight degradation for the eLAL
algorithm.

\begin{figure}
\begin{centering}
\includegraphics[width=0.6\columnwidth]{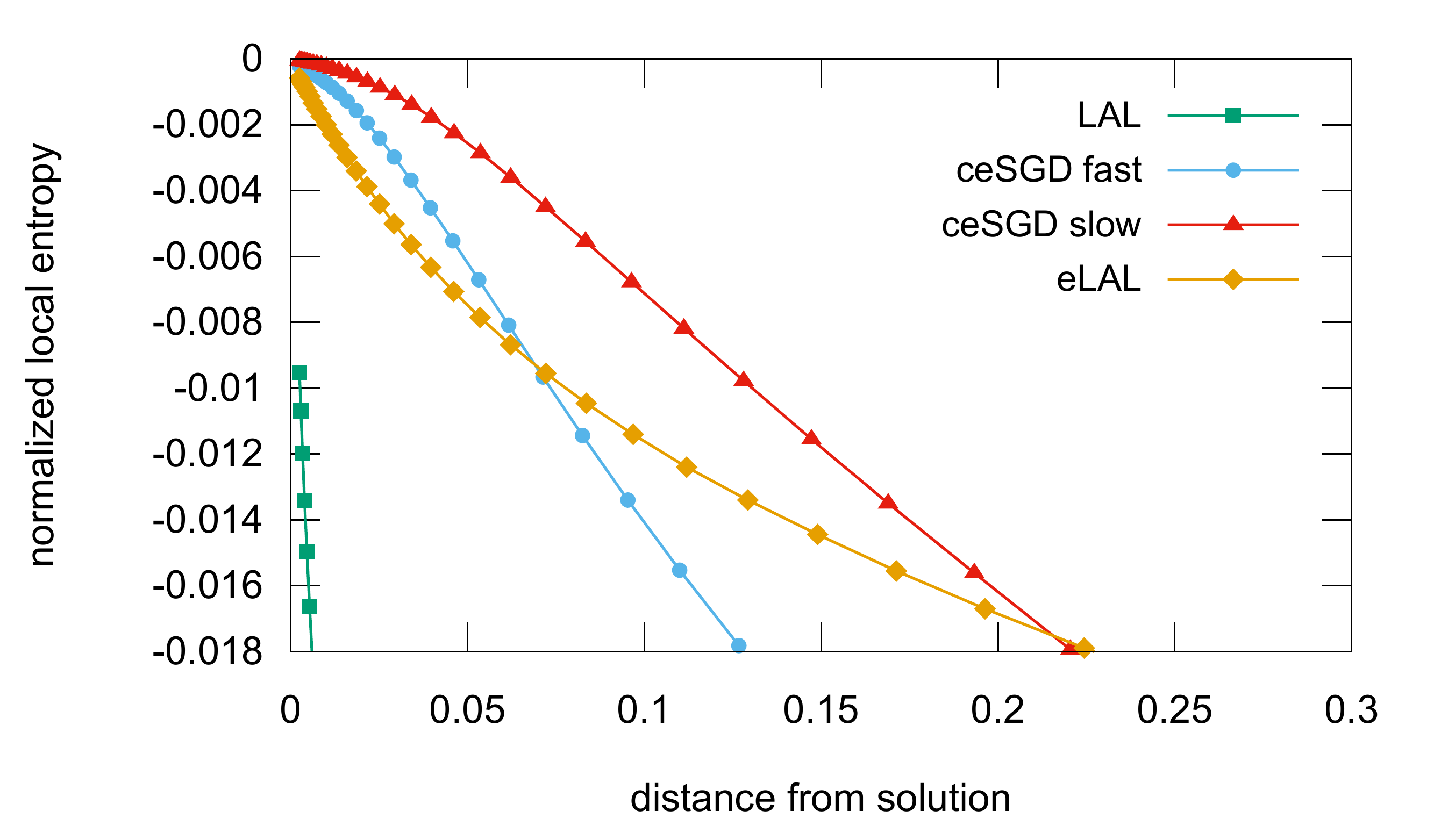}
\par\end{centering}
\caption{\label{fig:A-fmnist_wef_rndzall}Normalized local entropies as a function
of the distance for solutions found by various algorithms, using the
randomized-Fashion-MNIST dataset. This is the analogous of fig.~\ref{fig:fashion},
top panel, on the randomized dataset.}
\end{figure}

Fig.~\ref{fig:A-fmnist_wef_rndzall} shows the volumes around the
solutions, computed with the BP algorithm; this is the analogous of
fig.~\ref{fig:fashion}, top panel. We also kept the same scale for
easiness of comparison. Again, we observe a slight degradation of
the eLAL algorithm compared to ceSGD, albeit only at short distances.
It is still the case that LAL is by far the worst algorithm, and that
ceSGD slow is better than ceSGD fast. The volumes are overall smaller
than for the original dataset.

\begin{figure}
\begin{centering}
\includegraphics[width=0.6\columnwidth]{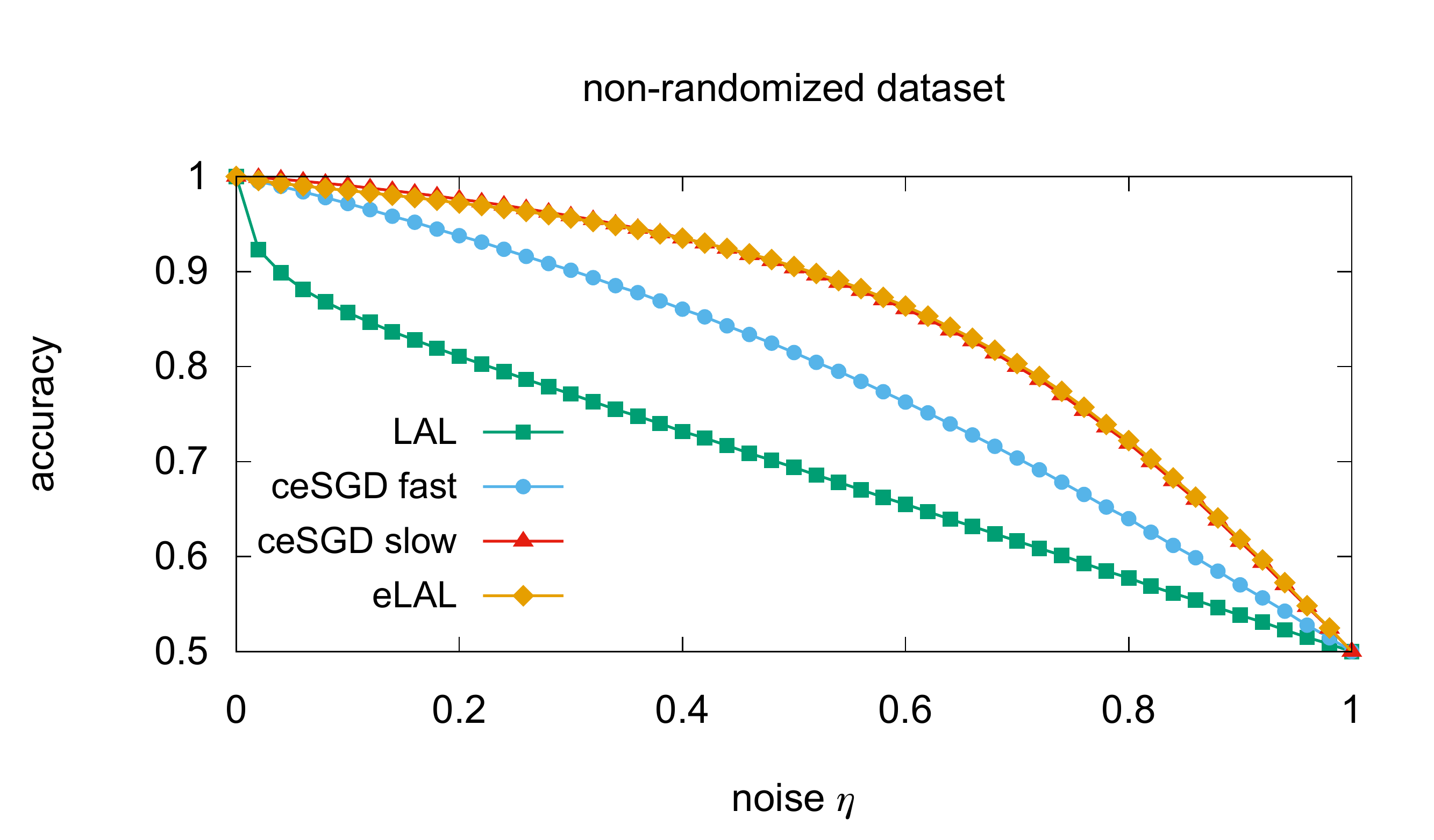}
\par\end{centering}
\begin{centering}
\includegraphics[width=0.6\columnwidth]{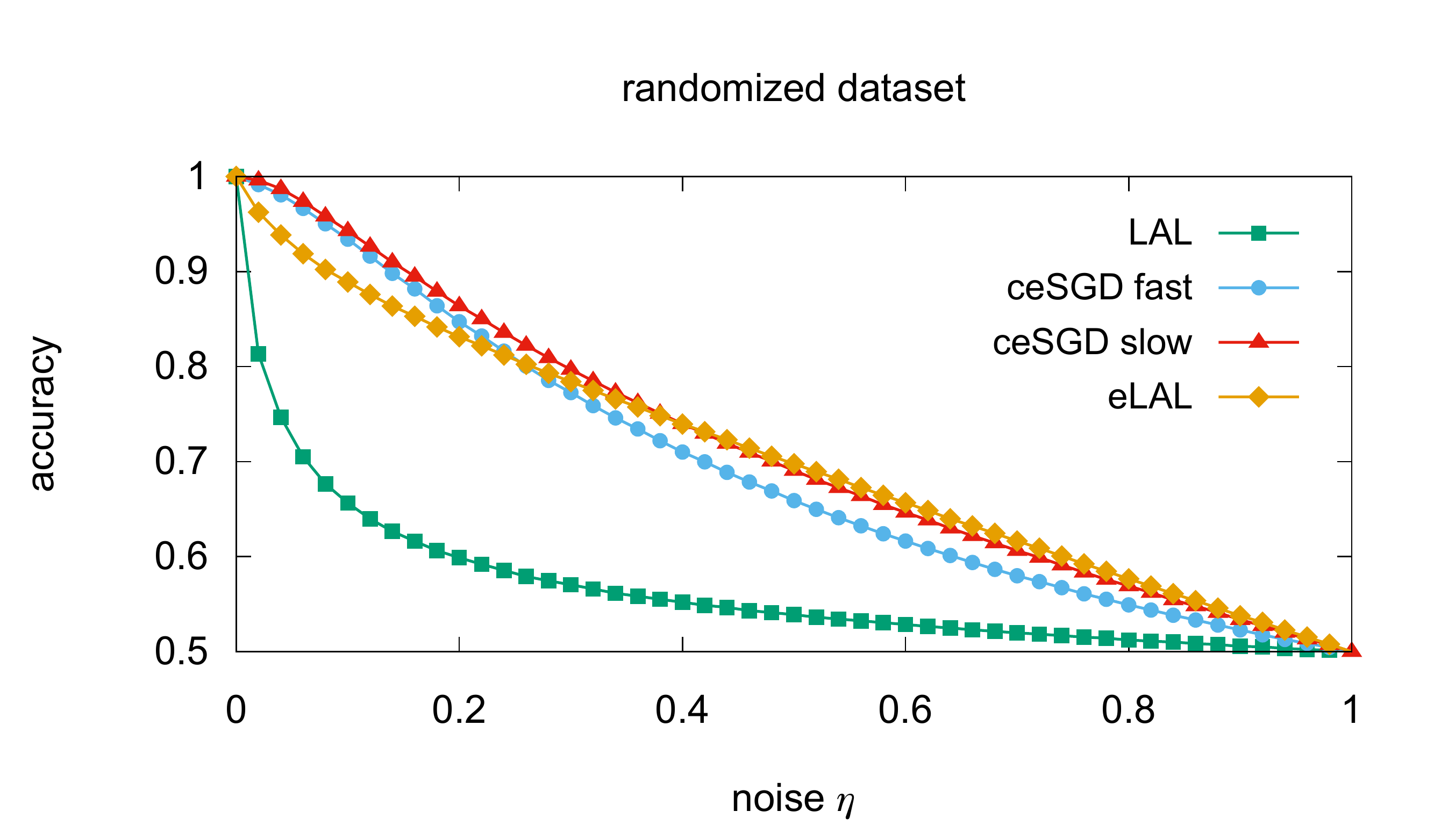}
\par\end{centering}
\caption{\label{fig:A-perturb_tests}Training accuracy in the presence of (biased)
noise, for the original Fashion-MNIST dataset (top panel) and for
its randomized version (bottom panel). In the top panel, eLAL and
ceSGD slow are hardly distinguishable.}
\end{figure}

We also performed an additional experiment on both datasets, measuring
the robustness towards the presence of noise in the input. For each
input image, the noise was added by replacing a randomly selected
fraction $\eta$ of pixels with random binary values. Each new value
of a pixel $i$ was extracted with the same bias observed in the original
dataset for that pixel, $\mathbb{E}_{\mu}\left[\xi_{i}\right]$: in
this way, the new corrupted images were still rather close to the
original input distribution (although the correlation with the desired
output label, and the internal correlations between pixels, were degraded)
and the networks would be able to operate in the same regime in which
they were trained. This model of noise is intended to provide a rough
proxy for the generalization capabilities of the networks without
the need for a validation set, since it measures the amount of overfitting
within a manifold which should approximate the distribution of the
data to be classified. The results are shown in fig.~\ref{fig:A-perturb_tests},
and they are fully consistent with the picture emerging from the study
of the local volumes.
\end{document}